\documentclass[10pt,journal,compsoc]{IEEEtran}
\usepackage{graphicx}
\usepackage{amsmath,amssymb} % define this before the line numbering.
\usepackage{color}
\usepackage{todonotes}

\usepackage{algorithm}
\usepackage{algpseudocode}  
\usepackage{multirow}
\usepackage{makecell}
\usepackage{booktabs}
\usepackage{subfigure}
\usepackage{hyperref}
\usepackage{url}
\usepackage{ulem}
\usepackage{tabularx,booktabs}
\usepackage{arydshln}
\ifCLASSOPTIONcompsoc
  \usepackage[nocompress]{cite}
\else
  \usepackage{cite}
\fi
\ifCLASSINFOpdf
\else
\fi
\begin{document}
%$\mathbb{Z}$

\title{An Open-source Benchmark of Deep Learning Models for Audio-visual Apparent and Self-reported Personality Recognition}

\author{Rongfan Liao,~\IEEEmembership{}
        Siyang Song$^*$, and
        Hatice Gunes

\IEEEcompsocitemizethanks{\IEEEcompsocthanksitem Rongfan Liao is with SONY China Software Center. E-mail: rongfan.liao@sony.com 

\IEEEcompsocthanksitem Siyang Song and Hatice Gunes are with the AFAR Lab, Department of Computer Science and Technology, University of Cambridge, Cambridge, CB3 0FT, United Kingdom.
E-mail: ss2796@cam.ac.uk, Hatice.Gunes@cl.cam.ac.uk
($^*$ Corresponding Author: Siyang Song, E-mail: ss2796@cam.ac.uk)
}% <-this % stops a space
\thanks{Manuscript accepted for publication on January 22, 2024.}}


\markboth{IEEE TRANSACTIONS ON , ~Vol.~14, No.~8, August~2015}%
{Shell \MakeLowercase{\textit{et al.}}: Bare Advanced Demo of IEEEtran.cls for IEEE Computer Society Journals}



\IEEEtitleabstractindextext{%
\begin{abstract}

%%% https://reader.elsevier.com/reader/sd/pii/S095070512100962X?token=DA169188B2517CC25B3A45EFA91F6481EBB8265980FE4CACD1D452DFF04F3D31E953925D146EF278D76F60A53B418DD4&originRegion=eu-west-1&originCreation=20220805140616

Personality determines a wide variety of human daily and working behaviours, and is crucial for understanding human internal and external states. In recent years, a large number of automatic personality computing approaches have been developed to predict either the apparent personality or self-reported personality of the subject based on non-verbal audio-visual behaviours. However, the majority of them suffer from complex and dataset-specific pre-processing steps and model training tricks. In the absence of a standardized benchmark with consistent experimental settings, it is not only impossible to fairly compare the real performances of these personality computing models but also makes them difficult to be reproduced. In this paper, we present the first reproducible audio-visual benchmarking framework to provide a fair and consistent evaluation of eight existing personality computing models (e.g., audio, visual and audio-visual) and seven standard deep learning models on both self-reported and apparent personality recognition tasks. Building upon a set of benchmarked models, we also investigate the impact of two previously-used long-term modelling strategies for summarising short-term/frame-level predictions on personality computing results. We conduct a comprehensive investigation into all the benchmarked models to demonstrate their capabilities in modelling personality traits on two publicly available datasets, audio-visual apparent personality (ChaLearn First Impression) and self-reported personality (UDIVA) datasets. The experimental results conclude: (i) apparent personality traits, inferred from facial behaviours by most benchmarked deep learning models, show more reliability than self-reported ones; (ii) visual models frequently achieved superior performances than audio models on personality recognition; (iii) non-verbal behaviours contribute differently in predicting different personality traits; and (iv) our reproduced personality computing models generally achieved worse performances than their original reported results. We make all the code and settings of this personality computing benchmark publicly available at \url{https://github.com/liaorongfan/DeepPersonality}.

% Personality is crucial for understanding human internal and external states. The majority of existing personality computing approaches suffer from complex and dataset-specific pre-processing steps and model training tricks. In the absence of a standardized benchmark with consistent experimental settings, it is not only impossible to fairly compare the real performances of these personality computing models but also makes them difficult to be reproduced. In this paper, we present the first reproducible audio-visual benchmarking framework to provide a fair and consistent evaluation of eight existing personality computing models (e.g., audio, visual and audio-visual) and seven standard deep learning models on both self-reported and apparent personality recognition tasks. We conduct a comprehensive investigation into all the benchmarked models to demonstrate their capabilities in modelling personality traits on two publicly available datasets, audio-visual apparent personality (ChaLearn First Impression) and self-reported personality (UDIVA) datasets. The experimental results conclude: (i) apparent personality traits, inferred from facial behaviours by most benchmarked deep learning models, show more reliability than self-reported ones; (ii) visual models frequently achieved superior performances than audio models on personality recognition; and (iii) non-verbal behaviours contribute differently in predicting different personality traits. 

\end{abstract}

%%% 2022/10/02：1. inclusion and exclusion criteria

\begin{IEEEkeywords}
Self-reported (true) personality recognition, Apparent personality (impression) recognition, Audio-visual personality computing benchmark, Spatio-temporal modelling, Deep Learning
\end{IEEEkeywords}}

\maketitle

% is not selected <OR> if conference mode is selected - because compsoc
% conference papers position the abstract like regular (non-compsoc)
\IEEEdisplaynontitleabstractindextext

% \fi
\IEEEpeerreviewmaketitle

\ifCLASSOPTIONcompsoc
\IEEEraisesectionheading{\section{Introduction}\label{sec:introduction}}
\else
\section{Introduction}
\label{sec:introduction}
\fi

\IEEEPARstart{P}{ersonality} defines a characteristic sets of human cognitive processes and emotional patterns that evolve from various biological and environmental factors \cite{corr2020cambridge}, which are well associated with a wide range of human behaviours and status such as purchasing behaviours \cite{youn2000impulse}, health conditions \cite{sano2015recognizing,jaiswal2019automatic}, social relationships \cite{asendorpf1998personality}, and even criminal activities \cite{briley2014genetic}. Consequently, automatic personality computing systems have drawn a lot of  attention in recent years, and have been frequently developed for real-world human behaviour understanding applications, including computer assisted tutoring systems \cite{afzal2019personality}, human resource management \cite{mansour2021relating}, job  interviews \cite{kaya2017multi}, and recommendation systems \cite{xia2014socially}. In these systems, the complex personality 
is usually described by trait-based models \cite{EYSENCK1991773, mitchell2007analysis,de2000cloninger,mccrae2008five} which focus on modelling personality aspects that are stable over time for the target person but differ in others \cite{matthews2003personality}.

Existing automatic personality computing approaches can be categorized into two types: (i) Self-reported personality recognition (SPR) that recognises the target subject's true personality traits; and (ii) apparent personality recognition (APR) that predicts external human observers' impression on the target subject (also called first impression or apparent personality). The majority of these approaches were focused on APR and were evaluated on the ChaLearn First Impression dataset \cite{ponce2016chalearn}. They recognise apparent personality using either human visual behaviours (e.g., facial behaviours) \cite{interpret_img,song2022learning,moreno2020estimation,helm2020single,eddine2017personality,tellamekala2022dimensional,gurpinar2016combining,persemon,beyan2019personality,persemon} or audio-visual behaviours \cite{deep_bimodal,bi_modal_lstm,gurpinar2016multimodal,crnet,aslan2019multimodal}. More specifically, a large part of these approaches \cite{8066355,audiovisual_resnet,interpret_img,deep_bimodal,eddine2017personality,tellamekala2022dimensional} attempt to infer personality traits from every single frame or short segment \cite{moreno2020estimation}, and then make the clip-level personality prediction by combining all frame/segment-level predictions, where video-level personality labels were used as the frame/short segment-level labels to train models. While subjects with different personalities can behave similarly in a frame or a short video segment, models trained by such strategies would be problematic as they pair similar behaviour samples with different personality labels \cite{song2020spectral}. Alternatively, recent studies frequently emphasized that long-term behaviours enable more reliable personality inference \cite{song2021self,crnet,beyan2019personality}, as personality traits are stable over time \cite{matthews2003personality}. These approaches propose to encode a clip-level personality representation from long-term behaviours of the subject, and generally achieve superior recognition results than most frame/segment-level approaches.

However existing approaches usually employ different pre-processing, post-processing and training strategies. Such differences lead to their reported results not being fairly reflected by their models' capabilities in recognizing personality traits. Meanwhile, very few deep learning-based approaches \cite{song2022learning,curto2021dyadformer,shao2021personality,dodd2023framework} have been proposed to recognise self-reported personality traits. They infer personality traits from audio \cite{welch2021listeners}, static image \cite{xu2021prediction}, video \cite{hickman2022automated}, or audio-visual clip \cite{salam2022learning} recorded from various scenarios, such as dyadic dialogue, self-evaluating surveys and self-interviews \cite{kassab2023vptd}. In summary, there is no comprehensive study that fairly demonstrates and compares the performances of existing audio-visual personality computing models and widely-used deep learning models on both APR and SPR (\textbf{Research gap 1}). Moreover, only a small number of them made their codes publicly available, which are even built on different platforms, e.g., Keras \cite{moreno2020estimation, helm2020single}, TensorFlow \cite{hayat2019use}, caffe \cite{persemon}, Torch \cite{bi_modal_lstm}, Matlab \cite{gurpinar2016multimodal,eddine2017personality} and Chainer \cite{2016} (listed in Table \ref{tb:avg_chalearn_10fold}). Also, some of these codes are incomplete and unimplementable. As a result, it is difficult for other researchers to reproduce or extend most of them \textbf{(Research gap 2)}.

% \textcolor{blue}{In summary, There exists two crucial research gaps in personality computing field}: \textcolor{blue}{(1)} there is no comprehensive study that fairly demonstrates and compares the performances of existing personality computing models and widely-used deep learning models \textcolor{blue}{on} both apparent personality and self-reported personality recognition (\textbf{Research gap 1}) and \textcolor{blue}{(2)} there is no \textcolor{blue}{easily-implemented} and well-evaluated baseline for incoming researchers to develop \textcolor{blue}{new approaches based on previously published methods or try existed ones.}  (\textbf{Research gap 2}).

In this paper, we bridge the aforementioned research gaps by introducing two main contributions: (i) we benchmark eight existing audio-visual or visual only automatic apparent personality recognition approaches as well as seven widely-used static or spatio-temporal deep learning models with standardized pre-processing, training and post-processing strategies, on both widely-used self-reported personality (i.e., UDIVA \cite{palmero2021context}) and apparent personality datasets (ChaLearn First Impression \cite{junior2019first}). The reported results provide a fair evaluation and comparison of all these models' capabilities for both apparent personality and self-reported personality recognition tasks (addressing research gap 1); and (ii) we make the code of all benchmarked models as well as their standardized pre-processing, training and post-processing scripts publicly available to further the science of \emph{automatic personality recognition}. This provides new researchers entering this field a set of strong personality computing baselines, which will facilitate their exploration of new models and will enable them to apply these to new datasets in personality computing. To the best of our knowledge, this is the first study that benchmarks and fairly evaluates the existing personality computing approaches and standard deep learning models on both SPR and APR tasks, with consistent and reproducible settings.

%%% 2022/05/13: 根据最终结果还需修改这段
% Since both short-term and long-term non-verbal behaviours may contain crucial cues related to personality, we additionally propose a two-stage framework that models both short-term and long-term audio-visual behaviours for true personality and perception recognition. This flexible framework can applies any of the benchmarking models or another standard deep learning model for the short-term behaviour modelling stage, while most existing standard time-series models can be applied to the long-term behaviour modelling stage. We show that the proposed two-stage framework can further improve the personality recognition for \textcolor{red}{most} benchmarked models. 

% To the best of our knowledge, this is the first work that benchmarks and fairly evaluates multiple existing personality computing approaches and standard deep learning models on both true personality and apparent personality recognition tasks, with consistent experimental settings. Our \textcolor{red}{benchmark code} are made publicly available at \footnote{\url{https://anonymous.4open.science}} to provide researchers a set of strong baselines and facilitate them to explore new models and arbitrary datasets in personality computing. 

% \setlength{\tabcolsep}{1pt}

\newcolumntype{C}{>{\centering\arraybackslash}X}
\setlength{\extrarowheight}{1pt}
\begin{table*}[ht]
\caption{Publicly available code repositories of existing automatic APR approaches, which can be categorised into three types based on their usability: (1) the fully-reproducible approaches that provide all python-based training/inference/model scripts; (2) the approaches whose code repositories are somehow problematic due to bugs, not-maintained code frameworks or incomplete code; and (3) the code repositories that do not rely on widely-used deep learning python libraries (Pytorch, Tensorflow, Keras, etc.).}
	\begin{center}
    \resizebox{1\linewidth}{!} {
		\begin{tabular}{l  l  l  l  l   }
			\toprule
                & Method & Platform  & Link  & Details    \\
            \hline 
            \multirow{2}*{(1)} 
                & Moreno-Armendazi et al.\cite{moreno2020estimation}  &      Keras       &   github.com/miguelmore/personality           &   Jupyter notebook-based script         \\ 
                & Helm et al.\cite{ helm2020single}                   &      Keras       &   github.com/dahe-cvl/apa\_paper         &          Python2.7-based code           \\ 
            
            \hline
            %\cdashline{1-5}[1pt/1pt]
            
            \multirow{3}*{(2)}
                & Hayat et al.\cite{hayat2019use}                     &   TensorFlow     &   github.com/HassanHayat08       &          Unimplementable (Bugs exist)   \\

                & Güclütürk et al.\cite{2016}                         &      Chainer     &   github.com/yagguc/deep\_impression        &         Training code is missing / ResNet-based approach        \\
                
                & Zhang et al. \cite{zhang2016deep}                   &     TensorFlow   &   github.com/zishansami102/First-Impression &      Unimplementable (Bugs exist)  \\              
                \hline              
                
            \multirow{4}*{(3)}
                & Zhang et al.\cite{persemon}                         &      Caffe       &   github.com/ZhangLeUestc/PersEmoN       &    C++ based code                       \\
                & Subraman et al.\cite{bi_modal_lstm}                 &     Torch (Lua)  &   github.com/InnovArul/first-impressions       &        Torch (Lua) is out of date  \\

                & Gürpinar et al. \cite{gurpinar2016multimodal}       &     MatConv      &   github.com/frkngrpnr/lapfi       &    MATLAB based code       \\
                & Bekhouche et al. \cite{eddine2017personality}       &     MatConv       &   github.com/Bekhouche/CVPR2017       &  MATLAB based code   \\
			\bottomrule
		\end{tabular}
        }
	\end{center}
\label{tb:avg_chalearn_10fold}
\end{table*}
\setlength{\tabcolsep}{1.4pt}

\section{Related Work}

\noindent In this section, we systematically review and summarize recent audio-visual automatic apparent personality recognition  (Sec. \ref{subsec:APP}) and self-reported personality recognition (Sec. \ref{subsec:APR}) approaches.

\subsection{Automatic apparent personality recognition}
\label{subsec:APP}

\noindent Existing approaches frequently attempt to recognize apparent personality from non-verbal facial behaviours, where a large part of them infer apparent personality only from a single facial display \cite{interpret_img,moreno2020estimation,helm2020single}. Joo et al. \cite{joo2015automated} conduct studies on  face images of $650$ American politicians of white ethnicity, which extracts Histogram of oriented gradients (HOG) features as the low-level static facial representation to predict personality traits. Dhall et al. \cite{dhall2016first} utilize both hand-crafted and deep-learned features to describe   Twitter profile facial images and infer personality traits from them, where background information was also considered. Moreno-Armendáriz et al. \cite{moreno2020estimation} first extract one portrait picture as the representation for each video, and then train a CNN to predict video-level perception. In addition, although facial videos are provided, some studies also directly infer apparent personality from static facial appearance without considering temporal dynamics. For example, Ventura et al.\cite{interpret_img} use Classification Activation Map (CAM) \cite{zhou2016learning} to investigate the relationship between facial expressions and apparent personality, which feeds each frame to a CNN to infer video-level perception traits.

While very few studies \cite{hayat2019use} infer personality traits from only audio behaviours, the majority of the studies focus on audio-visual methods that frequently show enhanced performances than the corresponding single modality (e.g., visual-based or audio-based) systems. Besides some early studies \cite{celiktutan2014continuous,celiktutan2014maptraits} investigated the relationship between frame-level facial cues and apparent personality traits in the context of human-robot interactions, most existing approaches attempt to predict video-level personality traits. Zhang et al. \cite{deep_bimodal} first choose 6 images from each video, and extend the Descriptor Aggregation Network (DAN) to deep learn facial perception features at the frame-level. Meanwhile, a linear regressor is used to process log filter bank-based clip-level audio representation. The late fusion scheme is employed to get the final prediction by averaging audio and visual predictions. Gucluturk et al. \cite{audiovisual_resnet} propose a two-stream ResNet to deep learn both audio and visual personality cues at the frame-level, which are combined at fully connected (FC) layer to predict perceptions. Subramaniam et al.\cite{bi_modal_lstm} divide each audio-visual clip into several short segments, where one facial frame is selected from each segment to produce a segment-level visual feature. Then, the segment-level audio-visual features are combined via a FC layer to make the apparent personality prediction. To recognize apparent personality from clip-level (long-term) audio-visual behaviours, Gürpinar et al.\cite{gurpinar2016multimodal} employ a pre-trained network to extract frame-level facial emotion features. The video-level visual representation is then obtained by computing statistics of all frame-level features while clip-level audio feature is extracted via OpenSMILE \cite{eyben2010opensmile}. The final prediction is made by the weighted average of audio and visual predictions. As discussed in Sec. \ref{sec:introduction}, these static or short-term behaviour-based approaches usually re-use video-level personality labels as the frame/segment-level labels during the training, which may result in problematic models.

To avoid the above problem, many recent approaches are proposed to recognize apparent personality from long-term behaviours (i.e., applying video-level behaviour representation to recognize video-level personality traits). For example, Beyan et al. \cite{beyan2019personality} summarize a video into a set of dynamic images, and then select a small part of key dynamic images based on their spatio-temporal saliency. Then, the produced key dynamic image sequence is used as the video-level representation for the perception prediction. Helm et al.\cite{helm2020single} extract a face image sequence by down-sampling the target video, whose features is used as the video-level facial behaviour representation. Then, a 3D-CNN is employed to infer apparent personality from this video-level representation. Gürpınar et al.\cite{gurpinar2016combining} jointly extract facial expression-related features from all aligned face images and the ambient information from the first frame of a down-sampled video, both of which are combined as the video-level visual representation for apparent personality prediction. Li et al. \cite{crnet} propose to down-sample each video into $32$ frames, and extract video-level visual features from the corresponding facial region sequence as well as the whole image sequence. In addition, the long-term audio features and text features are also combined with the visual features for APR. Since the above long-term behaviour modelling approach depend on down-sample videos, Song et al. \cite{song2021self} propose a self-supervised learning strategy to use all frames for video-level representation construction. This method trains a person-specific layer for each individual using facial behaviours of the whole video, whose weights are then used as the video-level personality representation.

\subsection{Automatic self-reported personality recognition}
\label{subsec:APR}

\noindent Compared to apparent personality, only a small number of studies attempt to infer self-reported personality traits. Qin et al. \cite{qin2018modern} extract five types of hand-crafted features (i.e., HOG, Local Binary Pattern (LBP), Gabor, Scale-invariant feature transform (SIFT) and Generalized Search Trees (Gist)) from each face image, which are fed to standard regressors (e.g., decision tree) to estimate self-reported personality (16PF) and intelligence. Besides such frame-level SPR approaches, recently proposed audio-visual self-reported personality analysis datasets \cite{palmero2021context,cafaro2017noxi} lead some studies to develop video-level SPR approaches. Similar to APR solutions, Curto et al. \cite{curto2021dyadformer} propose a Transformer-based model to extract individual and interpersonal behaviour features from each short dyadic interaction segment using variable time windows, which can jointly recognize self-reported personality traits for both individuals. In addtion, Celiktutan et al. \cite{celiktutan2017multimodal} specifically investigated the SPR under the human-robot interaction scenarios. Since personality is well associated with human cognitive process, Song and Shao et al. \cite{song2022learning,shao2021personality} propose the first audio-visual approach that employs Neural Architecture Search (NAS) to explore a person-specific network for each subject using all available frames, whose architecture and parameters are encoded as the person-specific cognitive process graph representation for SPR. This NAS-based idea was then followed by \cite{salam2022learning}, which also search a personalized deep learning architectures for each subject to recognize Big Five personality traits.

\section{The proposed benchmarking framework}

\begin{figure*}
\centering 
\includegraphics[width=0.96\textwidth]{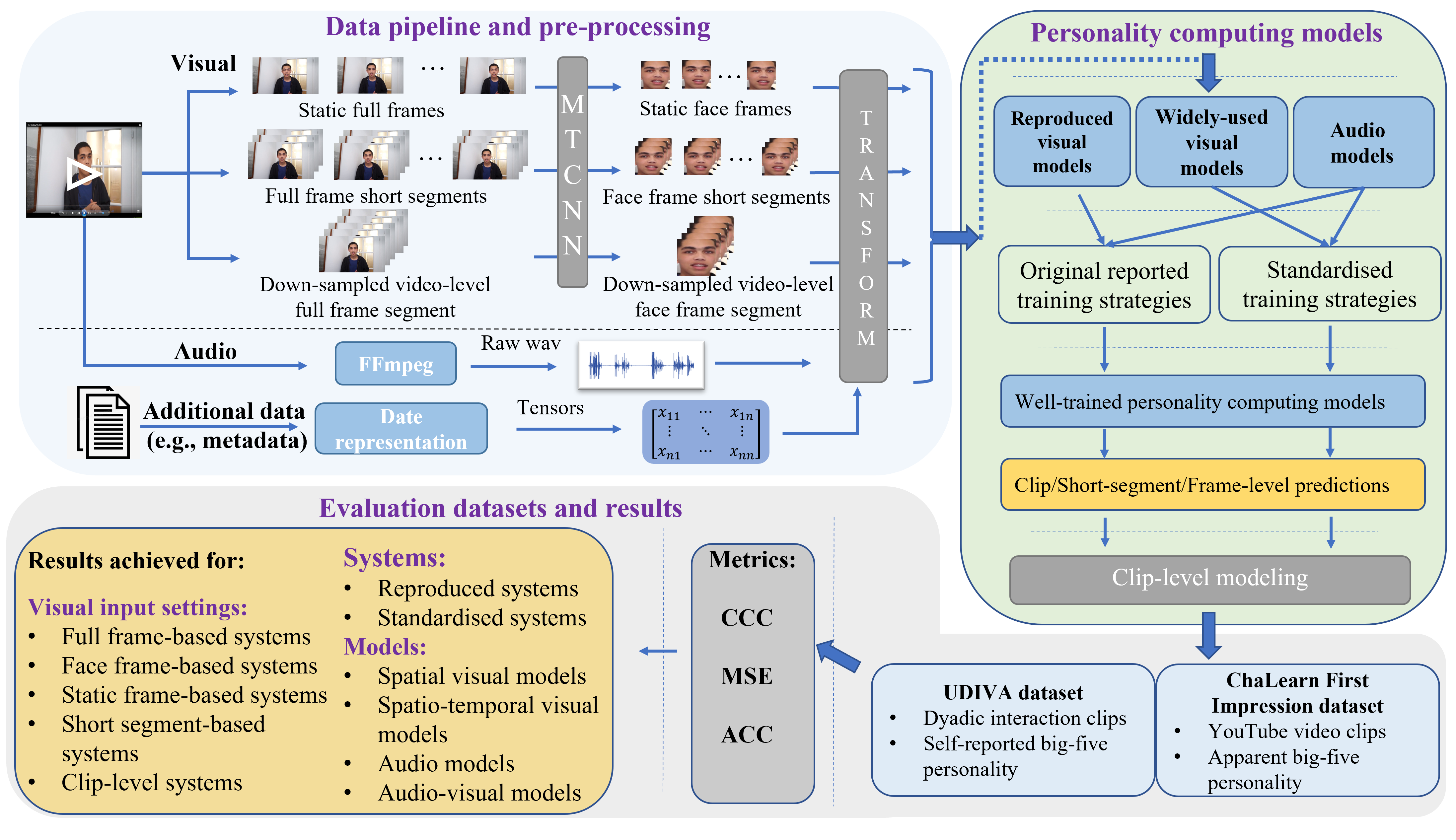} 
\caption{Illustration of our personality computing benchmark framework.} 
\label{Fig:pipeline}
\end{figure*}

\noindent In this paper, we propose a benchmarking framework that aims to provide a rigorous and reproducible evaluation of the existing personality computing models and the widely-used deep learning models for both automatic self-reported and apparent personality recognition. This framework will equip the future researchers with a set of strong audio-visual personality computing baselines, with standardized data pre-processing, training and post-processing strategies. Specifically, this section introduces our benchmarking framework by presenting its coding infrastructure (Sec. \ref{subsec: Coding}), benchmarked models and their settings (Sec. \ref{subsec: models}), and datasets used for evaluation (Sec. \ref{subsec: dataset}). The pipeline of the proposed personality benchmarking framework is illustrated in Fig. \ref{Fig:pipeline}.

\subsection{Coding infrastructure}
\label{subsec: Coding}

\noindent The goal of our paper is to fairly compare deep learning models’ capabilities for personality computing. None of the previous studies show a fair comparison between existing models mainly due to four reasons. First, previous approaches frequently adopted different kinds of pre-trained weights. Second, different data pre-processing strategies have been employed. Third, the training and validation strategies vary from different approaches. This would lead to different criteria for choosing the final model weights. The aforementioned three variables may largely impact the personality computing model performance. Another potential reason for the performance boost after the release of the test dataset is that some model architectures and weights might be tuned according to the test dataset rather than the validation set. To facilitate a fair comparison, our benchmark framework emphasises a unified framework focusing on evaluating the deep learning models’ capacities in predicting personality traits. Specifically, our benchmarking framework unifies the components of the data input, data pre-processing, data post-processing, model initialisation, training, validation, evaluation and coding platform/libraries for all deep learning models. The only difference in experiments resides in different models' architectures and their training hyper-parameter settings, as the optimal hyper-parameter settings are model-dependent.

% \textcolor{blue}{
% This way, the personality computing results achieved for all benchmarked models can be compared based on the same premises. 
% Incremental improvements are often observed post-competition, and potential biases could be introduced by model cherry-picking or over-learning on specific datasets when a challenge is over and a test set is released. Based on our framework, by employing diverse and representative datasets during the model evaluation, dubious generalizations can also be reflected. 
% }

\subsubsection{Data input and pre-processing} 
For all experiments, we consistently employ the same training, validation and test sets that are predefined in the corresponding self-reported personality and apparent personality datasets for training the models and reporting the results. Instead of using complex and dataset-specific data pre-processing pipelines, we employ a widely-used and unified data pre-processing  pipeline for each compared model (e.g., static models and spatio-temporal models) as follows:
\begin{itemize}

    \item \textbf{Static image-based models:} We first evenly divide each video into $K$ short segments as suggested in \cite{8066355,bi_modal_lstm} and only select the one frame of each segment (To make experiments to be reproducible, we consistently use the first frame of each segment), where we follow the winner \cite{8066355} of ChaLearn 2016 impression challenge to set $K$ as $100$ for experiments conducted on ChaLearn First Impression dataset, and empirically set $K$ as $2000$ for UDVIA dataset. We then apply MTCNN \cite{zhang2016joint} to obtain an cropped and aligned face image from each selected frame. During training, validation or testing, we again follow  \cite{8066355} to resize the resolution of these images to $456 \times 256$, where only a $224 \times 224$ sub-region in the center of the image is used as the final full frame. Meanwhile, the face region is contained in a cropped image with the resolution of $112 \times 112$ using the Dlib library \cite{dlib}. Specifically, standard data augmentation including center crop, random horizontal flip and pixel normalization are employed to process all training images. We follow previous studies \cite{8066355,interpret_img,audiovisual_resnet} that compute personality traits prediction of each test video by averaging all frame-level predictions.

    \item \textbf{Spatio-temporal models:} For each spatio-temporal model, we benchmark two types of systems: (i) short segment-level system, where we divide each video into several segments, where each segment that contains $n$ frames (i.e., we empirically employed $n=32$ or $n=64$ according to (1) the setting employed in the original publication; and (2) the experimental results) is used as the input. The personality trait prediction of each test video is computed by averaging all segment-level predictions; and (ii) video-level system, where we again follow \cite{crnet} that evenly divides each video into $32$ segments, and select the first frame of each segment to construct a sequence of $32$ frames to represent the entire video. The sequence is then used as the input for spatio-temporal models to produce the video-level personality prediction. For both systems, MTCNN and Dlib library are again employed to obtain each full frame and its corresponding face image based on the same setting as the static image-based models.

    \item \textbf{Audio models:}  As most audio-visual studies did not provide the details of their raw audio signal extraction, we follow previous studies \cite{8066355,suman2022multi} to use the FFmpeg tool \footnote{https://github.com/FFmpeg/FFmpeg} to extract the raw audio signal from original audio-visual clips for all audio-based models. The processing details for each audio model are introduced in Sec. \ref{subsec: existing models}.

\end{itemize}

\subsubsection{Training, validation and testing protocol} 

For all benchmarked static models and short segment-based spatio-temporal models, we follow the previous studies \cite{8066355,interpret_img,audiovisual_resnet} to re-use the clip-level label as the corresponding frame-level or short segment-level labels to train them. We also use the clip-level label as the label to train models that take clip-level sequence as the input. For all experiments, we consistently employ the same training, validation and testing protocol. Each static image-based model or short segment-based spatio-temporal model is trained on the training dataset with the shuffled data samples and evaluated on the validation set, where early stopping is used to prevent overfitting. We choose the model that obtains the best performance on the validation set as the final model for each experiment, which generates the reported testing results. The detailed training hyper-parameter settings for all benchmarked models are listed in Table \ref{tab:training details}.

\subsubsection{Evaluation metrics}

\noindent To measure the performance of the trained models, we first employ two standard evaluation metrics (i.e., ACC and Mean Squared Error (MSE)) proposed in \cite{ponce2016chalearn} and \cite{palmero2021context}. The ACC is defined as:
\begin{equation}
    \mbox{ACC} = 1 - \frac{1}{N}\sum_{n = 1}^N | y_{p_n} - y_{g_n} |
\end{equation}
and the MSE can be computed as:
\begin{equation}
    \mbox{MSE} = \frac{1}{N} \sum_{n = 1}^N \left( y_{p_n} - y_{g_n}  \right)^2 
\end{equation}
where $N$ indicates the number of test videos; $y_{p_i}$ and $y_{g_i}$ denote the prediction and ground-truth of the $n_{th}$ test video, respectively. We also follow recent studies \cite{song2022learning,song2021self,tellamekala2022dimensional} that employ Concordance Correlation Coefficient (CCC, Eq. \ref{eq:ccc}) to measure the correlation between predictions and ground-truths, which is defined as:
\begin{equation}
\mbox{CCC} = \frac{2 \sigma_{Y_{p}} \sigma_{Y_g} \mbox{PCC}_{Y_p,Y_g}}{\sigma_{Y_p}^2 + \sigma_{Y_g}^2 + \left( \mu_{Y_p} - \mu_{Y_g}  \right)^2},
\label{eq:ccc}
\end{equation}
where $\mu_{Y_p}$ and $\mu_{Y_g}$ denote the mean values of predictions and ground-truths of all test videos, respectively; $\sigma_{Y_{p}}$ and $\sigma_{Y_{g}}$ are the corresponding standard deviations; and $\mbox{PCC}_{Y_p,Y_g}$ denotes the Pearson Correlation Coefficient (PCC) between $Y_p$ and $Y_g$. In this paper, we follow previous studies that individually employ MSE and CCC to evaluate the SPR systems while we use ACC and CCC to evaluate the APR systems.

\subsection{Benchmarked personality computing models}
\label{subsec: models}

\noindent In this section, we provide the details of all the benchmarked models including (i) seven visual models, six audio models and five audio-visual models that have been employed by existing personality computing approaches; (ii) seven typical spatial/spatio-temporal deep learning CNN/Transformer models that have been widely-used in image or video processing tasks but have not yet been employed for personality recognition; and (iii) two previously used clip-level encoding models that allow all frame/short segment-level personality predictions of a clip to be combined for clip-level personality recognition.

\subsubsection{Model inclusion and exclusion criteria}

\noindent To the best of our knowledge, there are more than one hundred visual or audio-visual personality computing models that have been proposed in the literature. In this paper, we propose the following inclusion and exclusion criteria to benchmark the most representative personality computing models:
\begin{itemize}

    \item \textbf{Inclusion criteria:} The main criteria of this benchmark is to choose representative approaches that have been evaluated and compared in the well-known and widely-used publicly available audio-visual personality computing dataset: ChaLearn First Impression. As a result, we first choose the Top-3 ranked models \cite{deep_bimodal,bi_modal_lstm,audiovisual_resnet} from the ChaLearn First Impression Recognition Challenge. Then, we choose the best end-to-end audio-visual deep learning models that have been published in recent three years, respectively, which (\cite{persemon} (2019), \cite{crnet} (2020), and \cite{suman2022multi} 2022) claimed that they have achieved the state-of-the-art performances on ChaLearn 2016 dataset and provided all details for model settings and training (the best model of 2021 \cite{song2021self} is not an end-to-end approach). In addition, we also reproduce two models \cite{interpret_img,hayat2019use} that can visualise the relationship between human audio/visual non-verbal behaviours and personality traits.

    % firstly, top ranked models on ChaLearn2016 challenge. We choose the top three leading methods including models from Zhang et al.\cite{deep_bimodal,bi_modal_lstm,audiovisual_resnet}. 

    % Secondly, recently published works claiming the state-of-the-art performances on ChaLearn2016 impression dataset, which are PersEmoN\cite{persemon}, CR-Net\cite{crnet} and multi-modal personality prediction system\cite{suman2022multi}.thirdly, models that bring insights or interpretability into personality recognition such as Interpret-CNN\cite{interpret_img} and Interpret-audios\cite{hayat2019use}

    % \item \textbf{Exclusion criteria:} In this paper, we exclude methods that are \textbf{not} based on: (i) audio-visual deep learning models; (ii) an end-to-end framework, such as \cite{tellamekala2022dimensional,song2021self,shao2021personality,song2022learning} which require relatively complex pre-processing or pre-training; (iii) evaluated on the ChaLearn 2016 dataset \cite{curto2021dyadformer,suen2019tensorflow,shao2021personality,song2022learning,curto2021dyadformer,salam2022learning}; (iv) using the Big-five personality traits as the target recognition task; and (v) treating personality recognition as a regression task. 

    \item \textbf{Exclusion criteria:} In this paper, we exclude personality computing methods that (i) are not based on audio-visual deep learning models; (ii) whose personality recognition network cannot be directly trained in an end-to-end manner, i.e., the ‘end-to-end’ mentioned here refers to the network that can be directly trained by pairing the pre-processed video signals or/and audio representations with personality traits labels. In contrast, methods that require separated training for different parts of the network  \cite{song2021self, 2016} or manual feature engineering \cite{ilmini2016persons} are excluded; (iii) were not evaluated on the ChaLearn First Impression dataset \cite{song2022learning,hayat2019use,2016,salam2022learning,suman2022multi}; (iv) did not employ the Big-Five personality traits as the target of the recognition task; and (v) did not treat personality recognition as a regression task.

\end{itemize}

\subsubsection{Existing personality computing models} 
\label{subsec: existing models}

\noindent We first present seven deep learning visual models, six audio models and five audio-visual models that have been proposed to recognize apparent personality traits, which are evaluated on the ChaLearn First Impression dataset \cite{ponce2016chalearn}. For each deep learning-based model, the last FC layer has five neurons to jointly predict five personality traits. The original low-level settings of these models for reproducing are provided in the Supplementary Material, where the original pre-processing strategies employed for these models may slightly different from our standardised pre-processing strategy (i.e., we strictly follow these settings to additionally reproduce these models). \\

\noindent \textbf{Visual models:} The seven existing personality computing visual models reproduced in this paper are explained as follows:
\begin{itemize}

    \item \textbf{(i) DAN \cite{8066355}:} DAN is a CNN model that infers personality from a static image (including both face and background). It consists of several convolution-ReLU blocks and an additional block that is equipped with both average-pooling and max-pooling, which is added between the last convolutional layer and the final FC layer.

    % Before being fed to the model, each image is resized to $456 \times 256$ resolution and MSE is employed as the loss function.

    % The training process is conducted using the SGD optimizer \cite{sgd} with initial learning rate of 0.05, weight decay of 0.0005 and momentum of 0.8

    \item \textbf{(ii) CAM-DAN$^{+}$ \cite{interpret_img}:} CAM-DAN$^{+}$ has a similar architecture to the DAN network, which also infers personality from each static image.  It applies a max pooling and an average pooling as two parallel branches in addition to convolution layers, and is equipped with a Class Activation Map (CAM) module in order to visualise the most salient facial regions for personality recognition.     
    % CAM-DAN$^{+}$ has a similar architecture to the DAN network, which also infers personality from each static image. This model applies a max pooling and an average pooling as two parallel branches in addition to convolution layers, and is equipped with a Class Activation Map (CAM) module in order to visualise the most salient facial regions for personality recognition. 

    % The training process is conducted using the SGD optimizer \cite{sgd} with initial learning rate of 0.05, weight decay of 0.0005 and momentum of 0.8, where MSE is employed as the loss function.

    \item \textbf{(iii) Bi-modal CNN-LSTM  \cite{bi_modal_lstm}:} The visual part of this approach consists of three 3D convolutional layers and two FC layers. It randomly takes six face images from each video as the video-level representation, and concatenates them as the clip-level input.
    % The model is trained using the SGD optimizer with initial learning rate of 0.05, weight decay of 0.0005 and momentum of 0.9, where MSE is employed as the loss function. 

    \item \textbf{(iv) ResNet  \cite{audiovisual_resnet}:} The visual part of this model is a $17$-layer deep residual network, where an FC layer is attached at the top of the model. 
    % It takes a static image (both face and background) as the input. 
    % The training process is conducted using the SGD optimizer with initial learning rate of 0.05, weight decay of 0.0005 and momentum of 0.9, where MSE is employed as the loss function.

    \item \textbf{(v) CRNet \cite{crnet}:} The visual module of the CR-Net consists of two $34$-layer streams to learn personality features from down-sampled video-level face sequences and full image sequences. Then, a CR-block is attached to obtain personality classification features, based on which the regression feature is extracted. The Extra Trees Regressor (ETR) is finally employed to predict personality trait intensities from the extracted regression features. In this system, each video is divided into $32$ segments, where a single frame is randomly selected from each segment (i.e., each video is down-sampled to 32 frames).
    
    %The visual module of the CR-Net consists of two streams, where each uses a $34$-layer deep residual network as the backbone to learn personality features from either down-sampled video-level face sequences and full image sequences. Then, a CR-block is attached to first obtain personality classification features (processed by softmax) from the latent feature produced by each backbone, based on which the regression feature is extracted. The Extra Trees Regressor (ETR) is finally employed to predict personality trait intensities from the extracted regression features. To obtain the video-level face sequence and full image sequence, each video is divided into $32$ segments, where a single frame is randomly selected from each segment (i.e., each video is down-sampled to 32 frames) and the Bell Loss function \cite{crnet} is also applied.}

    % The SGD optimizer with initial learning rate of 0.002, weight decay of 0.005 and momentum of 0.9 is employed for training, 

    \item \textbf{(vi) PersEmoN \cite{persemon}:} The model first employs four convolution layers and a FC layer to extract features from each face image of the video. Then, three parallel FC layers are used to simultaneously predict valence, arousal and personality traits. Besides, a coherence module (an FC layer) is introduced to process both images from the personality dataset and emotion dataset, allowing the final learned representation to be dataset invariant, while a RAM module that consists of two FC layers is employed to produce the apparent personality attributes from arousal and valence inputs, aiming to enhance the learned personality representations. 
    % The training process of this model is achieved by using the SGD optimizer with initial learning rate of 0.001, weight decay of 0.0005 and momentum of 0.9, while MSE is used as the loss function.

    \item \textbf{(vii) Amb-Fac \cite{suman2022multi}:} The visual part of this system is a two-stream ResNet-101 model that extracts two different visual features: ambient features (extracted from full frames) and facial features. The system consists of three steps:  (i) Pre-processing: the system selects six equally spaced images from each video clip, and applies MTCNN to obtain their face areas; (ii) ambient and facial feature extraction: ResNet-101 pre-trained on ImageNet is re-trained to extract ambient features and facial features from each selected frame; and (iii) the ambient features and facial features of each selected frame are individually fed to a MLP to make frame-level personality prediction. The video-level personality trait prediction is obtained by averaging all frame-level predictions.
    % The training process is conducted using the SGD optimizer with initial learning rate of 0.01, weight decay of 0.0005 and momentum of 0.9, where MSE is employed as the loss function.
    
    % \textbf{(vii) Amb-Fac \cite{suman2022multi}:} The visual part of this system is a two-stream ResNet-101 model that extracts two different visual features: ambient features (extracted from full frames) and facial features. The system consists of three steps:  (i) Pre-processing: the system selects six equally spaced images from each video clip, and applies MTCNN to obtain the face area for extracting facial features; (ii) ambient and facial feature extraction: ResNet-101 pre-trained on ImageNet is re-trained to extract ambient features and facial features from each selected frame; and (iii) the ambient features and facial features of each selected frame individually are fed to a MLP (containing FC layers and a sigmoid layer) to make frame-level personality prediction. The video-level personality trait prediction is obtained by averaging all frame-level predictions. 

\end{itemize}

For frame-level personality recognition models, all frame-level predictions of a clip are fused to obtain the final clip-level personality prediction, and thus the video-level human facial behaviours are partially considered. Such models are not theoretically optimal for SPR, as it is difficult to infer self-reported personality traits from a single frame. Since no previous study has comprehensively investigated their performances on SPR, this paper conducts the first comprehensive evaluation for them on both APR and SPR tasks. However, this frame-level SPR strategy needs to be questioned and deeply investigated before adopting it in future studies.

\noindent \textbf{Audio models:} The six existing personality computing audio models reproduced in this paper are explained as follows:
\begin{itemize}

    \item \textbf{(i) FFT \cite{interpre-aud}:} The clip-level raw audio signal is represented by a $61208$ dimensional vector. Meanwhile, the obtained vector is normalised to have zero mean and unit variance, and converted to the frequency domain using FFT. The produced spectral signal is then fed to a network that consists of three temporal convolution layers followed by max pooling and dropout.

    %\item \textbf{(i) FFT \cite{interpre-aud}:} The clip-level raw audio signal is obtained from each audio-visual clip at 4000 HZ, and is represented by a $61208$ dimensional vector. The obtained audio signal is normalised to have zero mean and unit variance, and converted to the frequency domain using FFT. The produced spectral signal is then fed to a network that consists of three temporal convolution layers followed by max pooling and dropout. 
    % Here, the training is conducted using SGD optimizer with initial learning rate of 0.00083, weight decay of 0.065 and a momentum of 0.9, and the MSE is applied as the loss function.  

    \item \textbf{(ii) MFCC/Logfbank \cite{8066355}:} The clip-level raw audio signal is obtained from each audio-visual clip at 44,100 Hz, which are divided into 3059 audio frames. A 13-D MFCC feature and a 26-D logfbank feature are extracted from each audio frame. Then, the clip-level 39767-D MFCC feature and 79534-D logfbank feature are produced by concatenating all frame-level MFCC and logfbank features, respectively.

    % \item \textcolor{blue}{\textbf{(ii) MFCC/Logfbank \cite{8066355}:} The clip-level raw audio signal is obtained from each audio-visual clip at 44,100 Hz, which are divided into 3059 audio frames. A 13-D MFCC feature and a 26-D logfbank feature are extracted from each audio frame. Then, the clip-level 39767-D MFCC feature and 79534-D logfbank feature are produced by concatenating all frame-level MFCC and logfbank features, respectively.}     
    % During the model training, SGD optimizer with initial learning rate of 0.05, weight decay of 0.0005 and momentum of 0.8 is applied, and the MSE is employed as the loss function.

    \item \textbf{(iii) Bi-modal CNN-LSTM \cite{bi_modal_lstm}:} The audio signal of each clip is first divided into six non-overlapping segments, where the mean and standard deviation of several properties of each audio segment are extracted, and combined as a 68 dimension vector using pyAudioAnalysis \cite{pyaudioanalysis}. We then concatenate six vectors from six segments and fed it to a FC-LSTM-FC block to jointly predict five personality traits.
    %\item \textcolor{blue}{\textbf{(iii) Bi-modal CNN-LSTM \cite{bi_modal_lstm}:} The audio signal of each clip is first divided into six non-overlapping segments, where the mean and standard deviation of several properties of each audio segment are extracted, and combined as a 68 dimension vector using pyAudioAnalysis \cite{pyaudioanalysis}. We then concatenate six vectors from six segments as a matrix with size of $6 \times 68$, which is fed to a FC-LSTM-FC block to jointly predict five personality traits.}     
    % During the model training, SGD optimizer with initial learning rate of 0.05, weight decay of 0.0005 and momentum of 0.9 is applied, and the MSE is employed as the loss function.

    \item \textbf{(iv) ResNet \cite{audiovisual_resnet}:} The entire audio signal of each clip is represented as a 244,832-D vector using the librosa library \cite{librosa}. For each training iteration, it randomly selects a continuous 50,176-D sub-vector from the clip-level vector as the clip-level audio representation. Then, a $17$-layer ResNet is employed, where a FC layer is attached at the top of it.     
    
    % \item \textbf{(iv) ResNet \cite{audiovisual_resnet}:} The entire audio signal of each clip is sampled at 16,000 HZ using the librosa library\cite{librosa}, resulting in a 244,832-D vector to represent each clip-level audio signal. For each training iteration, we randomly select a continuous 50,176-D sub-vector from the clip-level vector as the clip-level audio representation. Then, a $17$-layer ResNet is employed, where a FC layer is attached at the top of it. 
    % The model training is achieved using the SGD optimizer with initial learning rate of 0.0002, weight decay of 0.0005 and momentum of 0.9 while the MSE is employed as the loss function. 

    \item \textbf{(v) CRNet \cite{crnet}:} The entire audio signal of each clip is converted to a 244,832-D vector using the librosa library \cite{librosa}, which is then used as the clip-level audio representation. The ResNet-34 model is employed to process this audio representation to extract the clip-level audio personality features, from which the CR-block and ETR are used to jointly predict five personality traits.
    % The entire audio model is trained using the SGD optimizer with initial learning rate of 0.0002, weight decay of 0.0005 and momentum of 0.9 and the Bell loss is employed as the loss function.  

     \item \textbf{(vi) VGGish \cite{suman2022multi}:} The raw audio clip is re-sampled and encoded as spectrogram features, which are framed into non-overlapping examples of 0.96s. A pre-trained VGGish CNN is then introduced to extract audio personality features from each spectrogram frame. The features of all frames are finally concatenated and fed to a MLP consisting of FC layers and a sigmoid layer to make a clip-level personality prediction.   
    
     % \item \textbf{(vi) VGGish \cite{suman2022multi}:} The raw audio clip is re-sampled to 16 kHz. The spectrogram of the audio is then created, and these features are framed into non-overlapping examples of 0.96s. A pre-trained VGGish CNN is then introduced to extract audio personality features from each spectrogram frame. The features of all frames are finally concatenated and fed to a MLP consisting of FC layers and a sigmoid layer to make a clip-level personality prediction. The Bell loss is employed as the loss function.
     % The entire audio model is trained using the SGD optimizer with initial learning rate of 0.001, weight decay of 0.0005 and a momentum of 0.9, and the Bell loss is employed as the loss function.

\end{itemize}

\noindent \textbf{Audio-visual models:} Based on the reproduced audio and visual systems, we also reproduce four audio-visual models that have been used for APR on the ChaLearn First Impression dataset as follows:
\begin{itemize}

    \item \textbf{(i) DAN-MFCC/Logfbank \cite{8066355}:} Following the same settings introduced in \cite{8066355}, we build an audio-visual model that is made up of the corresponding visual and the audio models (the visual model (i) and the audio model (ii)) described above. The final clip-level prediction is obtained by averaging clip-level audio prediction and clip-level video prediction.

    \item \textbf{(ii) Bi-modal CNN-LSTM \cite{bi_modal_lstm}:} The audio and visual clip of each subject is firstly divided into 6 segments, where each visual segment is processed by the visual model (iii) described above and each audio segment is processed by the audio model (iii) described above. Then, each pair of segment-level audio and visual latent features are concatenated as a 160-D vector. Finally, the clip-level personality prediction is obtained by feeding the audio-visual vectors of six segments to an LSTM model with 128 hidden units. Here, the LSTM contains a single hidden layer with 128 neurons. 
    % The entire model is trained using the SGD optimizer with initial learning rate of 0.05, weight decay of 0.0005 and momentum of 0.9, and MSE is employed as the loss function.

    \item \textbf{(iii) ResNet \cite{audiovisual_resnet}:} A two-steam ResNet model is employed, which consists of the visual model (iv) and the audio model (iv) described above. The audio feature (256-D) and the latent frame-level visual feature (256-D) are concatenated as a 512-D feature to a fully-connected layer to jointly predict the five personality traits. Specifically, the model randomly takes an audio and visual sample at each training iteration. 
    % The entire model is trained using the SGD optimizer with initial learning rate of 0.0002, weight decay of 0.0005 and momentum of 0.9, and MSE is employed as the loss function.

     \item \textbf{(iv) CRNet \cite{crnet}:} A three-stream ResNet-34 model is employed, which consists of two visual streams (the visual model (vi) described above), and an audio stream (the audio model (v) described above). Specifically, the two-stream visual model takes both the face image sequence and full frame sequence as the visual inputs. The latent features produced by audio and visual streams are combined as a single 512-D vector by element-wise sum.  
     % The entire model is trained using the SGD optimizer with initial learning rate of 0.002, weight decay of 0.005 and momentum of 0.9, where the Bell loss is employed as the loss function.  

     \item \textbf{(v) Amb-Fac-VGGish \cite{suman2022multi}:} A multi-modal system that contains a visual stream (the visual model (vii) described above) which takes both face and full frames as the input, and an VGGish-based audio stream (the audio model (vii) described above) which learns personality features from the audio spectrogram. 
     % The entire model is trained using the SGD optimizer with initial learning rate of 0.001, weight decay of 0.005 and momentum of 0.9, where the Bell loss is employed as the loss function.
    
\end{itemize}

\subsubsection{Widely-used static/spatio-temporal visual deep learning models} 
\label{subsec:widely-used models}

\setlength{\tabcolsep}{2pt}
\begin{table}[t!]
	\begin{center}
        \resizebox{1\linewidth}{!} {
		    \begin{tabular}{|l| c| c | c | c | c |}
			\toprule
                & Model                                  & Learning rate   & Weight decay & Momentum & Optimizer   \\
                \hline \hline
            \multirow{6}*{Audio} 
                & FFT \cite{interpre-aud}                & 0.01          & 0.0005             & 0.9   & SGD  \\
                & MFCC/Logfbank \cite{8066355}           & 0.00083       & 6.5                & 0.9   & SGD  \\
                & Bi-modal CNN-LSTM \cite{bi_modal_lstm} & 0.05          & 0.0005             & 0.9   & SGD  \\
                & ResNet \cite{audiovisual_resnet}       & 0.0002        & 0.0005             & 0.9   & SGD  \\
                & CRNet \cite{crnet}                     & 0.002         & 0.005              & 0.9   & SGD/Adam  \\
                & VGGish \cite{suman2022multi}           & 0.001         & 0.0005             & 0.9   & SGD  \\
                \hline
            \multirow{13.5}*{Video} 
                & DAN \cite{8066355}                     & 0.05          & 0.0005             & 0.8   & SGD  \\  
                & Bi-modal CNN-LSTM\cite{bi_modal_lstm}  & 0.05          & 0.0005             & 0.9   & SGD  \\ 
                & ResNet \cite{audiovisual_resnet}       & 0.05          & 0.0005             & 0.9   & SGD  \\ 
                & CRNet \cite{crnet}                     & 0.002         & 0.005              & 0.9   & SGD  \\ 
                & CAM-DAN$^{+}$ \cite{interpret_img}     & 0.05          & 0.0005             & 0.8   & SGD/Adam  \\
                & PersEmoN \cite{persemon}               & 0.001         & 0.0005             & 0.9   & SGD  \\
                & Amb-Fac \cite{suman2022multi}          & 0.01          & 0.0005             & 0.9   & SGD  \\
                & SENet \cite{senet}                     & 0.01          & 0.0005             & 0.8   & SGD  \\
                & HRNet \cite{hrnet}                     & 0.01          & 0.0005             & 0.9   & SGD  \\
                & VIT \cite{vit}                         & 0.001         & 0.0005             & 0.9   & SGD  \\
                & Swin-Transformer \cite{swin_trans}     & 0.01          & 0.0005             & 0.9   & SGD  \\
                & 3D-Resnet \cite{3d_resnet}             & 0.01          & 0.005              & 0.9   & SGD  \\
                & Slow-Fast \cite{slowfast}              & 0.002         & 0.0005             & 0.9   & SGD  \\
                & TPN \cite{tpn}                         & 0.001         & 0.0005             & 0.9   & SGD  \\
                & VAT \cite{vat}                         & 0.01          & 0.0005             & 0.9   & SGD  \\
                
                \hline
            \multirow{3}*{Audio-visual}
                % & DAN-MFCC \cite{8066355}                & 0.0008  & -0.0132  & -0.0222  & SGD  \\
                & Bi-modal CNN-LSTM \cite{bi_modal_lstm} &   0.05          & 0.0005             & 0.9  & SGD  \\
                & ResNet \cite{audiovisual_resnet}       & 0.001          & 0.0005             & 0.9   & SGD  \\
                & CRNet \cite{crnet}                     & 0.002          & 0.005             & 0.9   & SGD  \\
                % & Amb-Fac-VGGish \cite{suman2022multi}   & -0.0348 &  0.0468  & 0.0397   & SGD  \\
			\bottomrule
		\end{tabular}
        }
	\end{center}
	\caption{Training hyper-parameter settings for all benchmarked models}  
\label{tab:training details}
\end{table}
\setlength{\tabcolsep}{1.4pt}

% ------------------------------------------------------------------------------------------------------------------------------------------------

\noindent Since visual information is more informative for personality recognition (validated in Sec. \ref{subsec: all_results}), we additionally benchmark six standard visual deep learning models that have been widely used for static image or video analysis, including a VAT model \cite{vat} that has been previously applied to self-reported personality recognition \cite{palmero2021context}. Specifically, these include three models that take the static face/full image as the input and four models that infer personality from spatio-temporal visual data. The brief description of these models' unique characteristics are explained as follows:
\begin{itemize}

    \item \textbf{(i) SENet \cite{senet}:} SENet is a CNN model whose  ”Squeeze-and-Excitation” (SE) block adaptively calibrates channel-wise feature responses, i.e., it explicitly models inter-dependencies between features extracted from different convolution channels.

    \item \textbf{(ii) HRNet \cite{hrnet}:} HRNet is a hierarchical CNN model that has multi-level high-to-low resolution convolutions in parallel, which maintains high-resolution representations through the whole propagation process and produces strong high-resolution representations by repeatedly conducting fusion for representations extracted from parallel convolutions. 

    \item \item \textbf{(iii) VIT \cite{vit}:} VIT is a transformer-style model that computes relationships among pixels in various small patches of the input image based on the attention operation, where feature extraction for each patch is also controlled by several learnable embeddings. The VIT can effectively explore topological relationships between patches, making it able to capture global and wider range relations among pixels at the cost of a higher training complexity.

    \item \textbf{(iv) Swin-Transformer \cite{swin_trans}:}  Swin-Transformer is a hierarchical transformer model which applies a set of shifted windows to the input image, allowing the attention operations only to be conducted to non-overlapping local image regions, resulting in greater efficiency.

    \item \textbf{(v) 3DResNet \cite{3d_resnet}:} 3DResNet is a spatio-temporal CNN model which contains shortcut (residual) connections that allow a features to bypass one layer and move to the next layer during the propagation and back-propagation, where the 2D convolution operations in the standard ResNet \cite{he2016deep} are replaced with spatio-temporal convolution operations.

    \item \textbf{(vi) Slow-Fast \cite{slowfast}:} Slow-Fast is a spatio-temporal CNN model for video recognition. It contains a slow pathway to capture spatial semantics, which operates at a low frame rate in the temporal dimension of the input, as well as a fast pathway, operating at a high frame rate in the temporal dimension, aiming to capture dynamics.

    \item \textbf{(vii) TPN \cite{tpn}:} TPN is a spatio-temporal CNN that has hierarchical Temporal Pyramid structure, which can flexibly integrate multiple 2D or 3D backbone networks in a plug-and-play manner. As a result, it learns hierarchical spatial and temporal features from the input video, i.e., it can capture action instances at various tempos.

    \item \textbf{(viii) VAT \cite{vat}:} VAT is a spatio-temporal transformer that consists of 3D convolution operations to learn action features from each subject. Then, a set of head networks aggregating features from the spatio-temporal context to predict predict actions and regresses tighter bounding boxes.

\end{itemize}
The training settings (hyper-parameter settings) of all benchmarked models are also listed in Table \ref{tab:training details}.

\subsubsection{Clip-level representation generation models} 

\noindent To combine frame/segment-level personality predictions of an audio-visual clip for a clip-level personality trait prediction, this section also introduces two standard methods that have been employed in existing personality computing publications. In particular, besides the widely-used strategy which simply averages all frame/segment-level predictions at the clip-level prediction (used by \cite{interpret_img,8066355,audiovisual_resnet,persemon}), we also benchmark the spectral representation \cite{song2020spectral,song2018human} to summarize frame/segment-level predictions at the clip-level, which has been used by \cite{song2021self}. This is because the spectral representations not only effectively encode the temporal dependencies among all frame-level predictions of a clip, but also summarise time-series signal of an arbitrary length into a representation of a fixed size without any distortion.

\begin{itemize}

    \item \textbf{Averaging frame/segment-level predictions (AFP):} The clip-level personality prediction is obtained by averaging all frame/segment-level predictions.
    
    % \item \textbf{Averaging frame/segment-level features (AFF):} The clip-level personality feature is obtained by averaging all frame/segment-level features. Then, we train a \textcolor{red}{XXX} to predict clip-level personality traits from the produced \textcolor{red}{clip-level personality feature}.

    \item \textbf{Spectral representation of frame/segment-level predictions (SFP):}  We first encode a pair of spectral heatmaps (please see \cite{song2020spectral} for details) from the five-channel (corresponding to five traits) frame/segment-level personality trait prediction time-series. Then, we train a 1D-CNN to predict clip-level personality traits from the produced spectral heatmaps.

    % \item \textbf{Spectral representation of frame/segment-level features (SFF):} We first encode a pair of \textcolor{red}{spectral heatmap} from the multi-channel frame/segment-level personality features time-series. Then, we train a \textcolor{red}{XXX} to predict clip-level personality traits from the produced \textcolor{red}{spectral heatmaps}.

\end{itemize}

% %%% 2022/02/04| 代码内容
% For the benchmark, firstly, we reproduce the top three winning methods in ChaLearn2016 Looking at People challenge: deep bi-modal regression\cite{8066355}, bi-modal LSTM\cite{lstm}, audio-visual network\cite{2016}; two well-known approaches with multi-modal and multi-dataset fusion: CR-net\cite{crnet}, PersEmoN\cite{persemon}; and two interpreting attempts to illustrating model predictions in terms of visual and audio clues: Interpreting-Image\cite{interpret_img}, Interpreting-Audio\cite{interpre-aud}.
% Secondly, we explore some widely used 2D models on personality prediction to provide a general comparison with aforementioned specifically designed models, which are SENet\cite{senet}, HRNet\cite{hrnet} and Swin-Transformer\cite{swin_trans}. Thirdly, we carry out experiments on some widely used 3D models to provide comprehensive comparisons and references for further researches and those models are 3DResNet\cite{3d_resnet}, Slow-Fast\cite{slowfast}, TPN\cite{tpn} and VAT\cite{vat}. In general, there are fourteen handy models in this benchmark. At last we proposed some long-term fusion strategies based on the model prediction or output features to form a unified two stage prediction framework and those strategies are ATP\cite{}, STA\cite{}, SPV\cite{}, SPH\cite{}, SEG\cite{}, SPG\cite{}. 

%% 1. 我们程序的平台
%% 2. 预处理机制
%% 3. 模型设定
%% 4. 训练机制
%% 5. 评价机制

%%% 2022/02/04| 评价机制

\subsection{Evaluation datasets}
\label{subsec: dataset}

\noindent There are several publicly available audio-visual personality computing datasets for apparent personality recognition \cite{ponce2016chalearn,biel2010voices,sanchez2011nonverbal}, and self-reported personality recognition \cite{miranda2018amigos,palmero2021context,correa2018amigos} (please see \cite{junior2019first} for a survey of existing personality databases). Among them, the two most widely used publicly available audio-visual self-reported personality and apparent personality datasets: the UDIVA Dataset \cite{palmero2021context} and ChaLearn First Impression dataset \cite{ponce2016chalearn}, are employed to evaluate all models described in Sec. \ref{subsec: models}. Both datasets provide the Big-Five personality traits (i.e., \textbf{Extraversion} (\textbf{Ext}), \textbf{Agreeableness} (\textbf{Agr}), \textbf{Openness} (\textbf{Ope}), \textbf{Conscientiousness} (\textbf{Con}), and \textbf{Neuroticism} (\textbf{Neu})) as the label for each audio-visual clip.

\textbf{UDIVA dataset} was released in 2021. It records 188 dyadic interaction clips between $147$ voluntary participants, with total $90.5$h of recordings. Each clip contains two audio-visual files, where each records a single participant's behaviours. For each dyadic interaction session, participants were matched based on their availability, language and three variables: gender, age group and the relationship among interlocutors. In particular, all participants were matched to enforce a close-to-uniform distribution among all possible combinations between these variables. During the recordings, participants were asked to sit at 90  degrees to the conversational partner around a table, and under the dyadic interactions based on five tasks: Talk, 'Animal games', Lego building, “Ghost blitz” card game, and Gaze events, where each pair of participants were interacting using one of the three languages (i.e., English, Spanish or Catalan). The self-reported personality labels are obtained for each participant based on the following rules: (i) parents of children up to 8 years old completed the Children Behavior Questionnaire (CBQ) \cite{rothbart2001investigations}; participants from 9 to 15 years old completed the Early Adolescent Temperament Questionnaire (EATQ-R) \cite{ellis2001revision}; and the rest of the participants completed both the Big Five Inventory \cite{soto2017next} and the Honesty-Humility axis of the HEXACO personality inventory \cite{ashton2009hexaco}. However, due to the limited materials, it is neither possible to investigate the reliability of these labels, nor understand the real internal states of these participants. Also, behaviours in such short audio-visual clips may not be reliable to represent participants’ personality traits as personality traits are long-term status.

\textbf{ChaLearn First Impression dataset} was released in 2016. It contains talking-to-the-camera $10,000$ audio-visual clips that come from $2,764$ YouTube users, where each video lasts for about $15$ seconds with $30$ fps. Each video is labelled with the Big-Five personality traits that are annotated by external human annotators using the Amazon Mechanical Turk. The intensity of each trait was normalised to the range of $[0,1]$. The dataset provides official splits for training ($6,000$ videos), validation  ($2,000$ videos) and test  ($2,000$ videos). It is the largest audio-visual dataset openly available for research purposes, which has been annotated with Big-Five apparent personality traits. However, these annotations may suffer from unintentionally bias caused by pre-conception of annotators on each individual (e.g., gender and ethnicity). As a result, the classifier might model these bias patterns.

\section{Experimental Results}

\noindent In this section, we first present the results of all benchmarked models for both self-reported personality and apparent personality traits recognition in Sec. \ref{subsec: all_results}. Then, in Sec. \ref{subsec: ablation} we discuss the influences of different settings on personality recognition performances. Finally, we provide a systematic discussion of the benchmarking results in Sec. \ref{subsec:result-dis}

%%% 29/08/2022：所有的表方法引用

\subsection{Benchmarking personality computing models}
\label{subsec: all_results}

\noindent This section reports the self-reported and apparent personality recognition results achieved by all benchmarked models. According to Table \ref{tb:CCC-UDIVA} and Table \ref{tb:MSE-UDIVA}, all benchmarked models failed to accurately infer the self-reported personality traits from human audio/visual/audio-visual behaviours, with only three visual models (e.g., Interpret-img, SENet, and HRNet) achieving more than $0.15$ CCC performances. Standard audio feature extraction methods can barely extract self-reported personality-related information from non-verbal audio signals, as most of the benchmarked models achieved near-zero CCC values, while only the predictions of the VGGish achieved CCC $> 0.1$ with the ground-truth. Some audio models even degraded the corresponding visual systems under the audio-visual setting. The low CCC and high MSE results of almost all models on self-reported personality recognition indicate that it may difficult and unreliable for standard deep learning models to be used to directly infer human self-reported personality traits from external non-verbal behaviours.

In contrast, the relatively higher CCC and ACC results shown in Table \ref{tb:CCC-chalearn} and Table \ref{tb:ACC-chalearn} suggest that apparent personality traits can be better predicted from human non-verbal behaviours when using deep learning models, i.e., five visual models and two audio-visual models achieved CCC $> 0.5$. In particular, the predictions produced by HRNet and VAT from only facial behaviours have more than $0.6$ correlation (measured by CCC) with the ground-truth. In addition, we found that the CR-Net and VGGish models can also deep learn valuable apparent personality-related cues from non-verbal audio signals. The aforementioned results suggest that compared to self-reported personality, apparent personality is generally more feasible and easier to be directly predicted from human non-verbal behaviours. Since apparent personality is defined as the external human observer's perception of the target subject, these results can be explained by the fact that it is straight-forward for deep learning models to be used as an external observer and make judgments based on behaviours that can be directly observed. In contrast, self-reported personality represents the internal state/attribute of the target person, which is not straightforward to be directly inferred from external behaviours using standard deep learning models. The detailed discussions on applying ML models for inferring self-reported personality traits can be found in \cite{song2022learning,shao2021personality}.

We also conducted a 10-fold cross-validation to further compare the performances of the six best models (i.e., two audio models: CRNet and VGGish; two visual models: HRNet and VAT; and two audio-visual models: CRNet and Amb-Fac-VGGish), as they achieved top or runner-up performances in audio, visual and audio-visual personality recognition, respectively. The detailed results provided in the Supplementary Material show that the performance of all models on APR are still much better than their corresponding performances on SPR. However, we found that visual and audio-visual models-based cross-validation APR results are clearly worse than the results achieved for the pre-defined training/validation/test protocol. Since the `data hungry’ experiments provided in the Supplementary Material show that the APR performances keep increasing on the test set when more training samples are provided, we assume that the pre-defined test set may contain simpler samples by chance.

% cross-validation might be a useful protocol for a more holistic evaluation as compared to results that depend on pre-defined dataset splits.

\setlength{\tabcolsep}{2pt}
\begin{table}[t!]
	\begin{center}
        \resizebox{1\linewidth}{!} {
		    \begin{tabular}{|l| c| c | c | c | c | c | l|}
			\toprule
                & Traits & Open  & Consc  & Extrav & Agree & Neuro & Avg.   \\
                \hline \hline
            \multirow{6}*{Audio} 
                & FFT \cite{interpre-aud}            & 0.0000  &  0.0000 & 0.0000  & 0.0000  & 0.0000 & 0.0000  \\
                & MFCC/Logfbank \cite{8066355}       & 0.0002  & -0.0002 & 0.0002  & 0.0001  & 0.0001 & 0.0001  \\
                & Bi-modal CNN-LSTM \cite{bi_modal_lstm}  & 0.0000  & -0.0001 & 0.0000  & 0.0000  & 0.0000 & 0.0000  \\
                & ResNet \cite{audiovisual_resnet}   & -0.0459 & \underline{0.1045}  & -0.0416 & \underline{0.0429}  & 0.0015 & 0.0123  \\
                & CRNet \cite{crnet}                & \underline{0.0005}  & 0.0290  & \underline{0.0201}  & 0.0335  & 0.0267 & \underline{0.0220}  \\
                & VGGish \cite{suman2022multi} & \textbf{0.0688}  &  \textbf{0.1882} & \textbf{0.1310}  & \textbf{0.1069}  & \textbf{0.0412} & \textbf{0.1072}  \\
                \hline
            \multirow{13.5}*{Video} 
                & DAN \cite{8066355}                    & 0.0009 &  0.0087 &  0.0061 &  0.0035 &  -0.0016 &  0.0036   \\ % frame 
                & Bi-modal CNN-LSTM\cite{bi_modal_lstm} & -0.0001 & 0.0000 &  0.0000 &  0.0001 &  0.0000 &  0.0000   \\ % face
                & ResNet \cite{audiovisual_resnet}      & 0.0648 &  0.2424 &  0.0349 &  0.0305 &  -0.0009 &  0.0743  \\ % frame
                & CRNet \cite{crnet}                    & -0.0206 &  0.0879 &  0.012 &  0.1454 &  0.0645 &  0.0578   \\ % frame/face
                & CAM-DAN$^{+}$ \cite{interpret_img}    &  0.0336  &  \textbf{0.3359} &  0.0270 &  \textbf{0.2014} &  \underline{0.2709} &  \underline{0.1738}  \\
                & PersEmoN \cite{persemon}              & -0.0095  &  0.0133 &  0.0045 &  0.0085 &  0.0058 &  0.0045  \\
                & Amb-Fac \cite{suman2022multi}         & 0.0532   &  0.1941 &  0.0442 &  0.0453 &  0.0204 &  0.0714  \\
                & SENet \cite{senet}                    &  \underline{0.1678}  &0.2776 &  \textbf{0.0538} &  0.0299 &  \textbf{0.3093} &  0.1510  \\
                & HRNet \cite{hrnet}                    &  \textbf{0.2175}  &  \underline{0.2998} & -0.0039 &  \underline{0.1680} &  0.1945 &  \textbf{0.1752}  \\
                & VIT \cite{vit}      &  -0.0184  &  0.1532 & -0.0002 & 0.0645  & 0.0607  & 0.0520 \\
                & Swin-Transformer \cite{swin_trans}    & -0.0273  &  0.0470 &  0.0361 & -0.0142 &  0.0860 &  0.0256  \\
                & 3D-Resnet \cite{3d_resnet}            & -0.0478  &  0.0102 &  \underline{0.0478} &  0.0499 & -0.0240 &  0.0072  \\
                & Slow-Fast \cite{slowfast}             & -0.0102  &  0.0076 &  0.0010 & -0.0063 &  0.0161 &  0.0016  \\
                & TPN \cite{tpn}                        &  0.0448  &  0.0348 &  0.0287 & -0.0177 & -0.0281 &  0.0125  \\
                & VAT \cite{vat}                        & -0.0139  & -0.0016 &  0.0016 & -0.0013 & -0.0006 & -0.0031  \\
                
                \hline
            \multirow{5.5}*{Aud-vis}
                & DAN-MFCC \cite{8066355}              &  \underline{0.0008} &  0.1154 & -0.0132 & -0.0222 &  \textbf{0.1219} &  \underline{0.0405} \\
                & Bi-modal CNN-LSTM \cite{bi_modal_lstm}           &  0.0001 & -0.0001 &  0.0001 & -0.0001 &  0.0005 &  0.0001 \\
                & ResNet \cite{audiovisual_resnet} & -0.0530 &  \underline{0.1290} &  0.0230 &  0.0310 &  0.0002 &  0.0260 \\
                & CRNet \cite{crnet}               & \textbf{0.0998} &  \textbf{0.1780} &  \textbf{0.1158} &  \textbf{0.2168} &  \underline{0.0449} &  \textbf{0.1311} \\
                & Amb-Fac-VGGish \cite{suman2022multi}       & -0.0348 &  0.0468 &  \underline{0.0302} &  \underline{0.0397} &  0.0041 &  0.0171  \\
			\bottomrule
		\end{tabular}
        }
	\end{center}
	\caption{The CCC results achieved for the self-reported personality recognition on the UDIVA dataset. The reported results are obtained by averaging the results achieved for four sessions. For different modality groups, audio, video, audio-visual, the values in bold denote the highest ones in the columns and the values with underline denote the second highest ones.}  
\label{tb:CCC-UDIVA}
\end{table}
\setlength{\tabcolsep}{1.4pt}

\setlength{\tabcolsep}{2pt}
\begin{table}[t!]
	\begin{center}
        \resizebox{1\linewidth}{!} {
		    \begin{tabular}{|l| c| c | c | c | c | c | l|}
			\toprule
                & Traits         & Open    & Consc   & Extrav  & Agree   & Neuro   & Avg.   \\
                \hline \hline
            \multirow{6}*{Audio} 
                & FFT \cite{interpre-aud}           & 1.0122  &  0.7114 &  1.5056 &  0.9342 &  1.2851 &  1.0897  \\
                & MFCC/logfbank \cite{8066355}      & 0.9489  &  1.0256 &  2.0083 &  1.2377 &  1.3526 &  1.3144  \\
                & Bi-modal CNN-LSTM \cite{bi_modal_lstm} & 0.9544  &  0.7717 &  \underline{1.2788} &  \underline{0.8676} &  \underline{1.1965} &  \underline{1.0138}  \\
            & ResNet \cite{audiovisual_resnet}  & 0.9540  &  0.6985 &  \textbf{1.2325} &  \textbf{0.8611} &  \textbf{1.0744} &  \textbf{0.9641}  \\
                & CRNet \cite{crnet}               & \textbf{0.9145}  &  \underline{0.6978} &  1.4821 &  0.9623 &  1.3858 &  1.0885  \\
                & VGGish \cite{suman2022multi} & \underline{0.9312}  &  \textbf{0.6576} &  1.3345 &  0.9002 &  1.2227 &  1.0092  \\
                \hline
            \multirow{13.5}*{Video}
                & DAN \cite{8066355}                   & 1.0841 &  1.1174 & 2.0979  & 1.4846 & 1.2400 & 1.4048   \\ % frame 
                & Bi-modal CNN-LSTM\cite{bi_modal_lstm}& 0.8947 &  0.6690 & 1.3459  & 0.9269 & 1.1518 & 0.9977  \\ % face
                & ResNet \cite{audiovisual_resnet}     & 0.9637 &  0.6709 & 1.5770  & 1.0364 & 1.1526 & 1.0801  \\ % frame
                & CRNet  \cite{crnet}                  & 1.0985 &  0.6877 &  1.3079 &  0.909 &  1.4140 &  1.0834   \\ % frame/face
                & CAM-DAN$^{+}$ \cite{interpret_img}   & 1.0223  & \textbf{0.5471}  & 1.4800  & \textbf{0.8446}  & 1.1493  & 1.0087  \\
                & PersEmoN \cite{persemon}             & 0.9034  & 0.6985  & 1.4097  & 0.9446  & \underline{1.1469}  & 1.0206  \\
                & Amb-Fac \cite{suman2022multi}        & \textbf{0.8529}  & \underline{0.6394}  & 1.4212  & 0.9334  & \textbf{1.1410}  & \underline{0.9976}  \\
                & SENet \cite{senet}                   & 1.1398  & 0.6571  & 1.5526  & 1.1313  & 1.2733  & 1.1508  \\
                & HRNet \cite{hrnet}                   & 1.1224  & 0.8163  & 1.8893  & 1.0362  & 1.5447  & 1.2818  \\
                & VIT \cite{vit}     & 1.1460  & 0.8931  & 2.1999  & 1.2980   & 1.2843  & 1.3641  \\
                & Swin-transformer \cite{swin_trans}   & 1.0398  & 0.7119  & 1.3394  & 1.0186  & 1.2767  & 1.0773  \\
                & 3D-Resnet \cite{3d_resnet}           & 1.3016  & 0.7051  & \underline{1.2360}  & 0.8741  & 1.3855  & 1.1004  \\
                & Slow-Fast \cite{slowfast}            & 1.0652  & 0.9862  & 2.0394  & 1.2631  & 1.2956  & 1.3299  \\
                & TPN \cite{tpn}                       & 0.9259  & 0.7596  & 1.2524  & 0.9458  & 1.4147  & 1.0597  \\
                & VAT \cite{vat}                       & \underline{0.8554}  & 0.7076  & \textbf{1.1398}  & \underline{0.8520}  & 1.2286  & \textbf{0.9566}  \\
                \hline
                
            \multirow{5.5}*{Aud-vis}
                & DAN-MFCC \cite{8066355}              & 1.1657  &  0.7358 &  2.0523 &  1.1006 &  \textbf{1.0745} &  1.2258  \\
                & Bi-modal CNN-LSTM \cite{bi_modal_lstm}           & \textbf{0.9281}  &  \underline{0.7315} &  \underline{1.2944} & \textbf{0.8625} &  1.3152 &  \textbf{1.0263}  \\
                & ResNet \cite{audiovisual_resnet} & 1.1417  &  0.7787 &  1.6838 &  1.0732 &  \underline{1.1512} &  1.1657  \\
                & CRNet \cite{crnet}                           & 1.2834  &  0.8644 &  \textbf{1.2876} &  1.0442 &  1.6987 &  1.2357  \\
                & Amb-Fac-VGGish \cite{suman2022multi}       & \underline{0.9391}  &  \textbf{0.6675} &  1.4956 &  \underline{0.9588} &  1.1590 &  \underline{1.0440}  \\
			\bottomrule
		\end{tabular}
        }
	\end{center}
	\caption{The MSE results achieved for the self-reported personality recognition on the UDIVA dataset. The reported results are obtained by averaging the results achieved for four sessions. For different modality groups, audio, video, audio-visual, the values in bold denote the smallest ones in the columns and the values with underline denote the second smallest ones.}  
\label{tb:MSE-UDIVA}
\end{table}
\setlength{\tabcolsep}{1.4pt}

%%% 2022/06/05 完善表格，引用都要加上，每个方法的名称首字母大写。同时，复现方法的visual部分结果也要加在表中，请根据Sec. 3.2.1的内容完成所有表格。
\setlength{\tabcolsep}{2pt}
\begin{table}[t!]
	\begin{center}
    \resizebox{1\linewidth}{!} {
		\begin{tabular}{|l| c| c | c | c | c | c | l|}
			\toprule
                & Traits & Open  & Consc  & Extrav & Agree & Neuro & Avg.   \\
            \hline \hline

            \multirow{6}*{Audio} 
                & FFT \cite{interpre-aud}                & 0.0002  & -0.0001 & -0.0002 & -0.0002 & -0.0003  & -0.0002 \\
                & MFCC/logfbank \cite{8066355}           & 0.1968  & 0.1497  & 0.1738  & 0.1295  & 0.1780   & 0.1655 \\ 
                & Bi-modal CNN-LSTM \cite{bi_modal_lstm} & -0.0004 & -0.0005 & 0.0004  & -0.0005 & -0.0008  & 0.0004 \\ 
                & ResNet \cite{audiovisual_resnet}       & 0.1293  & 0.0830  & 0.0458  & 0.1101  & 0.1548   & 0.1046 \\ 
                & CRNet \cite{crnet}                     & \underline{0.4122}  & \underline{0.3406}  & \underline{0.3846}  & \underline{0.2857}  & \underline{0.4306}   & \underline{0.3707} \\
                & VGGish \cite{suman2022multi}           & \textbf{0.4516}  & \textbf{0.4493}  & \textbf{0.4429}  & \textbf{0.3127}  & \textbf{0.4500}   & \textbf{0.4213} \\
            \hline

            \multirow{13.5}*{Visual}
                & DAN \cite{8066355}                    & 0.5693 & 0.6254 & 0.6070 & 0.4855 & 0.6025 & 0.5779  \\ % frame 
                & Bi-modal CNN-LSTM\cite{bi_modal_lstm} & 0.0000 & 0.0000 & 0.0000 & 0.0000 & 0.0000 & 0.0000  \\ % face
                & ResNet \cite{audiovisual_resnet}      & 0.1561 & 0.1902 & 0.1355 & 0.0838 & 0.1373 & 0.1406  \\ % frame
                & CRNet  \cite{crnet}                   & 0.3748 & 0.3646 & 0.3987 & 0.2390 & 0.3226 & 0.3399   \\ % frame/face
                & CAM-DAN$^{+}$ \cite{interpret_img}    & 0.5882 & 0.6550 & 0.6326 & 0.5003 & 0.6199 & 0.5992  \\ % frame
                & PersEmoN \cite{persemon}              & 0.2067 & 0.2441 & 0.2675 & 0.1369 & 0.1768 & 0.2064  \\ % frame/face from other dataset
                & Amb-Fac \cite{suman2022multi}         & 0.5858 & 0.6750 & 0.5997 & 0.4971 & 0.5765 & 0.5868  \\ % frame | frame shows better result
                & SENet \cite{senet}                    & 0.5300 & 0.5580 & 0.5815 & 0.4493 & 0.5708 & 0.5379  \\ % face
                & HRNet \cite{hrnet}                    & \underline{0.5923} & \textbf{0.6912} & \underline{0.6436} & \underline{0.5195} & \underline{0.6273} & \underline{0.6148}  \\ % face
                & VIT\cite{vit}                           & 0.0184 & 0.0817 & 0.0247 & 0.0175 & 0.0318 & 0.0348  \\
                & Swin-transformer \cite{swin_trans}    & 0.2223 & 0.2426 & 0.2531 & 0.1224 & 0.1942 & 0.2069  \\ % face
                & 3D-Resnet \cite{3d_resnet}            & 0.3248 & 0.3601 & 0.3601 & 0.2120 & 0.3352 & 0.3185  \\ % face
                & Slow-Fast \cite{slowfast}             & 0.0256 & 0.0320 & 0.0185 & 0.0105 & 0.0184 & 0.0210  \\ % face   
                & TPN \cite{tpn}                        & 0.4427 & 0.4767 & 0.4998 & 0.3230 & 0.4675 & 0.4420  \\ % face
                & VAT \cite{vat}                        & \textbf{0.6216} & \underline{0.6753} & \textbf{0.6836} & \textbf{0.5228} & \textbf{0.6456} & \textbf{0.6298}  \\ % face
            \hline
            
            \multirow{5.5}*{Aud-vis}
                & DAN-MFCC \cite{8066355}                     & 0.4341  & 0.4645  & 0.4553  & 0.3519  & 0.4588  & 0.4329  \\ % frame      with audio data
                & Bi-modal CNN-LSTM \cite{bi_modal_lstm} & 0.0000  & 0.0000  & 0.0000  & 0.0000  & 0.0000  & 0.0000  \\ % face       with audio data
                & ResNet \cite{audiovisual_resnet}       & 0.4150  & 0.3671  & 0.3889  & 0.2679  & 0.4181  & 0.3714  \\ % frame      with audio data
                & CRNet \cite{crnet}                     & \underline{0.5193}  & \underline{0.5106}  & \underline{0.5024}  & \underline{0.4026}  & \underline{0.5119}  & \underline{0.4894}  \\ % frame/face with audio data
                & Amb-Fac-VGGish \cite{suman2022multi}   & \textbf{0.5618}  & \textbf{0.6421}  & \textbf{0.5921}  & \textbf{0.4620}  & \textbf{0.5734}  & \textbf{0.5663}  \\ % frame/face with audio data
			\bottomrule
		\end{tabular}
        }
	\end{center}
	\caption{The CCC results achieved for the apparent personality recognition on the ChaLearn First Impression dataset.For different modality groups, audio, video, audio-visual, the values in bold denote the highest ones in the columns and the values with underline denote the second highest ones.}  
\label{tb:CCC-chalearn}
\end{table}
\setlength{\tabcolsep}{1.4pt}

\setlength{\tabcolsep}{2pt}
\begin{table}[t!]
	\begin{center}
    \resizebox{1\linewidth}{!} {
		\begin{tabular}{|l| c| c | c | c | c | c | l |}
			\toprule
                & Traits & Open  & Consc  & Extrav & Agree & Neuro & Avg.   \\
            \hline \hline
            \multirow{6}*{Audio} 
                & FFT\cite{interpre-aud}                & 0.8303  & 0.8174  & 0.8502  & 0.8309  & 0.8331  & 0.8324  \\
                & MFCC/logfbank \cite{8066355}          & 0.8891  & 0.8790  & 0.8835  & 0.8967  & 0.8802  & 0.8857  \\ 
                & Bi-modal CNN-LST \cite{bi_modal_lstm} & 0.8835  & 0.8747  & 0.8785  & 0.8937  & 0.8774  & 0.8816  \\ 
                & ResNet \cite{audiovisual_resnet}      & 0.8822  & 0.8780  & 0.8782  & 0.8958  & 0.8820  & 0.8832  \\ 
                & CRNet \cite{crnet}                    & \underline{0.9001}  & \underline{0.8895}  & \underline{0.8951}  & \underline{0.9022}  & \textbf{0.8964}  & \underline{0.8967}  \\ 
                & Vggish-feat \cite{suman2022multi}     & \textbf{0.9010}  & \textbf{0.8962}  & \textbf{0.8959}  & \textbf{0.9028}  & \underline{0.8953}  & \textbf{0.8982}  \\
            \hline
            \multirow{13.5}*{Visual}
                & DAN \cite{8066355}                    & 0.9098 & 0.9106 & 0.9096 & 0.9102 & 0.9061 & 0.9093  \\ % frame
                & Bi-modal CNN-LSTM\cite{bi_modal_lstm} & 0.8832 & 0.8742 & 0.8778 & 0.8933 & 0.8770 & 0.8811  \\ % face
                & ResNet \cite{audiovisual_resnet}      & 0.8896 & 0.8835 & 0.8837 & 0.8968 & 0.8830 & 0.8873   \\ % frame
                & CRNet  \cite{crnet}                   & 0.8987 & 0.8932 & 0.8952 & 0.9018 & 0.8908 & 0.8960   \\ % frame/face
                & CAM-DAN$^{+}$ \cite{interpret_img}    & \textbf{0.9115} & 0.9139 & \underline{0.9126} & \textbf{0.9118} & \underline{0.9089} & \textbf{0.9118}  \\ % frame
                & PersEmoN \cite{persemon}              & 0.8934 & 0.8893 & 0.8913 & 0.8994 & 0.8866 & 0.8920  \\ % frame/face from other dataset
                & Amb-Fac \cite{suman2022multi}         & \underline{0.9101} & \underline{0.9141} & 0.9082 & 0.9095 & 0.9038 & 0.9091  \\ % frame 
                & SENet \cite{senet}                    & 0.9076 & 0.906  & 0.908  & 0.9097 & 0.9061 & 0.9075  \\ % face
                & HRNet \cite{hrnet}                    & \underline{0.9101} & \textbf{0.9154} & 0.9111 & \underline{0.9113} & 0.9084 & \underline{0.9113}  \\ % face
                & VIT \cite{vit}    & 0.8832 & 0.8760 & 0.8778 & 0.8934 & 0.8778 & 0.8817  \\
                & Swin-transformer \cite{swin_trans}    & 0.8937 & 0.8870 & 0.8893 & 0.8983 & 0.8860 & 0.8909  \\ % face
                & 3D-Resnet \cite{3d_resnet}            & 0.8964 & 0.8921 & 0.8933 & 0.9008 & 0.8915 & 0.8948  \\ % face
                & Slow-Fast \cite{slowfast}             & 0.8780 & 0.8604 & 0.8443 & 0.8809 & 0.8613 & 0.8650  \\ % face    
                & TPN \cite{tpn}                        & 0.9025 & 0.8963 & 0.9019 & 0.9013 & 0.8992 & 0.9003  \\ % face
                & VAT \cite{vat}                        & \textbf{0.9115} & 0.9123 & \textbf{0.9153} & 0.9099 & \textbf{0.9098} &\textbf{0.9118}  \\ % face
            \hline
            
            \multirow{5.5}*{Aud-vis}
                & DAN-MFCC \cite{8066355}                     & 0.9049 & \underline{0.9020} & \underline{0.9024} & \underline{0.9081} & 0.9015 & 0.9038 \\ % frame       with audio data(separated)
                & Bi-modal CNN-LSTM \cite{bi_modal_lstm} & 0.8833 & 0.8744 & 0.8779 & 0.8935 & 0.8773 & 0.8813 \\ % face        with audio data
                & ResNet \cite{audiovisual_resnet}       & 0.8996 & 0.8918 & 0.8945 & 0.9015 & 0.8948 & 0.8964 \\ % frame       with audio data
                & CRNet \cite{crnet}                     & \underline{0.9075} & 0.9019 & 0.9017 & 0.9055 & \underline{0.9034} & \underline{0.9040} \\ % frame/face  with audio data
                & Amb-Fac-VGGish \cite{suman2022multi}   & \textbf{0.9127} & \textbf{0.9169} & \textbf{0.9117} & \textbf{0.9133} & \textbf{0.9088} & \textbf{0.9127} \\ % frame/face  with audio data
			\bottomrule
		\end{tabular}
        }
	\end{center}
	\caption{The ACC results achieved for the apparent personality recognition on the ChaLearn First Impression dataset.For different modality groups, audio, video, audio-visual, the values in bold denote the highest ones in the columns and the values with underline denote the second highest ones.}  
 \label{tb:ACC-chalearn}
\end{table}
\setlength{\tabcolsep}{1.4pt}

%2022/08/29：

% 1. audio/visual/audio-visual: 柱状图
% 2. visual: 两个柱状图：标准方法下  True personality 和 Impression， frame-level methods vs segment-level methods vs video-level methods 的方法均值对比， 都用CCC
% 3. Reproduced 的结果和原始结果对比表

\subsection{Ablation studies}
\label{subsec: ablation}

\noindent In this section, we evaluate the influences of different pre-processing, post-processing and model settings on personality recognition performances. The statistical significance analyses for different settings are additionally provided in the Supplementary Material, where we found that the differences in most settings brought much more impact on APR performances over SPR performances. This can be explained by the fact that the benchmarked models can hardly infer self-reported personality traits from human non-verbal behaviours, and thus no matter what the settings are, they always achieved very unreliable predictions.

%%% 29/08/2022：添加文献
\subsubsection{Full frames VS. Face regions} 

\begin{figure}	
    \centering
	\subfigure[CCC results achieved for the UDIVA dataset.]{\label{subfig:ccc_per_udiva_face_frame}
		\includegraphics[width=8.8cm]{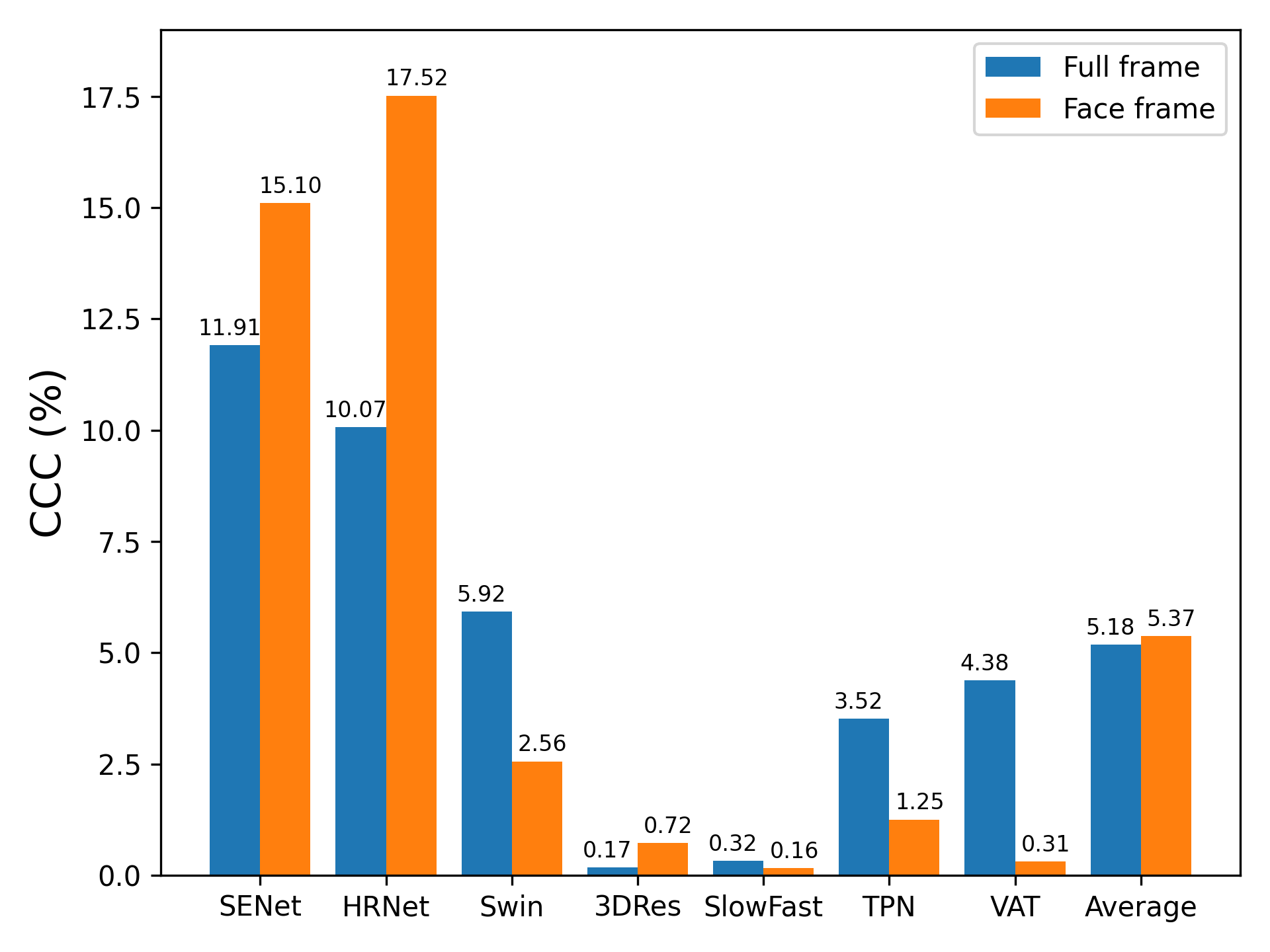}}
	\subfigure[MSE results achieved for the UDIVA dataset.]{\label{subfig:rmse_per_udiva_face_frame}
		\includegraphics[width=8.8cm]{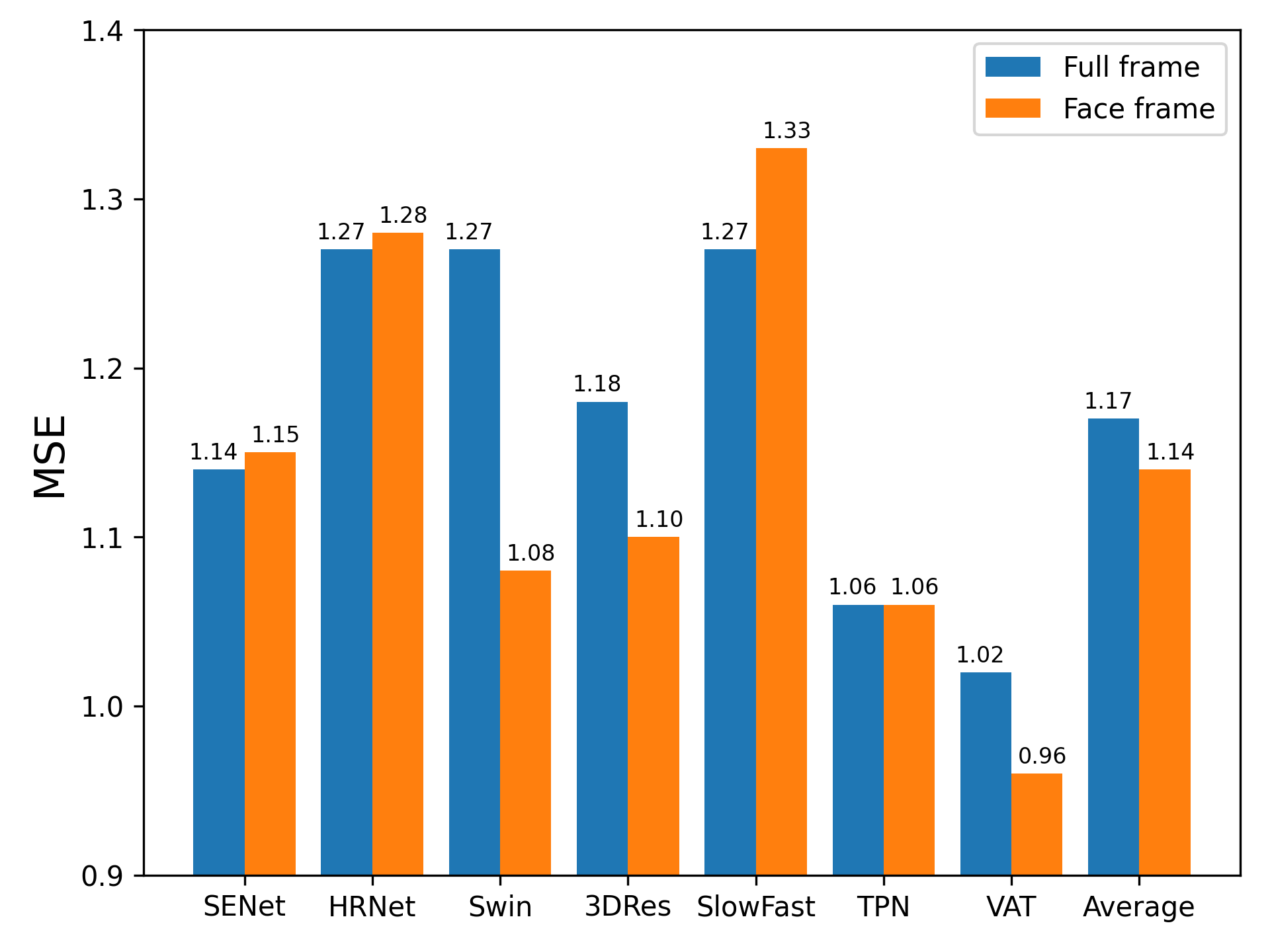}}
	\caption{Comparison between the self-reported personality recognition results achieved by different visual models using either aligned faces or original full frames.} 
    \label{fig:UDIVA_face_frame}
\end{figure}

\begin{figure}
	\centering
	\subfigure[CCC results achieved for the ChaLearn First Impression dataset.]{\label{subfig:ccc_per_impression_face_frame}
		\includegraphics[width=8.8cm]{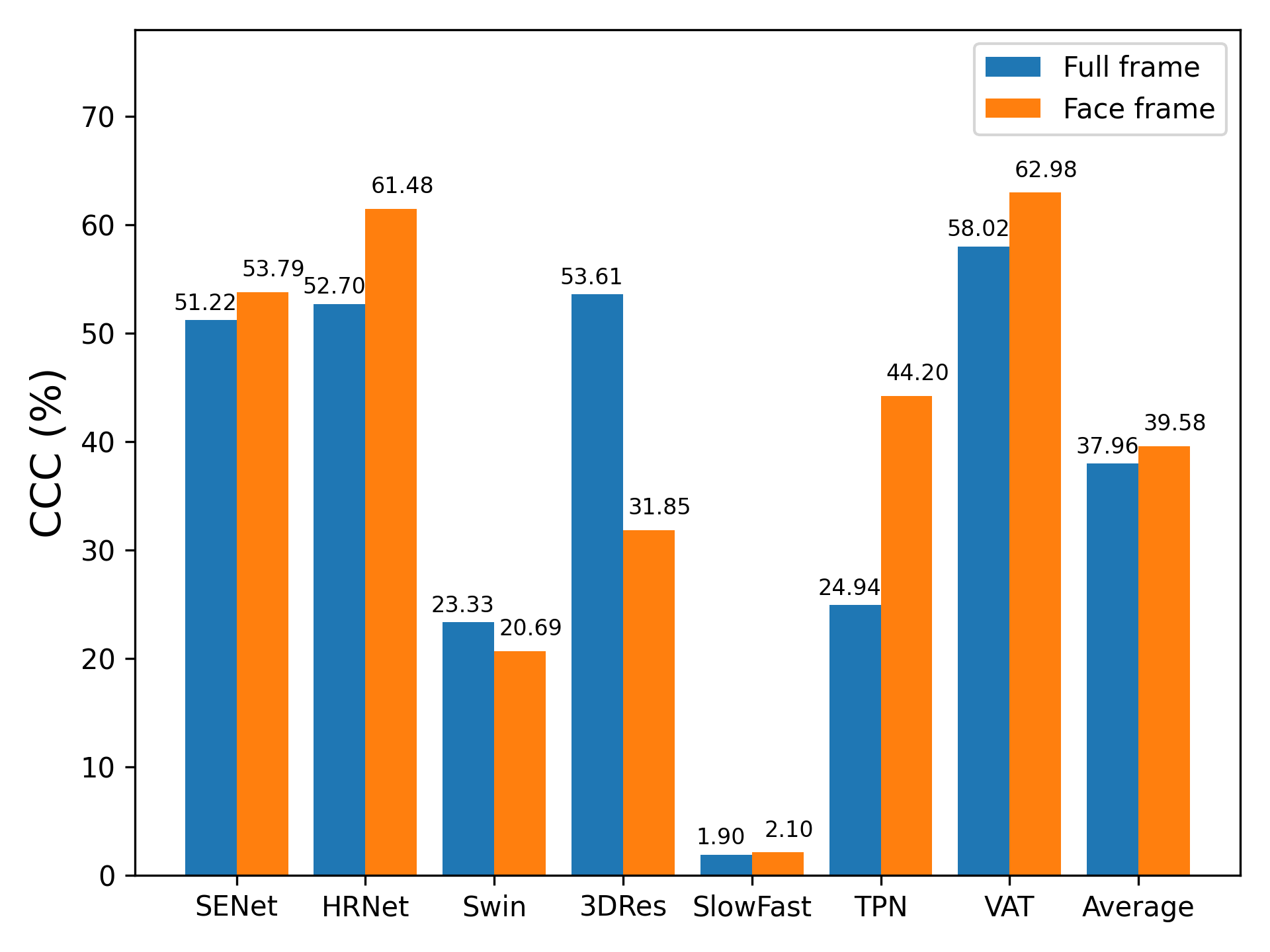}}
	\subfigure[ACC results achieved for the ChaLearn First Impression dataset.]{\label{subfig:acc_per_impression_face_frame}
		\includegraphics[width=8.8cm]{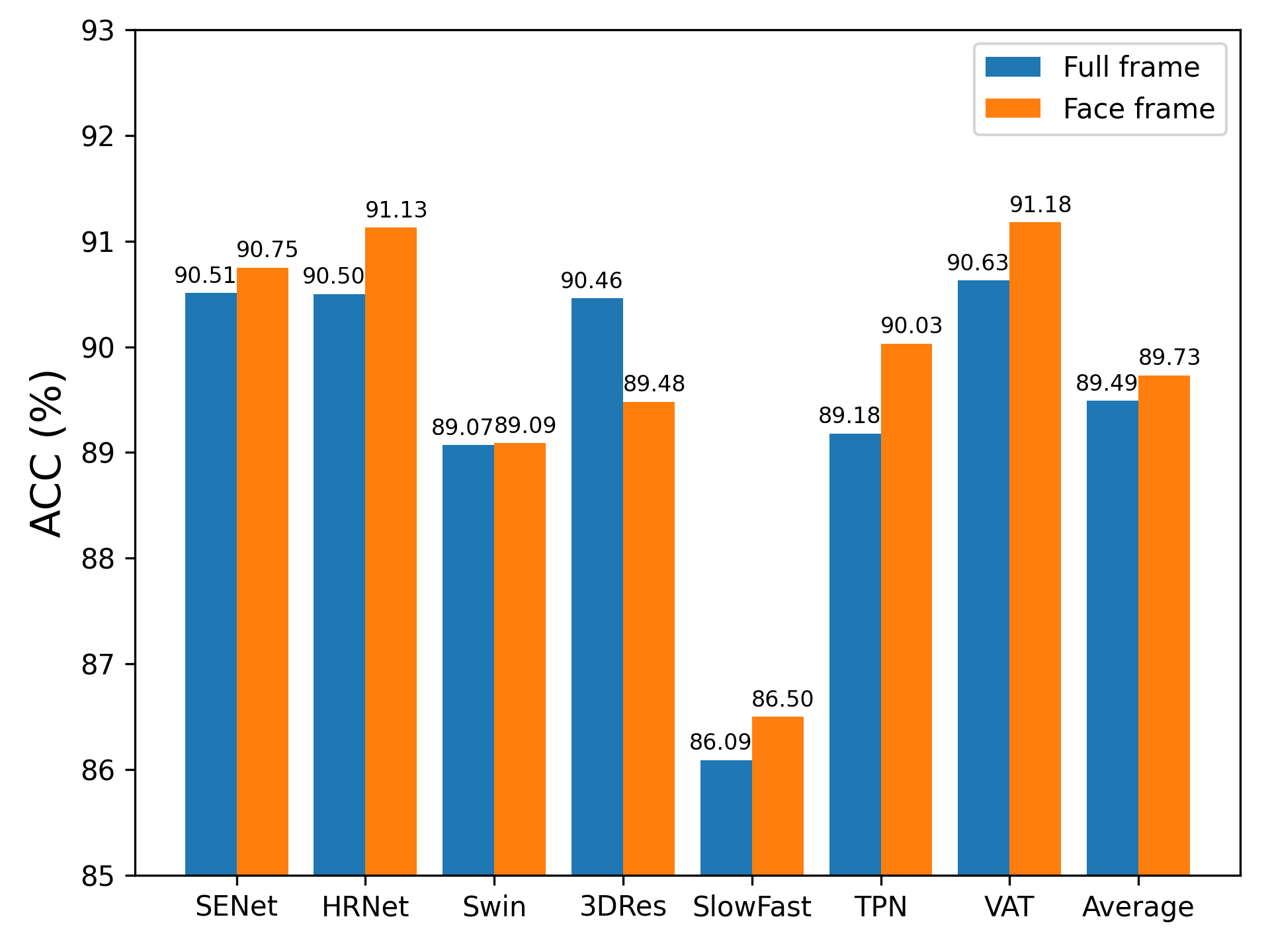}}
	\caption{Comparison between the apparent personality recognition CCC results achieved by different visual models using either aligned faces or original full frames.} 
    \label{fig:Chalearn_face_frame}
\end{figure}

\noindent We compare the results achieved by feeding aligned face regions and full frames (containing both the face and backgrounds) to standard visual deep learning models in Fig. \ref{fig:UDIVA_face_frame} and Fig. \ref{fig:Chalearn_face_frame}, where the standard visual deep learning models generally provide more reliable self-reported and apparent personality predictions by using face regions, despite the fact that neither full frames nor face regions provide reliable self-reported personality predictions. In contrast, apparent personality predictions generated by all visual models using both face regions and full frames have CCC $ > 0.37$ on average with the ground-truth, where face region-based predictions still have $1.62\%$ CCC advantage over full frame-based predictions. However, it can be seen that using face regions and full frames does not cause a large difference for most models, where only HRNet is sensitive to this variable on both tasks. These results suggest that although some previous studies claimed that backgrounds and spatial contextual information can provide informative cues for personality recognition, they may also contain personality-unrelated noise that can negatively impact the personality recognition. This indicates that personality-related cues contained in background and spatial contextual information can not always compensate their negative impact for personality recognition. Interestingly, some of the benchmarked models (e.g., 3DResNet) achieved better performances with the full frame setting. As shown in Fig. \ref{fig:UDIVA_face_frame} and Fig. \ref{fig:Chalearn_face_frame}, we hypothesise that the ability of 3DResNet to extract personality-related cues from facial behaviours may be limited, as evidenced by the relatively poor performance of its face-based system compared to other face-based systems (i.e., other benchmarked models). In contrast, the full frame-based system of 3DResNet yielded the second best results among all full frame-based systems, indicating its potential to capture valuable personality-related cues from head/body movements (as the head/body movements are contained in full frame sequences) or even contextual factors such as the background setting (e.g., office, bedroom) contained in the full frames.

%%% 29/08/2022：添加图
\subsubsection{Spatial VS. Spatio-temporal visual models} 

\begin{figure}
	\centering
	\subfigure[CCC results achieved for the UDIVA dataset.]{\label{}
		\includegraphics[width=8.8cm]{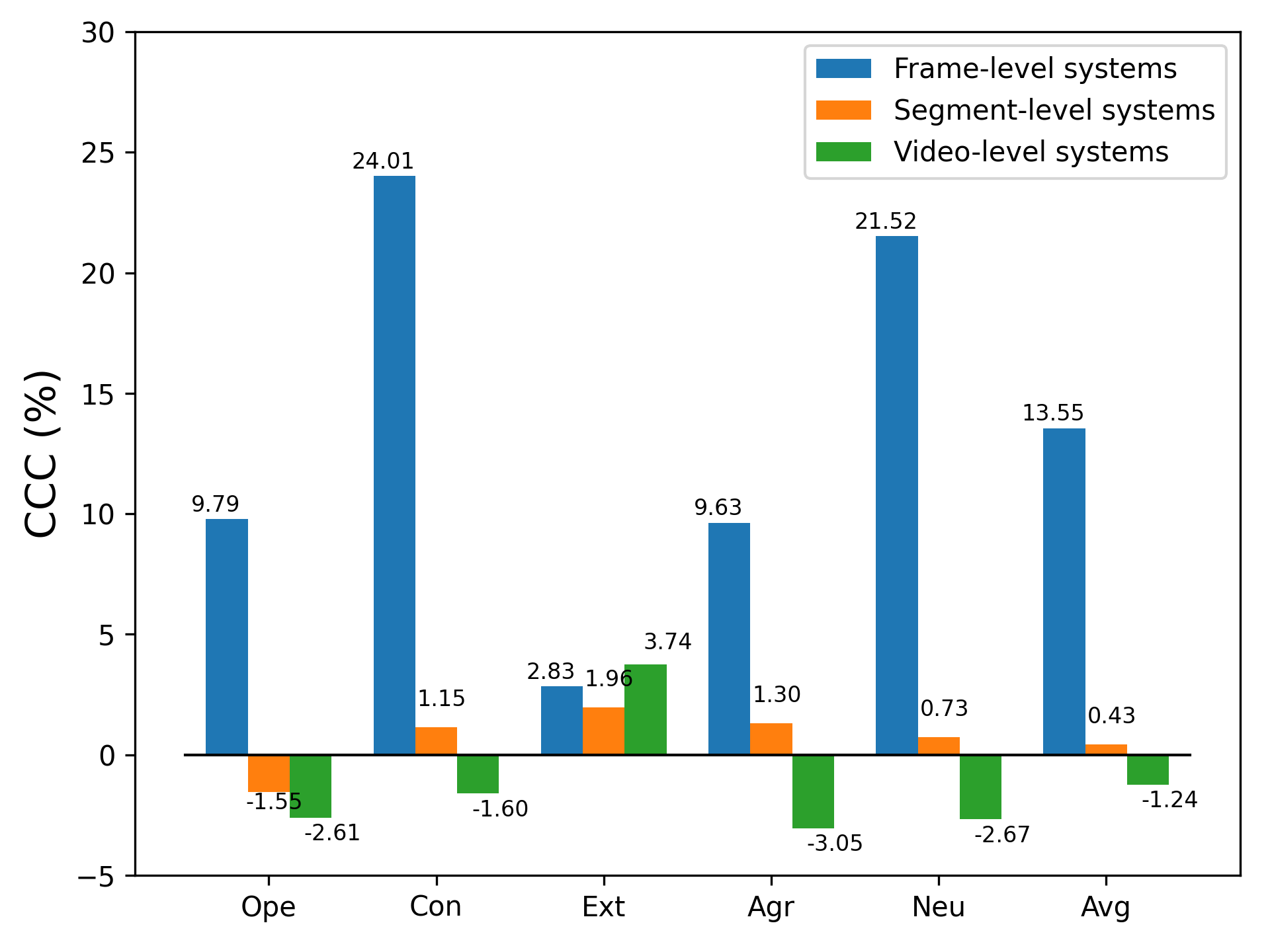}}
	\subfigure[CCC results achieved for the ChaLearn First Impression dataset.]{\label{}
		\includegraphics[width=8.8cm]{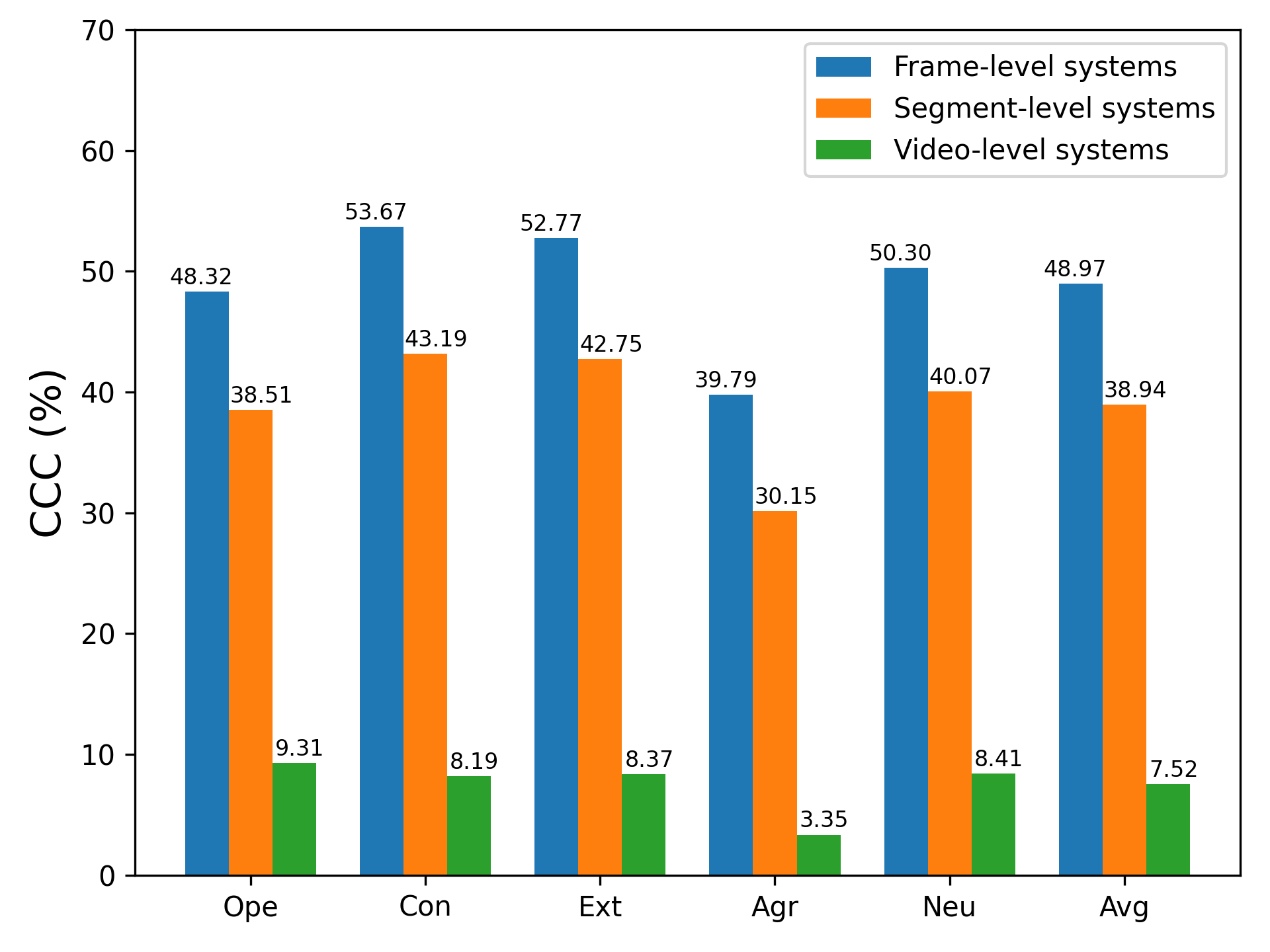}}
	\caption{Comparison between average results achieved by frame-level models and spatio-temporal models, where both short segment-level system and video-level system are evaluated for spatio-temporal models.} 
    \label{fig:frame-segment-video}
\end{figure}

\noindent Fig. \ref{fig:frame-segment-video} compares the average results achieved by spatial visual models (i.e., frame-level systems) and spatio-temporal visual models (i.e., short segment-level systems and video-level systems) on both datasets. It can be observed that the average CCC performance achieved by the spatial models outperforms the spatio-temporal settings for both tasks, i.e., while no model is reliable for inferring self-reported personality from facial behaviours, spatial models generally provide better self-reported personality predictions for four of the five traits than spatio-temporal models. Nevertheless, we found that a spatio-temporal model VAT achieved the best CCC performance in predicting apparent personality traits, with CCC $ > 0.6$ for predicting four of the five traits, which suggest that the VAT's temporal modelling strategy is suitable to extract apparent personality-related facial behaviour cues. Overall, spatial visual models consistently achieved better average performance than spatio-temporal models in predicting all apparent personality traits. These results indicate that the most standard spatio-temporal deep learning models (except VAT) can not effectively extract personality-related cues from human spatio-temporal facial behaviours, which may partially caused by the fact that they are not able to specifically model personality-related temporal contextual cues from the given video.

%%% 29/08/2022：添加图
\subsubsection{Short segment-level VS. video-level personality modelling}

\noindent Fig. \ref{fig:frame-segment-video} also compares short segment-level modelling and video-level modelling systems, where four widely-used spatio-temporal models introduced in Sec. \ref{subsec:widely-used models} are benchmarked. Although both short segment-level modelling and video-level modelling systems generally failed to provide reliable self-reported personality predictions, the short segment-level systems still achieved slightly better average performances than the video-level systems for four traits (i.e., Ope, Con, Agr, and Neu traits). Moreover, the short segment-level systems have clear advantages over the video-level systems for predicting all apparent personality traits. These results suggest that the short-term non-verbal behaviours contain crucial apparent personality-related cues while the standard video-level models which ignored short-term behaviours can not extract reliable apparent personality cues from the down-sampled long-term behaviours (i.e., down-sampling a long clip that contains around 400 frames to a segment containing 32 frames results in the loss of short-term behavioural cues).

%%% 29/08/2022：添加图（与Sec. 4.3.2的图结合）, 修改名称, HRNet, SENet，interpreimg 找相似处
\subsubsection{Audio  VS. Visual  VS. Audio-visual models} 

\begin{figure}
	\centering
	\subfigure[CCC results achieved for the UDIVA dataset.]{\label{}
		\includegraphics[width=8.8cm]{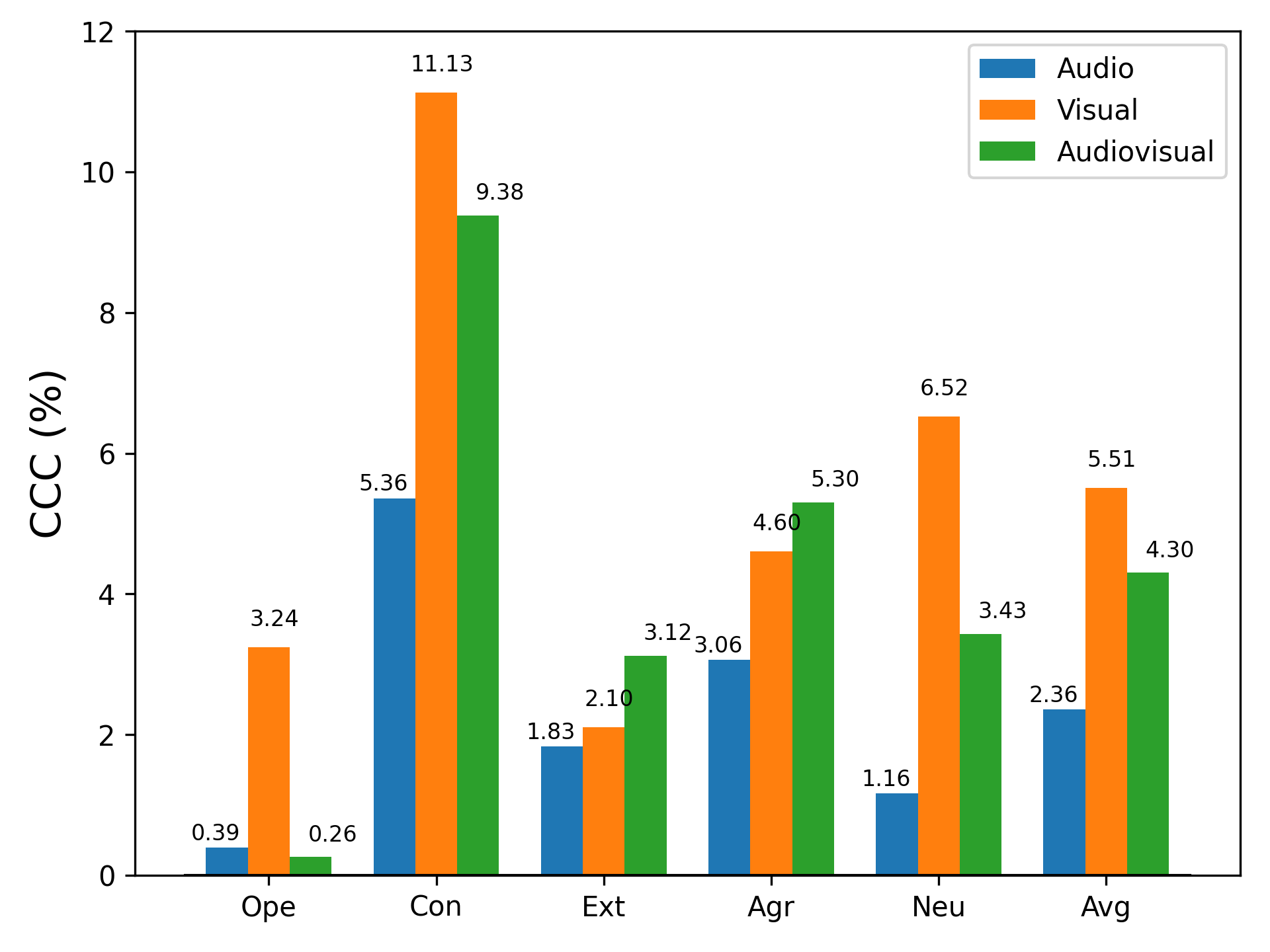}}
	\subfigure[CCC results achieved for the ChaLearn First Impression dataset.]{\label{}
		\includegraphics[width=8.8cm]{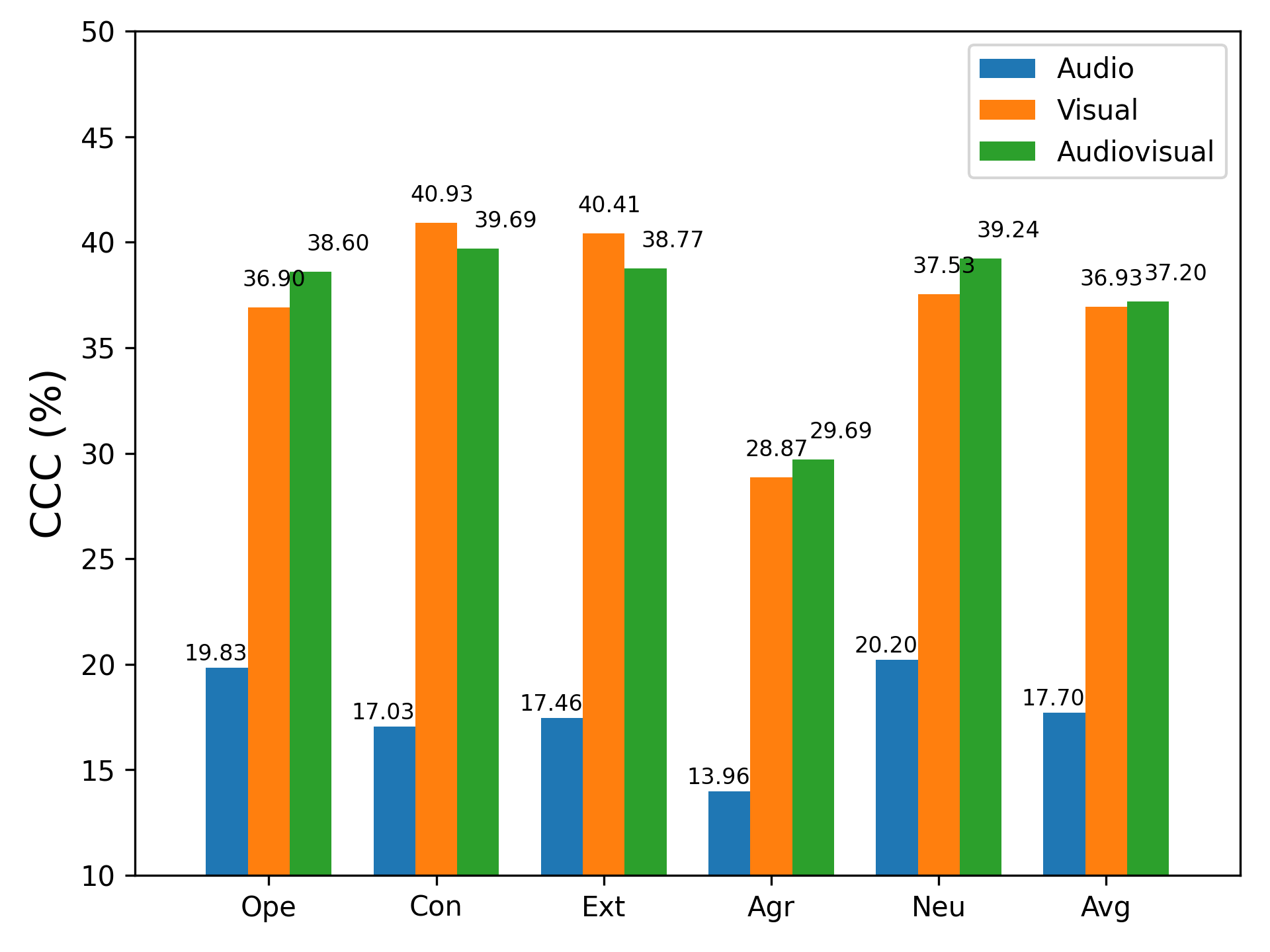}}
	\caption{Comparison between the average results achieved by the audio models, the visual models and the audio-visual models.} 
    \label{fig:audio-visual}
\end{figure}

\noindent As shown in Fig. \ref{fig:audio-visual}, visual models and audio-visual models on average generated superior personality recognition performances than audio models for all traits on both datasets. We found that CAM-DAN$^{+}$, HRNet and SENet visual models achieved relatively promising results for both SPR and APR tasks. However, the low CCC performance on SPR still shows that standard audio/visual machine learning models can hardly extract self-reported personality-related cues from audio-visual behaviours. As a result, the five audio-visual systems displayed in Table \ref{tb:CCC-chalearn} do not show clear and consistent advantages over their corresponding visual counterparts. In contrast, the audio and facial behaviours sometimes are more reliable in reflecting apparent personality traits. According to Table \ref{tb:CCC-chalearn}, adding the audio modality allows three out of the five audio-visual systems \cite{crnet,audiovisual_resnet,suman2022multi} achieved better performances over their corresponding visual methods in predicting apparent personality traits. However, when the extracted audio features are not well associated with apparent personality traits, they may even negatively impact the personality recognition ability of the visual system (e.g., he MFCC/logbank audio features degrade the DAN-based visual system \cite{8066355}). In summary, these results suggest that (i) non-verbal facial behaviours contain more discriminative personality-related cues than non-verbal audio signals for both self-reported and apparent personality recognition tasks; and (ii) if the audio modality can individually provide reliable personality predictions, it is expected to further enhance the predictions of the visual system.

\subsubsection{Summarising frame/segment-level personality predictions} 

\noindent We then compare the results achieved by the different strategies for summarising frame/segment-level personality predictions for obtaining clip-level personality prediction by: (i) simply averaging frame-level personality predictions as the clip-level personality prediction; and (ii) encoding a spectral representation of frame/segment-level predictions. Specifically, we first encode a pair of spectral heatmaps from the five-channel frame/segment-level personality trait predictions. Then, we train a regressor that contains seven 1D convolution layers followed by two fully connected layers to predict clip-level personality traits from the produced spectral heatmaps. Table \ref{tb:long-term} demonstrates the superiority of applying SFP to summarise frame/segment-level personality predictions over the other three strategies, as the performances of all models are improved using SFP. This also indicates that the spectral representation can effectively capture reliable personality-related spatio-temporal cues from frame/segment-level personality predictions. In contrast, the advantages of applying SFP to self-reported personality recognition are less evident. We assume this is because the frame-level predictions of self-reported personality traits lack reliability, making it challenging to capture dependable personality-related spatio-temporal cues from them.

\begin{table*}
\begin{tabularx}{\textwidth}{@{}l*{10}{C}c@{}}
\toprule
&Clip-level representation   & DAN\cite{8066355}   & ResNet\cite{audiovisual_resnet}  & CRNet\cite{hrnet}  & PersEmoN \cite{persemon} & CAM-DAN$^{+}$ \cite{interpret_img}  &  SENet\cite{senet}          & HRNet \cite{hrnet}           &  Swin\cite{swin_trans}     \\
\midrule 
\multirow{2}{*}{(C)}
&AFP       & 57.79        & 37.14         & 48.94          & 20.64           & 59.92          & 51.22         & 52.70          & 23.33 \\
&LRP       & 55.07(-2.72) & 30.46(-6.68)  & 44.11(-4.83)   & 26.47(+5.83)    & 58.45 (-1.47)  & 49.30(-1.92)  & 49.22(-3.48)   & 19.58(-3.75) \\
&MLP       & 51.41(-6.38) & 37.71(+0.57)  & 48.16(-0.78)   & 27.99(+7.35)    & 58.13 (-1.79)  & 51.81(+0.59)  & 53.39(+0.69)   & 20.07(-3.26)\\
&SFP       & 58.29(+0.50) & 39.80(+2.66)  & 50.03(+1.09)   & 28.81(+8.17)    & 62.09(+2.17)   & 51.54(+0.32)  & 53.14(+0.44)   & 23.66(+0.33) \\
\midrule
\multirow{2}{*}{(U)}
&AFP       & 7.36         & 8.98          & 13.11          & 0.68            & 10.59          & 11.91         & 10.07          & 6.44 \\
&LRP       & 0.05(-7.31)  & -0.06(-9.04)  & 9.52(-3.59)    & 6.38(+5.7)    & 15.32(+4.73)    & 9.61(-2.3)   & 17.50(+7.43)   & 5.80(-0.64) \\
&MLP       & 1.01(-6.35)  & 2.68 (-6.3)   & 6.73(-6.38)    & 5.12(+4.44)     & 8.03(-2.56)    & 9.07(-2.84)  & 8.63 (-1.44)   & 0.61(-5.83)\\
&SFP       & 3.55(-3.81)  & 10.21(+1.23)  & 1.46(-11.65)   & 0.20(-0.48)     & 1.06(-9.53)    & 2.39(-9.52)   & 2.00(-8.07)    & 0.11(-6.33) \\ 
\bottomrule
\end{tabularx}
\caption{Long-term modelling CCC (\%) results of the benchmarked frame/short segment-level personality computing visual models, where (C) denotes results achieved on ChaLearn Impression dataset, and (U) denotes results achieved on UDIVA dataset. The AFP, LRP, MLP and SFP represent the clip-level prediction computed from all frame-level predictions using averaging, linear regression, MLP and spectral representation, respectively.}
\label{tb:long-term}
\end{table*}

% 1 prediction from frame images
% 2 mlp on extracted frame images
% 3 conv-net on extracted frame images
% 4 conv-net on extracted frame features

% \begin{table*}
% \begin{tabularx}{\textwidth}{@{}l*{10}{C}c@{}}
% \toprule
% Methods   &  deep-bimodal     & interpret-img    &  senet           & hrnet           &  swin-transformer     \\ 
% \midrule  
%  1        & 0.9093            & 0.9118           & 0.9051           & 0.9050          &  0.8907 \\
%  2        & 0.9095(+0.0002) ) & 0.9117(-0.0001)  & 0.9053(+0.0002)  & 0.9053(0.0003)  & 0.8913(+0.0005)  \\ 
% \bottomrule
% \end{tabularx}
% \caption{ Long-term modelling results of different approaches on different models, where 1 denotes the original prediction accuracy values, 2 denotes convolutional neural network on extracted frame images}
% % \label{table:widely-used visual models}
% \end{table*}

\subsubsection{Other factors}

\noindent In addition, our analysis also revealed the following key findings: (i) jointly predicting all five traits yielded superior performance compared to models predicting each trait individually; (ii) models' performances are largely influenced by the temporal scale of the input, especially for APR; and (iii) incorporating metadata of the subjects did not lead to a clear improvement in the performance of non-verbal behaviour-based SPR, which remained poor; (iv) The results of self-reported personality recognition are highly influenced by the scenarios in which the data is recorded. Results achieved for the UDIVA show that the best performances of various models were achieved under the 'Ghost' scenarios. This scenario requires each participant to not only solve the problem but also interact with the other participant, both of which would foster rich personality-related behaviours expressed by them \cite{gundogdu2017investigating,zhang2001thinking}. However, combining behaviours expressed from all tasks does not show clear benefits for SPR models, suggesting that other tasks could not effectively trigger participants to express personality-related behaviours.

% SPR results are highly dependent on the data recording settings (e.g., task), where the majority of evaluated systems achieved the best performances under the ‘Ghost’ scenarios contained in the UDIVA dataset, while combining behaviours expressed from all tasks in the UDIVA dataset does not show clear benefits for SPR models. The detailed results are provided in the Supplementary Material.

\subsection{Discussion}
\label{subsec:result-dis}

\noindent In addition to reporting the results achieved by the benchmarked models and discussing the results achieved by the different settings, this section discusses other relevant aspects of the benchmarked models, including: (i) some reproduced models achieved lower results than the originally reported results; (ii) personality traits having different relationships with the non-verbal behaviours; (iii) the benchmarked models generally have poor performance in SPR; and (iv) challenges and research gaps in the current audio-visual personality computing models.

\subsubsection{Some reproduced approaches have lower performances than originally reported}

\noindent As compared in Table \ref{tb:reproduced}, an important finding is that 4 out of 7 reproduced models failed to achieve similar or superior results compared to their originally reported results on the ChaLearn First Impression dataset, even when we strictly followed the same pre-processing and training settings as reported in the original papers. This could be caused by various reasons, including incomplete reporting of training, evaluation and pre-processing details in the original papers \cite{crnet,persemon}, excluding the text modality \cite{crnet}, or random factors involved in their training and evaluation processes (e.g., randomly selected images from each video/segment \cite{bi_modal_lstm,audiovisual_resnet,crnet,persemon}, randomly cropped image regions \cite{audiovisual_resnet,crnet} as well as the randomly cropped 50176 dimensional audio features \cite{audiovisual_resnet}). %Moreover, the results achieved by most reproduced models are not as good as the corresponding systems that utilized the proposed standardised pre-processing and training steps. This validated the effectiveness of the proposed standardised pre-processing pipeline, which can be adopted by future researchers to report on their proposed models for fair comparisons.

%%% one column
\setlength{\tabcolsep}{2pt}
\begin{table}[t!]
	\begin{center}
    \resizebox{1\linewidth}{!} {
		\begin{tabular}{l c c c  l}
			\toprule
                & Model & Reproduced & Standardised  & Reported    \\
            % \hline 
            \hline
            % \multirow{6}*{Audio} 
                & DAN\cite{8066355}                      & 0.9131  & 0.9093  & 0.9130  \\
                & CAM-DAN$^{+}$\cite{interpret_img}      & 0.9120  & 0.9118  & 0.9120  \\
                & Amb-Fac-VGGish\cite{suman2022multi}    & 0.9127  & 0.9102  & 0.9146  \\ \hline
                & Bi-modal CNN-LSTM\cite{bi_modal_lstm}  & 0.8812  & 0.8813  & 0.9121  \\
                & ResNet\cite{audiovisual_resnet}        & 0.8939  & 0.8964  & 0.9109  \\
                & CRNet\cite{crnet}                      & 0.9079  & 0.9040  & 0.9188  \\
                & PersEmoN\cite{persemon}                & 0.8920  & 0.8920  & 0.9170  \\
                % & FFT\cite{interpre-aud}                 & 0.8598  & 0.8324  & 0.9009  \\

			\bottomrule
		\end{tabular}
        }
	\end{center}
	\caption{Average ACC results of five personality traits achieved for the reproduced models on the ChaLearn First Impression dataset. \textit{Reproduced} denotes the data was processed based on the method described in original papers; \textit{Standardised} indicates the results obtained from the system used our standardised data processing pipeline; and \textit{Reported} denotes the performances provided by the original papers.}  
 \label{tb:reproduced}
\end{table}
\setlength{\tabcolsep}{1.4pt}

%\textbf{Full images VS Face images:} As illustrated in Fig. \ref{fig:UDIVA_face_frame} and Fig. \ref{fig:Chalearn_face_frame}, standard deep learning models generally provide more reliable true personality and apparent personality predictions from face frames, despite that neither full frames nor face frames provide reliable true personality predictions. In contrast, apparent personality predictions generated by all visual models using both face frames and full frames have more than $0.37$ average CCC with the ground-truths, where face frame-based predictions still have $1.62\%$ CCC advantage over full frame-based predictions. These results may suggest that although some previous studies \cite{} claimed that backgrounds and spatial contextual information can provide informative cues for personality recognition, they may also contain personality-unrelated noises that can negatively impact on the personality recognition. In short, the results achieved by a set of benchmarked standard deep learning models show that the personality-related cues contained in backgrounds and spatial contextual information can not compensate their negative impacts for personality recognition.

%%% 29/08/2022：添加表: 每个trait audio平均结果， visual 平均结果， audio-visual 平均结果，雨
\subsubsection{Relationships between the personality traits and non-verbal behaviours}

\noindent Fig. \ref{fig:frame-segment-video} show that self-reported Conscientiousness and Neuroticism traits are relatively reliable to be inferred by static facial displays than other traits. Self-reported Extraversion trait is less related to static facial behaviours but more associated with long-term spatio-temporal facial behaviours. In terms of apparent personality traits, it can be seen that they have similar relationship patterns with facial behaviours of three temporal scales (frame-level, short segment-level and video-level), i.e., the behaviours of all three temporal scales are more reliable to infer Conscientiousness, Extraversion, Neuroticism, and Openness traits, while all of their associations with the Agreeableness trait are clearly weaker. 

While facial behaviours are clearly more informative than audio behaviours in predicting all personality traits, Fig. \ref{fig:audio-visual} illustrates that the self-reported Conscientiousness and Neuroticism traits can be better reflected by facial behaviours while Extraversion and Openness traits are less associated with facial behaviours. Again, both audio and facial behaviours are more reliable in inferring apparent Conscientiousness, Extraversion, Neuroticism, and Openness traits but are less correlated with the Agreeableness trait. 

Based on these results, we can conclude that each true or apparent personality trait has a unique relationship with human non-verbal behaviours, including their temporal scales and modalities. In particular, we found that: (i) non-verbal audio-visual behaviours are more reliable in inferring apparent personalities; (ii) facial behaviours are more informative than audio behaviours in inferring personality traits; (iii) self-reported Conscientiousness and Neuroticism traits can be easier to be inferred from non-verbal facial behaviours than other self-reported traits; and (iv) apparent Agreeableness trait is more difficult to be predicted from facial behaviours than other apparent personality traits.

\subsubsection{Poor performances in self-reported personality recognition}

\begin{figure}
    \centering
    \subfigure[Correlation between each trait and facial behaviours.]{
        % \label{subfig:ccc_per_impression_face_frame}
	\includegraphics[width=8.8cm]{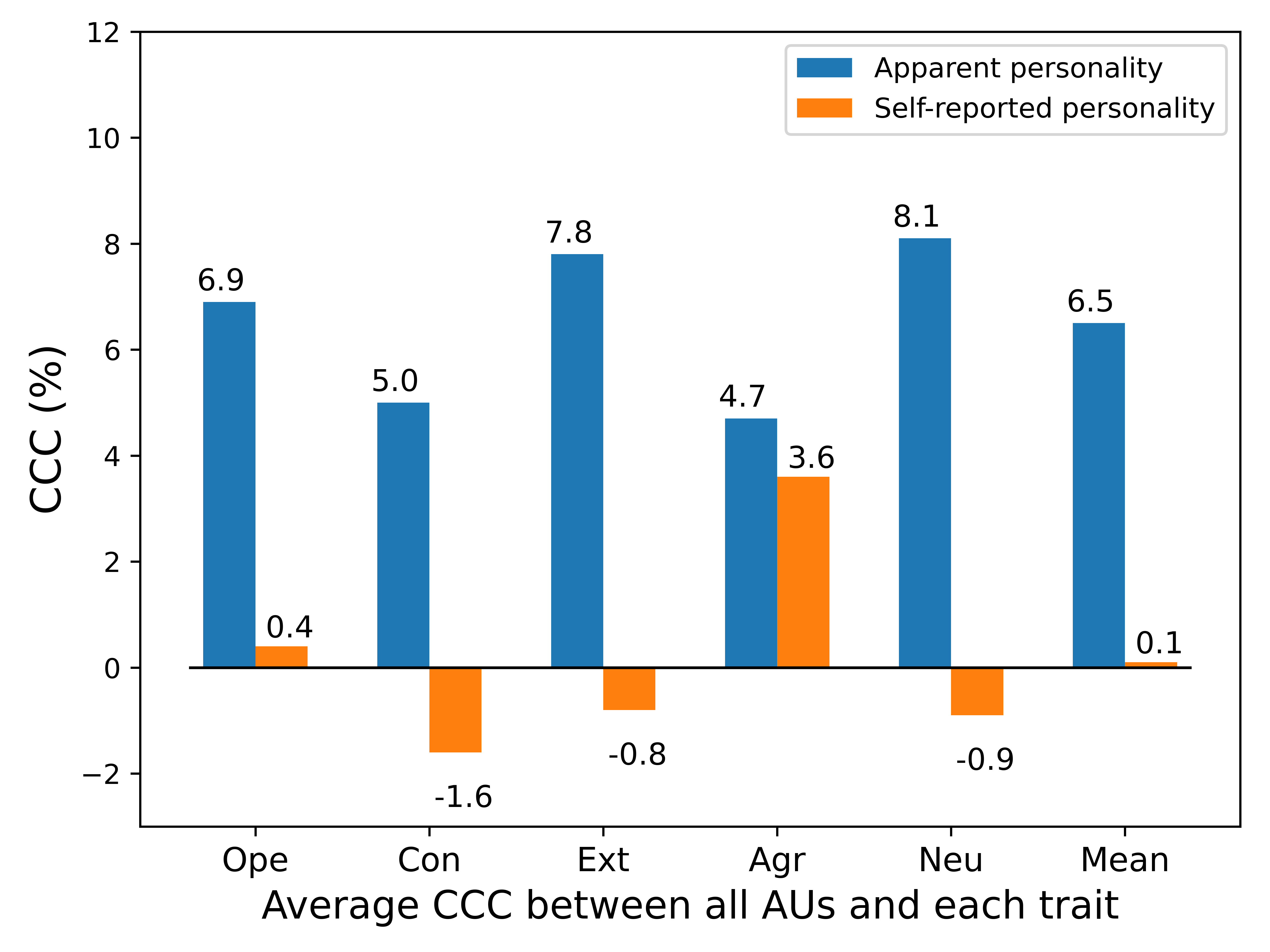}

    }
    
     \subfigure[Correlation between each trait and audio behaviours.]{
        % \label{subfig:acc_per_impression_face_frame}
	\includegraphics[width=8.8cm]{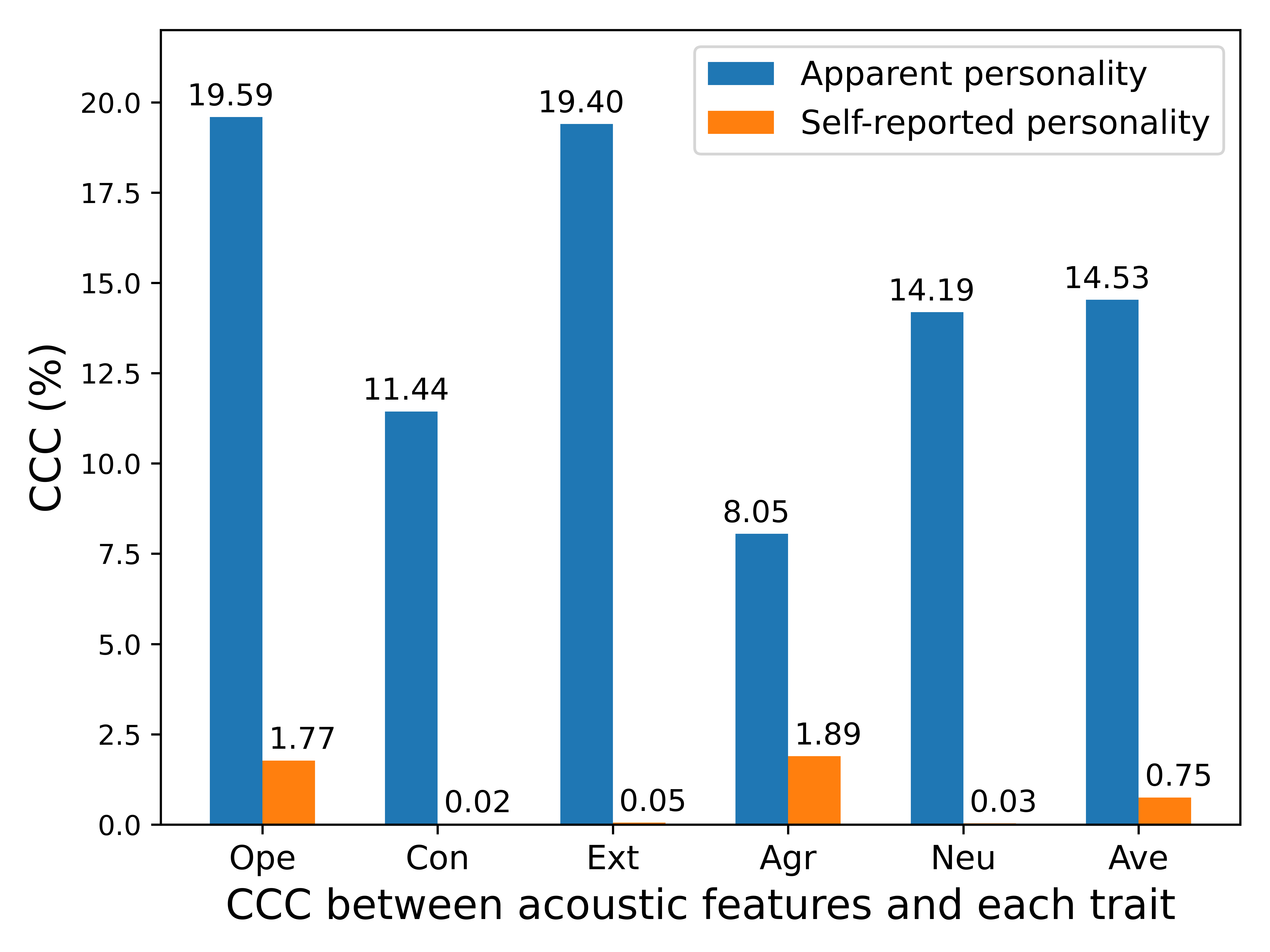}
    }
    \subfigure[Correlation between each AU and personality.]{
        % \label{subfig:acc_per_impression_face_frame}
	\includegraphics[width=8.8cm]{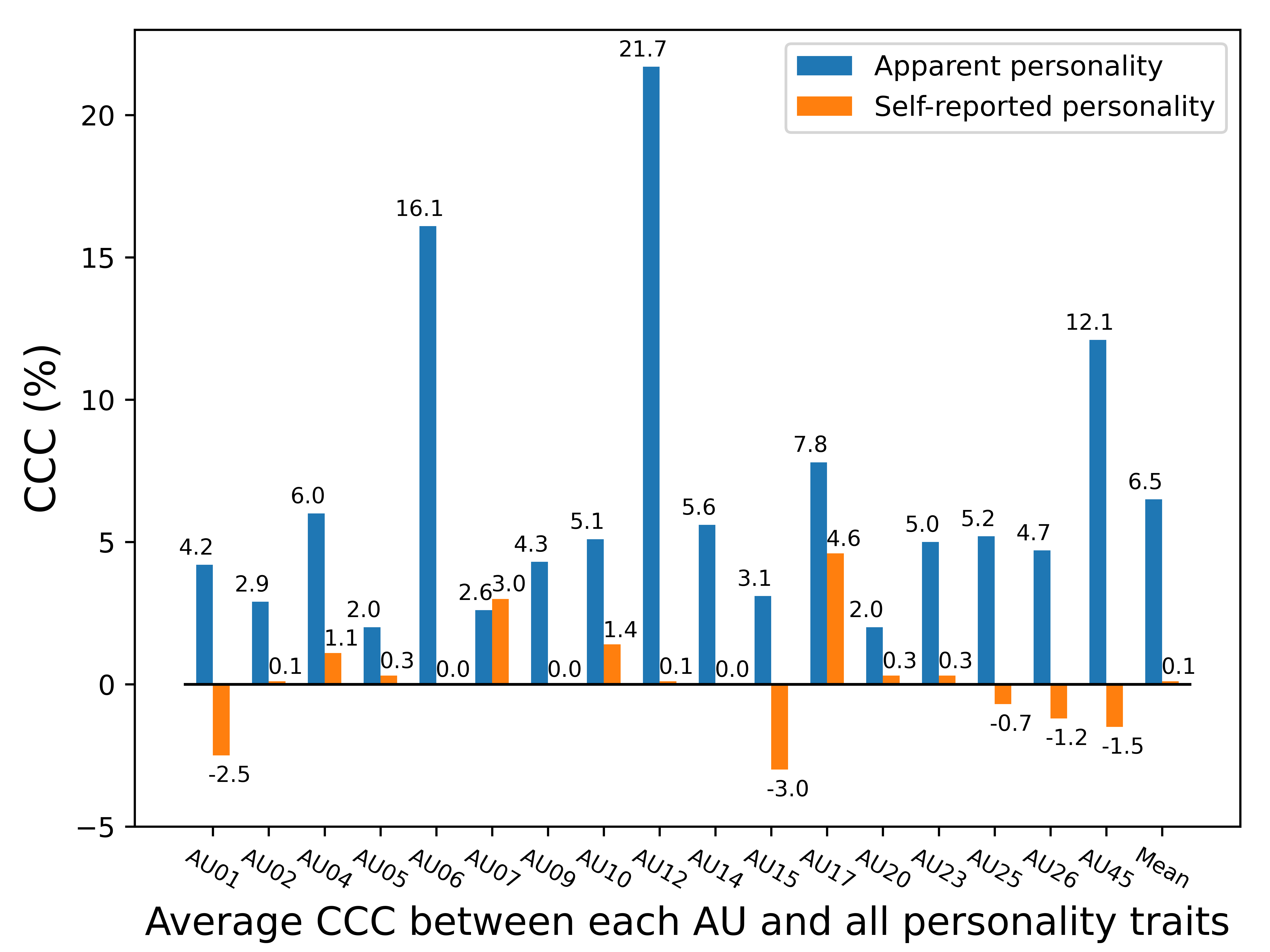}
    }

    \caption{Correlation among non-verbal behaviours and personality traits. The data are AU values extracted from every frame in each ChaLearn 2016 and UDIVA video.} 
    % \label{fig:Chalearn_face_frame}
    \label{Fig:primitives}
\end{figure}

\noindent To gain comprehensive insights, we followed \cite{song2020spectral} to compute the correlations between 17 automatically detected human facial action units (AUs) (using OpenFace 2.0  \cite{baltrusaitis2018openface}) as well as 988 acoustic features identified with OpenSmile \cite{eyben2010opensmile}, and each self-reported/apparent personality trait, respectively. Fig. \ref{Fig:primitives} clearly demonstrates that all human non-verbal behaviours have much lower correlations with self-reported personality traits. This could be a potential reason that non-verbal behaviour-based deep learning models can more effectively predict apparent personality traits over self-reported personality traits. We attribute this disparity to the fact that apparent personality traits are annotated by observing the subjects' behaviours, which is not taken into account during the annotation of self-reported personality traits.

% \begin{figure*}
% \centering 
% % \includegraphics[width=0.96\textwidth]{CCC_between_AUs_and_traits_2.png} 
% \includegraphics[width=0.8\textwidth]{images/CCC_between_AUs_and_traits_3.png} 
% \caption{\textcolor{blue}{Correlation among non-verbal behaviours and personality traits.}} 
% \label{Fig:primitives}
% \end{figure*}

Another possible reason for the lower performance achieved in self-reported personality recognition may be the limited number of training clips/subjects in the UDIVA dataset. Compared to the ChaLearn First Impression dataset (6000 training clips from 2624 subjects and 2000 validation clips from 1484 subjects), the UDIVA dataset comprises only 232 training clips from 99 candidates and 36 validation clips from 20 candidates. To mitigate the effects stemming from variations in the number of training and validation clips, we randomly select 232 training clips from 99 candidates and 36 validation clips from 20 candidates from the ChaLearn First Impression dataset (i.e., the list is provided in \footnote{\url{https://github.com/liaorongfan/DeepPersonality/tree/add_new_data_loader/datasets/chalearn16_few_candidates}} for the reproducible purpose). We also extracted 15-second segments from each UDIVA clip, aligning with the duration of clips in the ChaLearn First Impression dataset. Based on these reduced datasets, we evaluated the performance of the six best deep learning models. Results (reported in the Supplementary Material) show that less training/validation samples and shorter clip length lead to the significant performance drop in APR and SPR performances, respectively, which suggest that the lower performance achieved for SPR could be attributed to the limited number of clips/subjects in the UDIVA dataset. However, it can be observed that even with the limited number of short clips, the benchmarked models still achieved better results for APR.

Finally, we compared three types of systems: (i) systems taking individual static facial images, relying on static facial expressions to make predictions; (ii) systems taking selected static images with neutral facial display, making predictions based on identity cues; and (iii) systems taking anonymized multi-channel AU time-series, predicting personality based on temporal facial behaviours without considering identity information. The detailed results are provided in the supplementary material, where we found that APR models primarily relied on behavioural and expression cues rather than identity cues. In contrast, SPR models relied on identity cues, and thus achieved very poor performances.  This observation suggests that human non-verbal audio-visual behaviours might not directly reflect the self-reported personality, and taking into account other contextual aspects might be important, which in itself is an interesting finding that has not been investigated by prior works.

%six deep learning models are benchmarked, including two audio models (CRNet and VGGish), two visual models (HRNet and  VAT) and two audio-visual models (CRNet and  Amb-Fac-VGGish), as they achieved top-2 performances in each modality group.

\subsubsection{Challenges and research gaps}
\label{subsec:challenges and gaps}

\noindent Although an increasing number of automatic personality computing approaches have been proposed in recent years, both their reported results and our benchmarking results show that existing approaches are not very reliable for inferring personality traits, especially for self-reported personality. We can summarise the research gaps of existing personality computing approaches as follows:
\begin{itemize}
    
    \item (i) Most existing studies \cite{8066355,audiovisual_resnet,interpret_img,deep_bimodal,eddine2017personality,tellamekala2021apparent,tellamekala2022dimensional} attempt to infer personality traits directly from static frames or short audio-visual segments without considering contextual information. Since subjects that have different personalities can show similar behaviours in a single frame or a short audio-visual segment (e.g., different subjects can show a happy expression in a single frame), a static frame or a short audio-visual segment is not reliable for inferring personality. Also pairing a static frame or a short audio-visual segment with clip-level personality labels to train models results in an ill-posed machine learning problem, making the trained model theoretically not have good generalization capability. Our results show that the spectral representation-based long-term modelling of frame/short segment-level predictions produced by reliable frame/short segment-based systems (e.g., most frame/short segment-level APR models) leads to enhanced personality recognition results (Table \ref{tb:long-term}).

    \item (ii) Existing clip-level personality models can partially address the aforementioned ill-posed machine learning problem during model training. However, our benchmark shows that their results are worse as compared to the static frames or short audio-visual segments-based approaches. This can be explained by the fact that most clip-level spatio-temporal personality computing models \cite{crnet} (including the benchmarked standard spatio-temporal models) require down-sampling the original video, i.e., a large number of frames are discarded, i.e., crucial short-term personality-related cues are ignored during training and inference stages. While some clip-level personality computing studies \cite{song2021self,shao2021personality,song2022learning}  retain almost all frames in a video, these methods are time-consuming to implement and train as they need to individually train a person-specific network for each subject. In short, there lacks an efficient clip-level audio-visual personality computing framework that can extract personality-related cues from all frames of the input data.

    \item (iii) Most existing deep learning-based personality computing approaches simply pair the given audio-visual data and labels to train models, i.e., they do not take the personality-related domain knowledge into consideration when designing and training their models. As we can see from the benchmarked results, such strategies can not produce models that are able to provide reliable personality predictions. In other words, it is necessary for future studies to systematically review the physiological and psychological findings to concretely model the relationship between true/apparent personality traits and human non-verbal behaviours (e.g., the behaviours/temporal patterns that are more frequently displayed by subjects that have a certain personality traits pattern), and integrate such knowledge to design more advanced personality computing models and training strategies.

\end{itemize}

Besides the research gaps, we also discuss the challenges for both current and future researchers in developing new and more advanced personality models as follows:
\begin{itemize}
    
    \item (i) There is a need of open-source and easily implemented code that allow researcher to improve the existing approaches or develop more advanced models based on it. 
    
    \item (ii) Existing approaches usually have different and complex pre-processing strategies. A large number of the published approaches do not provide full details of their pre-processing steps, model settings and training strategies due to the limited page numbers. Subsequently, it is difficult for researchers to fully reproduce their methods and results, i.e., our benchmarking results are not as the same as the results reported in their original papers.
    
    \item (iii) Due to the limited number of publicly available datasets, most existing personality computing models are only evaluated on a single dataset (e.g., ChaLearn First Impression dataset). Based on the results achieved by our benchmark, we found the same model can have completely different performances in predicting self-reported or apparent personality traits on different a dataset. Therefore, we assume that personality computing models should be evaluated on more than one dataset (i.e., multiple apparent personality datasets or self-reported personality datasets) which have different data recording settings and scenarios. Otherwise, if a model achieves excellent performance on a single dataset, we can only conclude that this model is well adapted to the dataset at hand, but can not be made about its reliability and capability in predicting personality in general.

\end{itemize}

\section{Conclusion}

\noindent In this paper, we propose the first audio-visual personality computing benchmark for both self-reported and apparent personality recognition tasks, which are evaluated on the two widely-used and publicly available personality computing datasets: the ChaLearn First Impression dataset and the ChaLearn UDIVA self-reported personality dataset. We first benchmarked seven visual models, six audio models and five audio-visual models that have been published and evaluated on the ChaLearn First Impression datasets. We also benchmarked seven visual deep learning models that are widely used for visual problems, which have not been applied to video-based personality computing before. Our open-source, easy-to-use and standardised framework can be utilized to not only develop new personality computing models but also conduct quick yet robust evaluation of new models on both self-reported and apparent personality traits recognition tasks.

Building on our benchmarked models, we systematically evaluated various factors that impact personality computing. This led us to make the following conclusions: (i) using cropped and aligned face images as input generally led models to produce slightly better personality predictions than using images with background; (ii) the static models frequently achieved better performances than most spatio-temporal models for personality recognition; (iii) visual models frequently achieved superior results than audio models on personality recognition, indicating that subjects' facial behaviours contain more discriminative apparent personality-related cues than non-verbal audio behaviours; (iv) most models achieved better APR results than SPR even when different datasets were used, which suggests that subjects do not show their self-reported personality through non-verbal audio-visual behaviours; and (v) each personality trait has a unique relationship with human non-verbal behaviours (i.e., modalities and temporal scale), and thus even the same data contribute differently for personality recognition.
We also discussed the current research gaps and challenges in audio-visual personality computing.

Our work provides a standardized data loading, pre-processing and model training framework as well as a set of reproduced personality computing models, which partially address the discussed issues. We are aware that our benchmark cannot accommodate all machine learning strategies, (e.g., all previous personality computing models, pre/post-processing strategies). However, creating an extensive benchmark should be a joint community effort. Therefore, we have made our benchmark fully open-source, accompanied by detailed guidelines, and welcome contributors from the international research community to enhance various aspects of the pipeline. Our future endeavours will focus on annual improvements to the benchmark by incorporating newly published personality computing models, evaluating benchmarked models on novel personality datasets, and integrating additional modalities (e.g., language, interaction cues). Additionally, we aim to accommodate more machine learning strategies, including advanced self-supervised learning and data augmentation techniques, with our extensions and with the assistance of other researchers  working on this research area. We hope that this paper serves as an open call to all interested researchers to join forces with us to integrate their personality computing models into our publicly available framework. Given the challenges of directly inferring self-reported personality from audio-visual behaviours, our future work will also aim to investigate the impact of other contextual cues (e.g., data collection settings/scenarios), as previous studies \cite{song2022learning,shao2021personality} suggested that the behaviours expressed during self-presentation \cite{human2012your} or selfies (e.g., photos and videos) \cite{kaurin2018selfies} are more informative for displaying subjects’ self-reported personality traits. We will further follow  \cite{human2012your,kaurin2018selfies} to investigate more efficient and effective ways to model personality-related person-specific cognitive processes. In addition, more comprehensive and trustworthiness annotation, behaviour modalities and long-term datasets will also be essential for future self-reported personality recognition studies.

\ifCLASSOPTIONcompsoc
\section*{Acknowledgments}
\noindent\textbf{Funding:} The work of Siyang Song was funded partially by the EPSRC/UKRI under grant ref. EP/R030782/1, and partially by the European Union’s Horizon 2020 research and innovation programme project WorkingAge, under grant agreement No. 82623. Hatice Gunes is supported by the EPSRC/UKRI under grant ref. EP/R030782/1.

% \noindent\textbf{Open Access Statement:} For the purpose of \textit{open access}, the authors have applied a Creative Commons Attribution (CC BY) license to any Accepted Manuscript version arising. 

% \noindent\textbf{Data Access Statement:} This study involves secondary analyses of existing datasets, that are described and cited in the text. Licensing restrictions prevent sharing of the datasets. 

% \noindent\textbf{Code~Access:} \url{https://github.com/liaorongfan/DeepPersonality}. 

\else
  \section*{Acknowledgment}
\fi

\ifCLASSOPTIONcaptionsoff
  \newpage
\fi



%\bibliographystyle{IEEEtran}
%\begin{thebibliography}{1}

%H.~Kopka and P.~W. Daly, \emph{A Guide to {\LaTeX}}, 3rd~ed.\hskip 1em plus

%\end{thebibliography}
\normalem
\bibliographystyle{IEEEtran}

\bibliography{egbib}

\begin{thebibliography}{10}
\providecommand{\url}[1]{#1}
\csname url@samestyle\endcsname
\providecommand{\newblock}{\relax}
\providecommand{\bibinfo}[2]{#2}
\providecommand{\BIBentrySTDinterwordspacing}{\spaceskip=0pt\relax}
\providecommand{\BIBentryALTinterwordstretchfactor}{4}
\providecommand{\BIBentryALTinterwordspacing}{\spaceskip=\fontdimen2\font plus
\BIBentryALTinterwordstretchfactor\fontdimen3\font minus \fontdimen4\font\relax}
\providecommand{\BIBforeignlanguage}[2]{{%
\expandafter\ifx\csname l@#1\endcsname\relax
\typeout{** WARNING: IEEEtran.bst: No hyphenation pattern has been}%
\typeout{** loaded for the language `#1'. Using the pattern for}%
\typeout{** the default language instead.}%
\else
\language=\csname l@#1\endcsname
\fi
#2}}
\providecommand{\BIBdecl}{\relax}
\BIBdecl

\bibitem{corr2020cambridge}
P.~J. Corr and G.~Matthews, \emph{The Cambridge handbook of personality psychology}.\hskip 1em plus 0.5em minus 0.4em\relax Cambridge University Press, 2020.

\bibitem{youn2000impulse}
S.~Youn and R.~J. Faber, ``Impulse buying: its relation to personality traits and cues,'' \emph{ACR North American Advances}, 2000.

\bibitem{sano2015recognizing}
A.~Sano, A.~J. Phillips, Z.~Y. Amy, A.~W. McHill, S.~Taylor, N.~Jaques, C.~A. Czeisler, E.~B. Klerman, and R.~W. Picard, ``Recognizing academic performance, sleep quality, stress level, and mental health using personality traits, wearable sensors and mobile phones,'' in \emph{2015 IEEE 12th International Conference on Wearable and Implantable Body Sensor Networks (BSN)}.\hskip 1em plus 0.5em minus 0.4em\relax IEEE, 2015, pp. 1--6.

\bibitem{jaiswal2019automatic}
S.~Jaiswal, S.~Song, and M.~Valstar, ``Automatic prediction of depression and anxiety from behaviour and personality attributes,'' in \emph{2019 8th International Conference on Affective Computing and Intelligent Interaction (ACII)}.\hskip 1em plus 0.5em minus 0.4em\relax IEEE, 2019, pp. 1--7.

\bibitem{asendorpf1998personality}
J.~B. Asendorpf and S.~Wilpers, ``Personality effects on social relationships.'' \emph{Journal of personality and social psychology}, vol.~74, no.~6, p. 1531, 1998.

\bibitem{briley2014genetic}
D.~A. Briley and E.~M. Tucker-Drob, ``Genetic and environmental continuity in personality development: a meta-analysis.'' \emph{Psychological bulletin}, vol. 140, no.~5, p. 1303, 2014.

\bibitem{afzal2019personality}
S.~Afzal, B.~Dempsey, C.~D'Helon, N.~Mukhi, M.~Pribic, A.~Sickler, P.~Strong, M.~Vanchiswar, and L.~Wilde, ``The personality of ai systems in education: experiences with the watson tutor, a one-on-one virtual tutoring system,'' \emph{Childhood Education}, vol.~95, no.~1, pp. 44--52, 2019.

\bibitem{mansour2021relating}
D.~Mansour, A.~B. Bhardwaj, and A.~Chopra, ``Relating ocean (big five) to job satisfaction in aviation,'' in \emph{2021 International Conference on Computational Intelligence and Knowledge Economy (ICCIKE)}.\hskip 1em plus 0.5em minus 0.4em\relax IEEE, 2021, pp. 285--289.

\bibitem{kaya2017multi}
H.~Kaya, F.~Gurpinar, and A.~Ali~Salah, ``Multi-modal score fusion and decision trees for explainable automatic job candidate screening from video cvs,'' in \emph{Proceedings of the IEEE conference on computer vision and pattern recognition workshops}, 2017, pp. 1--9.

\bibitem{xia2014socially}
F.~Xia, N.~Y. Asabere, H.~Liu, Z.~Chen, and W.~Wang, ``Socially aware conference participant recommendation with personality traits,'' \emph{IEEE Systems Journal}, vol.~11, no.~4, pp. 2255--2266, 2014.

\bibitem{EYSENCK1991773}
\BIBentryALTinterwordspacing
H.~Eysenck, ``Dimensions of personality: 16, 5 or 3?—criteria for a taxonomic paradigm,'' \emph{Personality and Individual Differences}, vol.~12, no.~8, pp. 773--790, 1991. [Online]. Available: \url{https://www.sciencedirect.com/science/article/pii/019188699190144Z}
\BIBentrySTDinterwordspacing

\bibitem{mitchell2007analysis}
J.~T. Mitchell, N.~A. Kimbrel, N.~E. Hundt, A.~R. Cobb, R.~O. Nelson-Gray, and C.~M. Lootens, ``An analysis of reinforcement sensitivity theory and the five-factor model,'' \emph{European Journal of Personality: Published for the European Association of Personality Psychology}, vol.~21, no.~7, pp. 869--887, 2007.

\bibitem{de2000cloninger}
F.~De~Fruyt, L.~Van~de Wiele, and C.~Van~Heeringen, ``Cloninger's psychobiological model of temperament and character and the five-factor model of personality,'' \emph{Personality and individual differences}, vol.~29, no.~3, pp. 441--452, 2000.

\bibitem{mccrae2008five}
R.~R. McCrae and P.~T. Costa~Jr, ``The five-factor theory of personality.'' 2008.

\bibitem{matthews2003personality}
G.~Matthews, I.~J. Deary, and M.~C. Whiteman, \emph{Personality traits}.\hskip 1em plus 0.5em minus 0.4em\relax Cambridge University Press, 2003.

\bibitem{ponce2016chalearn}
V.~Ponce-L{\'o}pez, B.~Chen, M.~Oliu, C.~Corneanu, A.~Clap{\'e}s, I.~Guyon, X.~Bar{\'o}, H.~J. Escalante, and S.~Escalera, ``Chalearn lap 2016: First round challenge on first impressions-dataset and results,'' in \emph{European conference on computer vision}.\hskip 1em plus 0.5em minus 0.4em\relax Springer, 2016, pp. 400--418.

\bibitem{interpret_img}
C.~Ventura, D.~Masip, and A.~Lapedriza, ``Interpreting cnn models for apparent personality trait regression,'' in \emph{Proceedings of the IEEE conference on computer vision and pattern recognition workshops}, 2017, pp. 55--63.

\bibitem{song2022learning}
S.~Song, Z.~Shao, S.~Jaiswal, L.~Shen, M.~Valstar, and H.~Gunes, ``Learning person-specific cognition from facial reactions for automatic personality recognition,'' \emph{IEEE Transactions on Affective Computing}, 2022.

\bibitem{moreno2020estimation}
M.~A. Moreno-Armend{\'a}riz, C.~A.~D. Mart{\'\i}nez, H.~Calvo, and M.~Moreno-Sotelo, ``Estimation of personality traits from portrait pictures using the five-factor model,'' \emph{IEEE Access}, vol.~8, pp. 201\,649--201\,665, 2020.

\bibitem{helm2020single}
D.~Helm and M.~Kampel, ``Single-modal video analysis of personality traits using low-level visual features,'' in \emph{2020 Tenth International Conference on Image Processing Theory, Tools and Applications (IPTA)}.\hskip 1em plus 0.5em minus 0.4em\relax IEEE, 2020, pp. 1--6.

\bibitem{eddine2017personality}
S.~Eddine~Bekhouche, F.~Dornaika, A.~Ouafi, and A.~Taleb-Ahmed, ``Personality traits and job candidate screening via analyzing facial videos,'' in \emph{Proceedings of the IEEE conference on computer vision and pattern recognition workshops}, 2017, pp. 10--13.

\bibitem{tellamekala2022dimensional}
M.~K. Tellamekala, T.~Giesbrecht, and M.~Valstar, ``Dimensional affect uncertainty modelling for apparent personality recognition,'' \emph{IEEE Transactions on Affective Computing}, 2022.

\bibitem{gurpinar2016combining}
F.~G{\"u}rp{\i}nar, H.~Kaya, and A.~A. Salah, ``Combining deep facial and ambient features for first impression estimation,'' in \emph{European conference on computer vision}.\hskip 1em plus 0.5em minus 0.4em\relax Springer, 2016, pp. 372--385.

\bibitem{persemon}
L.~Zhang, S.~Peng, and S.~Winkler, ``Persemon: a deep network for joint analysis of apparent personality, emotion and their relationship,'' \emph{IEEE Transactions on Affective Computing}, 2019.

\bibitem{beyan2019personality}
C.~Beyan, A.~Zunino, M.~Shahid, and V.~Murino, ``Personality traits classification using deep visual activity-based nonverbal features of key-dynamic images,'' \emph{IEEE Transactions on Affective Computing}, 2019.

\bibitem{deep_bimodal}
C.-L. Zhang, H.~Zhang, X.-S. Wei, and J.~Wu, ``Deep bimodal regression for apparent personality analysis,'' in \emph{European conference on computer vision}.\hskip 1em plus 0.5em minus 0.4em\relax Springer, 2016, pp. 311--324.

\bibitem{bi_modal_lstm}
A.~Subramaniam, V.~Patel, A.~Mishra, P.~Balasubramanian, and A.~Mittal, ``Bi-modal first impressions recognition using temporally ordered deep audio and stochastic visual features,'' in \emph{European conference on computer vision}.\hskip 1em plus 0.5em minus 0.4em\relax Springer, 2016, pp. 337--348.

\bibitem{gurpinar2016multimodal}
F.~G{\"u}rpinar, H.~Kaya, and A.~A. Salah, ``Multimodal fusion of audio, scene, and face features for first impression estimation,'' in \emph{2016 23rd International conference on pattern recognition (ICPR)}.\hskip 1em plus 0.5em minus 0.4em\relax IEEE, 2016, pp. 43--48.

\bibitem{crnet}
Y.~Li, J.~Wan, Q.~Miao, S.~Escalera, H.~Fang, H.~Chen, X.~Qi, and G.~Guo, ``Cr-net: A deep classification-regression network for multimodal apparent personality analysis.'' \emph{International Journal of Computer Vision}, vol. 128, no.~12, 2020.

\bibitem{aslan2019multimodal}
S.~Aslan and U.~G{\"u}d{\"u}kbay, ``Multimodal video-based apparent personality recognition using long short-term memory and convolutional neural networks,'' \emph{arXiv preprint arXiv:1911.00381}, 2019.

\bibitem{8066355}
X.-S. Wei, C.-L. Zhang, H.~Zhang, and J.~Wu, ``Deep bimodal regression of apparent personality traits from short video sequences,'' \emph{IEEE Transactions on Affective Computing}, vol.~9, no.~3, pp. 303--315, 2018.

\bibitem{audiovisual_resnet}
Y.~G{\"u}{\c{c}}l{\"u}t{\"u}rk, U.~G{\"u}{\c{c}}l{\"u}, M.~A. van Gerven, and R.~van Lier, ``Deep impression: Audiovisual deep residual networks for multimodal apparent personality trait recognition,'' in \emph{European conference on computer vision}.\hskip 1em plus 0.5em minus 0.4em\relax Springer, 2016, pp. 349--358.

\bibitem{song2020spectral}
S.~Song, S.~Jaiswal, L.~Shen, and M.~Valstar, ``Spectral representation of behaviour primitives for depression analysis,'' \emph{IEEE Transactions on Affective Computing}, 2020.

\bibitem{song2021self}
S.~Song, S.~Jaiswal, E.~Sanchez, G.~Tzimiropoulos, L.~Shen, and M.~Valstar, ``Self-supervised learning of person-specific facial dynamics for automatic personality recognition,'' \emph{IEEE Transactions on Affective Computing}, 2021.

\bibitem{curto2021dyadformer}
D.~Curto, A.~Clap{\'e}s, J.~Selva, S.~Smeureanu, J.~Junior, C.~Jacques, D.~Gallardo-Pujol, G.~Guilera, D.~Leiva, T.~B. Moeslund \emph{et~al.}, ``Dyadformer: A multi-modal transformer for long-range modeling of dyadic interactions,'' in \emph{Proceedings of the IEEE/CVF international conference on computer vision}, 2021, pp. 2177--2188.

\bibitem{shao2021personality}
Z.~Shao, S.~Song, S.~Jaiswal, L.~Shen, M.~Valstar, and H.~Gunes, ``Personality recognition by modelling person-specific cognitive processes using graph representation,'' in \emph{proceedings of the 29th ACM international conference on multimedia}, 2021, pp. 357--366.

\bibitem{dodd2023framework}
E.~Dodd, S.~Song, and H.~Gunes, ``A framework for automatic personality recognition in dyadic interactions,'' 2023.

\bibitem{welch2021listeners}
B.~Welch, M.~R. van Mersbergen, and L.~B. Helou, ``Listeners' perceptions of speaker personality traits based on speech,'' \emph{Journal of Speech, Language, and Hearing Research}, vol.~64, no.~12, pp. 4762--4771, 2021.

\bibitem{xu2021prediction}
J.~Xu, W.~Tian, G.~Lv, S.~Liu, and Y.~Fan, ``Prediction of the big five personality traits using static facial images of college students with different academic backgrounds,'' \emph{Ieee Access}, vol.~9, pp. 76\,822--76\,832, 2021.

\bibitem{hickman2022automated}
L.~Hickman, N.~Bosch, V.~Ng, R.~Saef, L.~Tay, and S.~E. Woo, ``Automated video interview personality assessments: Reliability, validity, and generalizability investigations.'' \emph{Journal of Applied Psychology}, vol. 107, no.~8, p. 1323, 2022.

\bibitem{salam2022learning}
H.~Salam, V.~Manoranjan, J.~Jiang, and O.~Celiktutan, ``Learning personalised models for automatic self-reported personality recognition,'' in \emph{Understanding Social Behavior in Dyadic and Small Group Interactions}.\hskip 1em plus 0.5em minus 0.4em\relax PMLR, 2022, pp. 53--73.

\bibitem{kassab2023vptd}
K.~Kassab, A.~Kashevnik, A.~Mayatin, and D.~Zubok, ``Vptd: Human face video dataset for personality traits detection,'' \emph{Data}, vol.~8, no.~7, p. 113, 2023.

\bibitem{hayat2019use}
H.~Hayat, C.~Ventura, and A.~Lapedriza, ``On the use of interpretable cnn for personality trait recognition from audio.'' in \emph{CCIA}, 2019, pp. 135--144.

\bibitem{2016}
\BIBentryALTinterwordspacing
Y.~Güçlütürk, U.~Güçlü, M.~A.~J. van Gerven, and R.~van Lier, ``Deep impression: Audiovisual deep residual networks for multimodal apparent personality trait recognition,'' \emph{Computer Vision – ECCV 2016 Workshops}, p. 349–358, 2016. [Online]. Available: \url{http://dx.doi.org/10.1007/978-3-319-49409-8_28}
\BIBentrySTDinterwordspacing

\bibitem{palmero2021context}
C.~Palmero, J.~Selva, S.~Smeureanu, J.~C.~J. Junior, A.~Clap{\'e}s, A.~Mosegu{\'\i}, Z.~Zhang, D.~Gallardo, G.~Guilera, D.~Leiva \emph{et~al.}, ``Context-aware personality inference in dyadic scenarios: Introducing the udiva dataset.'' in \emph{WACV (Workshops)}, 2021, pp. 1--12.

\bibitem{junior2019first}
J.~C. S.~J. Junior, Y.~G{\"u}{\c{c}}l{\"u}t{\"u}rk, M.~P{\'e}rez, U.~G{\"u}{\c{c}}l{\"u}, C.~Andujar, X.~Bar{\'o}, H.~J. Escalante, I.~Guyon, M.~A. Van~Gerven, R.~Van~Lier \emph{et~al.}, ``First impressions: A survey on vision-based apparent personality trait analysis,'' \emph{IEEE Transactions on Affective Computing}, 2019.

\bibitem{zhang2016deep}
C.-L. Zhang, H.~Zhang, X.-S. Wei, and J.~Wu, ``Deep bimodal regression for apparent personality analysis,'' in \emph{European conference on computer vision}.\hskip 1em plus 0.5em minus 0.4em\relax Springer, 2016, pp. 311--324.

\bibitem{joo2015automated}
J.~Joo, F.~F. Steen, and S.-C. Zhu, ``Automated facial trait judgment and election outcome prediction: Social dimensions of face,'' in \emph{Proceedings of the IEEE international conference on computer vision}, 2015, pp. 3712--3720.

\bibitem{dhall2016first}
A.~Dhall and J.~Hoey, ``First impressions-predicting user personality from twitter profile images,'' in \emph{International Workshop on Human Behavior Understanding}.\hskip 1em plus 0.5em minus 0.4em\relax Springer, 2016, pp. 148--158.

\bibitem{zhou2016learning}
B.~Zhou, A.~Khosla, A.~Lapedriza, A.~Oliva, and A.~Torralba, ``Learning deep features for discriminative localization,'' in \emph{Proceedings of the IEEE conference on computer vision and pattern recognition}, 2016, pp. 2921--2929.

\bibitem{celiktutan2014continuous}
O.~Celiktutan and H.~Gunes, ``Continuous prediction of perceived traits and social dimensions in space and time,'' in \emph{Image Processing (ICIP), 2014 IEEE International Conference on}.\hskip 1em plus 0.5em minus 0.4em\relax IEEE, 2014, pp. 4196--4200.

\bibitem{celiktutan2014maptraits}
O.~Celiktutan, F.~Eyben, E.~Sariyanidi, H.~Gunes, and B.~Schuller, ``Maptraits 2014-the first audio/visual mapping personality traits challenge-an introduction: Perceived personality and social dimensions,'' in \emph{Proceedings of the 16th International Conference on Multimodal Interaction}.\hskip 1em plus 0.5em minus 0.4em\relax ACM, 2014, pp. 529--530.

\bibitem{eyben2010opensmile}
F.~Eyben, M.~W{\"o}llmer, and B.~Schuller, ``Opensmile: the munich versatile and fast open-source audio feature extractor,'' in \emph{Proceedings of the 18th ACM international conference on Multimedia}, 2010, pp. 1459--1462.

\bibitem{qin2018modern}
R.~Qin, W.~Gao, H.~Xu, and Z.~Hu, ``Modern physiognomy: an investigation on predicting personality traits and intelligence from the human face,'' \emph{Science China Information Sciences}, vol.~61, no.~5, p. 058105, 2018.

\bibitem{cafaro2017noxi}
A.~Cafaro, J.~Wagner, T.~Baur, S.~Dermouche, M.~Torres~Torres, C.~Pelachaud, E.~Andr{\'e}, and M.~Valstar, ``The noxi database: multimodal recordings of mediated novice-expert interactions,'' in \emph{Proceedings of the 19th ACM International Conference on Multimodal Interaction}, 2017, pp. 350--359.

\bibitem{celiktutan2017multimodal}
O.~Celiktutan, E.~Skordos, and H.~Gunes, ``Multimodal human-human-robot interactions (mhhri) dataset for studying personality and engagement,'' \emph{IEEE Transactions on Affective Computing}, vol.~10, no.~4, pp. 484--497, 2017.

\bibitem{zhang2016joint}
K.~Zhang, Z.~Zhang, Z.~Li, and Y.~Qiao, ``Joint face detection and alignment using multitask cascaded convolutional networks,'' \emph{IEEE Signal Processing Letters}, vol.~23, no.~10, pp. 1499--1503, 2016.

\bibitem{dlib}
N.~Boyko, O.~Basystiuk, and N.~Shakhovska, ``Performance evaluation and comparison of software for face recognition, based on dlib and opencv library,'' in \emph{2018 IEEE Second International Conference on Data Stream Mining \& Processing (DSMP)}.\hskip 1em plus 0.5em minus 0.4em\relax IEEE, 2018, pp. 478--482.

\bibitem{suman2022multi}
C.~Suman, S.~Saha, A.~Gupta, S.~K. Pandey, and P.~Bhattacharyya, ``A multi-modal personality prediction system,'' \emph{Knowledge-Based Systems}, vol. 236, p. 107715, 2022.

\bibitem{ilmini2016persons}
K.~Ilmini and T.~Fernando, ``Persons’ personality traits recognition using machine learning algorithms and image processing techniques,'' \emph{Advances in Computer Science: an International Journal}, vol.~5, no.~1, pp. 40--44, 2016.

\bibitem{interpre-aud}
H.~Hayat, C.~Ventura, and {\`A}.~Lapedriza, ``On the use of interpretable cnn for personality trait recognition from audio,'' in \emph{CCIA}, 2019.

\bibitem{pyaudioanalysis}
T.~Giannakopoulos, ``pyaudioanalysis: An open-source python library for audio signal analysis,'' \emph{PloS one}, vol.~10, no.~12, p. e0144610, 2015.

\bibitem{librosa}
B.~McFee, C.~Raffel, D.~Liang, D.~P. Ellis, M.~McVicar, E.~Battenberg, and O.~Nieto, ``librosa: Audio and music signal analysis in python,'' in \emph{Proceedings of the 14th python in science conference}, vol.~8.\hskip 1em plus 0.5em minus 0.4em\relax Citeseer, 2015, pp. 18--25.

\bibitem{senet}
J.~Hu, L.~Shen, S.~Albanie, G.~Sun, and E.~Wu, ``Squeeze-and-excitation networks,'' 2019.

\bibitem{hrnet}
J.~Wang, K.~Sun, T.~Cheng, B.~Jiang, C.~Deng, Y.~Zhao, D.~Liu, Y.~Mu, M.~Tan, X.~Wang, W.~Liu, and B.~Xiao, ``Deep high-resolution representation learning for visual recognition,'' 2020.

\bibitem{vit}
A.~Dosovitskiy, L.~Beyer, A.~Kolesnikov, D.~Weissenborn, X.~Zhai, T.~Unterthiner, M.~Dehghani, M.~Minderer, G.~Heigold, S.~Gelly \emph{et~al.}, ``An image is worth 16x16 words: Transformers for image recognition at scale,'' \emph{arXiv preprint arXiv:2010.11929}, 2020.

\bibitem{swin_trans}
Z.~Liu, Y.~Lin, Y.~Cao, H.~Hu, Y.~Wei, Z.~Zhang, S.~Lin, and B.~Guo, ``Swin transformer: Hierarchical vision transformer using shifted windows,'' 2021.

\bibitem{3d_resnet}
K.~Hara, H.~Kataoka, and Y.~Satoh, ``Can spatiotemporal 3d cnns retrace the history of 2d cnns and imagenet?'' in \emph{Proceedings of the IEEE Conference on Computer Vision and Pattern Recognition (CVPR)}, 2018, pp. 6546--6555.

\bibitem{slowfast}
C.~Feichtenhofer, H.~Fan, J.~Malik, and K.~He, ``Slowfast networks for video recognition,'' 2019.

\bibitem{tpn}
C.~Yang, Y.~Xu, J.~Shi, B.~Dai, and B.~Zhou, ``Temporal pyramid network for action recognition,'' in \emph{Proceedings of the IEEE Conference on Computer Vision and Pattern Recognition (CVPR)}, 2020.

\bibitem{vat}
R.~Girdhar, J.~Carreira, C.~Doersch, and A.~Zisserman, ``Video action transformer network,'' 2019.

\bibitem{he2016deep}
K.~He, X.~Zhang, S.~Ren, and J.~Sun, ``Deep residual learning for image recognition,'' in \emph{Proceedings of the IEEE conference on computer vision and pattern recognition}, 2016, pp. 770--778.

\bibitem{song2018human}
S.~Song, L.~Shen, and M.~Valstar, ``Human behaviour-based automatic depression analysis using hand-crafted statistics and deep learned spectral features,'' in \emph{2018 13th IEEE International Conference on Automatic Face \& Gesture Recognition (FG 2018)}.\hskip 1em plus 0.5em minus 0.4em\relax IEEE, 2018, pp. 158--165.

\bibitem{biel2010voices}
J.-I. Biel and D.~Gatica-Perez, ``Voices of vlogging,'' in \emph{Fourth International AAAI Conference on Weblogs and Social Media}, 2010.

\bibitem{sanchez2011nonverbal}
D.~Sanchez-Cortes, O.~Aran, M.~S. Mast, and D.~Gatica-Perez, ``A nonverbal behavior approach to identify emergent leaders in small groups,'' \emph{IEEE Transactions on Multimedia}, vol.~14, no.~3, pp. 816--832, 2011.

\bibitem{miranda2018amigos}
J.~A. Miranda-Correa, M.~K. Abadi, N.~Sebe, and I.~Patras, ``Amigos: A dataset for affect, personality and mood research on individuals and groups,'' \emph{IEEE Transactions on Affective Computing}, vol.~12, no.~2, pp. 479--493, 2018.

\bibitem{correa2018amigos}
J.~A.~M. Correa, M.~K. Abadi, N.~Sebe, and I.~Patras, ``Amigos: A dataset for affect, personality and mood research on individuals and groups,'' \emph{IEEE Transactions on Affective Computing}, 2018.

\bibitem{rothbart2001investigations}
M.~K. Rothbart, S.~A. Ahadi, K.~L. Hershey, and P.~Fisher, ``Investigations of temperament at three to seven years: The children's behavior questionnaire,'' \emph{Child development}, vol.~72, no.~5, pp. 1394--1408, 2001.

\bibitem{ellis2001revision}
L.~K. Ellis and M.~K. Rothbart, ``Revision of the early adolescent temperament questionnaire,'' in \emph{Poster presented at the 2001 biennial meeting of the society for research in child development, Minneapolis, Minnesota}.\hskip 1em plus 0.5em minus 0.4em\relax Citeseer, 2001.

\bibitem{soto2017next}
C.~J. Soto and O.~P. John, ``The next big five inventory (bfi-2): Developing and assessing a hierarchical model with 15 facets to enhance bandwidth, fidelity, and predictive power.'' \emph{Journal of personality and social psychology}, vol. 113, no.~1, p. 117, 2017.

\bibitem{ashton2009hexaco}
M.~C. Ashton and K.~Lee, ``The hexaco--60: A short measure of the major dimensions of personality,'' \emph{Journal of personality assessment}, vol.~91, no.~4, pp. 340--345, 2009.

\bibitem{gundogdu2017investigating}
D.~Gundogdu, A.~N. Finnerty, J.~Staiano, S.~Teso, A.~Passerini, F.~Pianesi, and B.~Lepri, ``Investigating the association between social interactions and personality states dynamics,'' \emph{Royal Society open science}, vol.~4, no.~9, p. 170194, 2017.

\bibitem{zhang2001thinking}
L.-f. Zhang and J.~Huang, ``Thinking styles and the five-factor model of personality,'' \emph{European Journal of Personality}, vol.~15, no.~6, pp. 465--476, 2001.

\bibitem{baltrusaitis2018openface}
T.~Baltrusaitis, A.~Zadeh, Y.~C. Lim, and L.-P. Morency, ``Openface 2.0: Facial behavior analysis toolkit,'' in \emph{2018 13th IEEE international conference on automatic face \& gesture recognition (FG 2018)}.\hskip 1em plus 0.5em minus 0.4em\relax IEEE, 2018, pp. 59--66.

\bibitem{tellamekala2021apparent}
M.~K. Tellamekala, T.~Giesbrecht, and M.~Valstar, ``Apparent personality recognition from uncertainty-aware facial emotion predictions using conditional latent variable models,'' in \emph{2021 16th IEEE International Conference on Automatic Face and Gesture Recognition (FG 2021)}.\hskip 1em plus 0.5em minus 0.4em\relax IEEE, 2021, pp. 1--8.

\bibitem{human2012your}
L.~J. Human, J.~C. Biesanz, K.~L. Parisotto, and E.~W. Dunn, ``Your best self helps reveal your true self: Positive self-presentation leads to more accurate personality impressions,'' \emph{Social Psychological and Personality Science}, vol.~3, no.~1, pp. 23--30, 2012.

\bibitem{kaurin2018selfies}
A.~Kaurin, L.~Heil, M.~Wessa, B.~Egloff, and S.~Hirschm{\"u}ller, ``Selfies reflect actual personality--just like photos or short videos in standardized lab conditions,'' \emph{Journal of Research in Personality}, vol.~76, pp. 154--164, 2018.

\end{thebibliography}

\end{document}

\IEEEtitleabstractindextext{%

\begin{IEEEkeywords}

\end{IEEEkeywords}}

% \maketitle

% is not selected <OR> if conference mode is selected - because compsoc
% conference papers position the abstract like regular (non-compsoc)
\IEEEdisplaynontitleabstractindextext

% \fi
\IEEEpeerreviewmaketitle

\ifCLASSOPTIONcompsoc
% \IEEEraisesectionheading{\section{Introduction}\label{sec:introduction}}
\else

%   \section*{Acknowledgments}
\else
\fi

\ifCLASSOPTIONcaptionsoff
  \newpage
\fi

\appendix
\subsection{Results of Impression dataset}
\subsubsection{Models trained on audio data from  ChaLearn2016 impressison dataset}

%% audio modality
% +++++++++++++++++++++++++++++++++++++++++++++++++++++++++++++++++++++++++++++++++++
\setlength{\tabcolsep}{2pt}
\begin{table}[H]
	\begin{center}
\resizebox{1\linewidth}{!} {
		\begin{tabular}{|l| c| c c c c c l|}
			\toprule
            & Traits         & Open    & Consc  & Extrav   & Agree   & Neuro   & Avg.   \\
            \hline \hline
        \multirow{5}*{ACC} 
                & FFT             & 0.8303  & 0.8174  & 0.8502  & 0.8309  & 0.8331  & 0.8324 \\
                & MFCC/logfbank   & 0.8891  & 0.8790  & 0.8835  & 0.8967  & 0.8802  & 0.8857  \\ 
                & Bi-modal CNN    & 0.8835  & 0.8747  & 0.8785  & 0.8937  & 0.8774  & 0.8816  \\ 
                & ResNet          & 0.8822  & 0.8780  & 0.8782  & 0.8958  & 0.8820  & 0.8832  \\ 
                & CR-Net          & 0.9001  & 0.8895  & 0.8951  & 0.9022  & 0.8964  & 0.8967  \\ 
            \hline
        \multirow{5}*{PCC} 
                & FFT           &  0.0002  & -0.0001  & -0.0002  & -0.0002  & -0.0003  & -0.0002 \\
                & MFCC/logfbank &  0.3232  &  0.2588  &  0.2792  &  0.2373  &  0.3027  &  0.2803   \\ 
                & Bi-modal CNN  & -0.0124  & -0.0485  &  0.0356  & -0.0484  & -0.0664  &  0.0280    \\ 
                & ResNet        &  0.1828  &  0.2248  &  0.1241  &  0.2077  &  0.2668  &  0.2012   \\ 
                & CR-Net        &  0.4976  &  0.4370  &  0.4941  &  0.3858  &  0.5224  &  0.4674   \\ 
            \hline
        \multirow{5}*{CCC} 
                & FFT           &  0.0002  & -0.0001  & -0.0002  & -0.0002  & -0.0003 & -0.0002 \\
                & MFCC/logfbank &  0.1968  &  0.1497  &  0.1738  &  0.1295  &  0.1780 &  0.1655 \\ 
                & Bi-modal CNN  & -0.0004  & -0.0005  &  0.0004  & -0.0005  & -0.0008 &  0.0004 \\ 
                & ResNet        &  0.1293  &  0.0830  &  0.0458  &  0.1101  &  0.1548 &  0.1046 \\ 
                & CR-Net        &  0.4122  &  0.3406  &  0.3846  &  0.2857  &  0.4306 &  0.3707 \\ 
			\bottomrule
		\end{tabular}
        }
	\end{center}
	\caption{Models trained on audio data from talk session of true personality dataset}  
\label{tb:BRC}
\end{table}
\setlength{\tabcolsep}{1.4pt}

\subsubsection{Models trained on visual data from ChaLearn2016 impressison dataset}

Models are trained with frame data
%% frame images
% +++++++++++++++++++++++++++++++++++++++++++++++++++++++++++++++++++++++++++++++++++
\setlength{\tabcolsep}{2pt}
\begin{table}[H]
	\begin{center}
\resizebox{1\linewidth}{!} {
		\begin{tabular}{|l| c| c c c c c l|}
			\toprule
            & Traits         & Open    & Consc  & Extrav   & Agree   & Neuro   & Avg.   \\
            \hline \hline

        \multirow{7}*{ACC} 
        & Interpret-img    & 0.9088 & 0.9088 & 0.9071 & 0.9079 & 0.9024 & 0.9070  \\
        & PersEmoN         & 0.8944 & 0.8862 & 0.8902 & 0.8979 & 0.8873 & 0.8912  \\
        & senet            & 0.9061 & 0.9044 & 0.9045 & 0.9077 & 0.9027 & 0.9051 \\
        & hrnet            & 0.9057 & 0.9035 & 0.9042 & 0.9083 & 0.9036 & 0.9050 \\
        & swin-transformer & 0.8942 & 0.8881 & 0.8877 & 0.8980 & 0.8853 & 0.8907 \\
        & 3D-resnet        & 0.8830 & 0.8745 & 0.8781 & 0.8933 & 0.8771 & 0.8812 \\
        & slow-fast        & 0.8773 & 0.8597 & 0.8398 & 0.8800 & 0.8598 & 0.8633 \\    
        & tpn              & 0.8890 & 0.8793 & 0.8833 & 0.8961 & 0.8827 & 0.8861 \\
        & vat              & 0.9036 & 0.9066 & 0.9048 & 0.9042 & 0.8996 & 0.9038 \\
            \hline
        \multirow{7}*{PCC} 
        & Interpret-img    &  0.6041  & 0.6648  & 0.6316  & 0.5135  & 0.6004  & 0.6029  \\
        & PersEmoN         &  0.3945  & 0.3869  & 0.4353  & 0.2691  & 0.3906  & 0.3753 \\
        & senet            & 0.5782  & 0.6083  & 0.6056  & 0.4979  & 0.6114  & 0.5803  \\
        & hrnet            & 0.5746  & 0.6058  & 0.6013  & 0.5171  & 0.6221  & 0.5842  \\
        & swin-transformer & 0.3984  & 0.4150  & 0.3844  & 0.2878  & 0.3605  & 0.3692  \\
        & 3D-resnet        & 0.0024  & 0.0012  & -0.0036 & -0.0166 & -0.0034 & -0.004  \\
        & slow-fast        & 0.3232  & 0.3275  & 0.2992  & 0.2248  & 0.2959  & 0.2941  \\
        & tpn              & 0.3088  & 0.2962  & 0.3066  & 0.2342  & 0.2955  & 0.2883  \\
        & vat              & 0.5775  & 0.6416  & 0.6096  & 0.4838  & 0.5934  & 0.5812  \\
            \hline
        \multirow{7}*{CCC} 
        & Interpret-img    &  0.5487  & 0.6090  & 0.5778  & 0.4364  & 0.5402  & 0.5424 \\
        & PersEmoN         &  0.2481  & 0.2935  & 0.3110  & 0.1614  & 0.2246  & 0.2477 \\
        & senet            & 0.5112  & 0.5397  & 0.5392  & 0.4233  & 0.5478  & 0.5122 \\
        & hrnet            & 0.5187  & 0.5449  & 0.5476  & 0.4539  & 0.5700  & 0.5270  \\
        & swin-transformer & 0.2698  & 0.2796  & 0.2495  & 0.1457  & 0.2220  & 0.2333  \\
        & 3D-resnet        & 0.0004  & 0.0002  & -0.0001 & -0.0014 & -0.0003 & 0.0002  \\
        & slow-fast        & 0.0451  & 0.0371  & 0.0250  & 0.0195  & 0.0313  & 0.0316  \\
        & tpn              & 0.1532  & 0.1608  & 0.1350  & 0.0863  & 0.1328  & 0.1336  \\
        & vat              & 0.5559  & 0.6243  & 0.5881  & 0.4532  & 0.5667  & 0.5577  \\
			\bottomrule
		\end{tabular}
        }
	\end{center}
	\caption{Models trained on frames from ChaLearn2016 impressison dataset}  
\label{tb:BRC}
\end{table}
\setlength{\tabcolsep}{1.4pt}
% +++++++++++++++++++++++++++++++++++++++++++++++++++++++++++++++++++++++++++++++++++++++

Models are trained with face data
%% frame images
% +++++++++++++++++++++++++++++++++++++++++++++++++++++++++++++++++++++++++++++++++++
\setlength{\tabcolsep}{2pt}
\begin{table}[H]
	\begin{center}
\resizebox{1\linewidth}{!} {
		\begin{tabular}{|l| c| c c c c c l|}
			\toprule
            & Traits         & Open    & Consc  & Extrav   & Agree   & Neuro   & Avg.   \\
            \hline \hline

        \multirow{7}*{ACC} 
        & senet              & 0.9076 & 0.906  & 0.908  & 0.9097 & 0.9061 & 0.9075  \\
        & hrnet              & 0.9101 & 0.9154 & 0.9111 & 0.9113 & 0.9084 & 0.9113  \\
        & swin-transformer   & 0.8937 & 0.8870 & 0.8893 & 0.8983 & 0.8860 & 0.8909  \\
        & 3D-resnet          & 0.8964 & 0.8921 & 0.8933 & 0.9008 & 0.8915 & 0.8948  \\
        & slow-fast          & 0.878 & 0.8604 & 0.8443 & 0.8809 & 0.8613 & 0.8650  \\    
        & tpn                & 0.9025 & 0.8963 & 0.9019 & 0.9013 & 0.8992 & 0.9003  \\
        & vat                & 0.9115 & 0.9123 & 0.9153 & 0.9099 & 0.9098 & 0.9118  \\
            \hline
        \multirow{7}*{PCC} 
        & senet              & 0.6025 & 0.6335 & 0.6498 & 0.5342 & 0.6464 & 0.6133 \\
        & hrnet              & 0.6302 & 0.7158 & 0.6801 & 0.5659 & 0.6666 & 0.6517  \\
        & swin-transformer   & 0.3875 & 0.4021 & 0.4268 & 0.2786 & 0.3659 & 0.3722  \\
        & 3D-resnet          & 0.4356 & 0.4731 & 0.4743 & 0.3474 & 0.4619 & 0.4385 \\
        & slow-fast          & 0.1644 & 0.2097 & 0.1582 & 0.1072 & 0.1352 & 0.1549 \\    
        & tpn                & 0.5577 & 0.5819 & 0.5928 & 0.4497 & 0.5795 & 0.5523  \\
        & vat                & 0.6512 & 0.7077 & 0.7087 & 0.5597 & 0.6784 & 0.6612  \\
            \hline
        \multirow{7}*{CCC} 
        & senet              & 0.5300 & 0.5580 & 0.5815 & 0.4493 & 0.5708 & 0.5379 \\
        & hrnet              & 0.5923 & 0.6912 & 0.6436 & 0.5195 & 0.6273 & 0.6148  \\
        & swin-transformer   & 0.2223 & 0.2426 & 0.2531 & 0.1224 & 0.1942 & 0.2069  \\
        & 3D-resnet          & 0.3248 & 0.3601 & 0.3601 & 0.2120 & 0.3352 & 0.3185  \\
        & slow-fast          & 0.0256 & 0.0320 & 0.0185 & 0.0105 & 0.0184 & 0.0210  \\    
        & tpn                & 0.4427 & 0.4767 & 0.4998 & 0.3230 & 0.4675 & 0.4420  \\
        & vat                & 0.6216 & 0.6753 & 0.6836 & 0.5228 & 0.6456 & 0.6298  \\
			\bottomrule
		\end{tabular}
        }
	\end{center}
	\caption{Models trained on face images from ChaLearn2016 impressison dataset}  
\label{tb:BRC}
\end{table}
\setlength{\tabcolsep}{1.4pt}
% +++++++++++++++++++++++++++++++++++++++++++++++++++++++++++++++++++++++++++++++++++++++

\subsubsection{Models trained on audio-visual data from ChaLearn2016 impressison dataset}

% +++++++++++++++++++++++++++++++++++++++++++++++++++++++++++++++++++++++++++++++++++
\setlength{\tabcolsep}{2pt}
\begin{table}[H]
	\begin{center}
\resizebox{1\linewidth}{!} {
		\begin{tabular}{|l| c| c c c c c l|}
			\toprule
            & Traits         & Open    & Consc  & Extrav   & Agree   & Neuro   & Avg.   \\
            \hline \hline

        \multirow{4}*{ACC} 
                & Deep-bimodal reg   & 0.9098 & 0.9106 & 0.9096 & 0.9102 & 0.9061 & 0.9093 \\ % frame       without audio data(separated)
                & Bi-modal lstm      & 0.8833 & 0.8744 & 0.8779 & 0.8935 & 0.8773 & 0.8813 \\ % face        with audio data
                & Audiovisual-resnet & 0.8996 & 0.8918 & 0.8945 & 0.9015 & 0.8948 & 0.8964 \\ % frame       with audio data
                & CRNet              & 0.9075 & 0.9019 & 0.9017 & 0.9055 & 0.9034 & 0.9040 \\ % frame/face  with audio data
            \hline
        \multirow{4}*{PCC} 
                & Deep-bimodal reg   &  0.6210  &  0.6746  & 0.6536  & 0.5445  &  0.6494  & 0.6286 \\ % frame       without audio data(separated)
                & Bi-modal lstm      & -0.0065  & -0.0107  & 0.0062  & 0.0231  & -0.0216  & 0.0019 \\ % face        with audio data
                & Audiovisual-resnet &  0.5005  &  0.4720  & 0.4852  & 0.3811  &  0.5096  & 0.4697 \\ % frame       with audio data
                & CRNet              &  0.5932  &  0.5870  & 0.5705  & 0.4685  &  0.6058  & 0.5650 \\ % frame/face  with audio data
            \hline
        \multirow{4}*{CCC} 
                & Deep-bimodal reg   & 0.5693  & 0.6254  & 0.6070  & 0.4855  & 0.6025  & 0.5779  \\ % frame      without audio data
                & Bi-modal lstm      & 0.0000  & 0.0000  & 0.0000  & 0.0000  & 0.0000  & 0.0000  \\ % face       with audio data
                & Audiovisual-resnet & 0.4150  & 0.3671  & 0.3889  & 0.2679  & 0.4181  & 0.3714  \\ % frame      with audio data
                & CRNet              & 0.5193  & 0.5106  & 0.5024  & 0.4026  & 0.5119  & 0.4894  \\ % frame/face with audio data
			\bottomrule
		\end{tabular}
        }
	\end{center}
	\caption{Models trained on frames and audio data from talk session of true personality dataset}  
\label{tb:BRC}
\end{table}
\setlength{\tabcolsep}{1.4pt}
% +++++++++++++++++++++++++++++++++++++++++++++++++++++++++++++++++++++++++++++++++++++++

\subsubsection{Models trained on visual data with all video frames and face images in impression dataset}

%% frame images
% +++++++++++++++++++++++++++++++++++++++++++++++++++++++++++++++++++++++++++++++++++
\setlength{\tabcolsep}{2pt}
\begin{table}[H]
	\begin{center}
\resizebox{1\linewidth}{!} {
		\begin{tabular}{|l| c| c c c c c l|}
			\toprule
            & Traits         & Open    & Consc  & Extrav   & Agree   & Neuro   & Avg.   \\
            \hline \hline

        \multirow{5}*{ACC} 
        & 3D-ResNet          & 0.8830 & 0.8745 & 0.8781 & 0.8933 & 0.8771 & 0.8812 \\
        & Slow-fast          & 0.8773 & 0.8597 & 0.8398 & 0.8800 & 0.8598 & 0.8633 \\    
        & TPN                & 0.8890 & 0.8793 & 0.8833 & 0.8961 & 0.8827 & 0.8861 \\
        & VAT                & 0.9036 & 0.9066 & 0.9048 & 0.9042 & 0.8996 & 0.9038 \\
            \hline
        \multirow{5}*{PCC} 
        & 3D-ResNet          & 0.0024  & 0.0012  & -0.0036 & -0.0166 & -0.0034 & -0.004  \\
        & Slow-fast          & 0.3232  & 0.3275  & 0.2992  & 0.2248  & 0.2959  & 0.2941  \\
        & TPN                & 0.3088  & 0.2962  & 0.3066  & 0.2342  & 0.2955  & 0.2883  \\
        & VAT                & 0.5775  & 0.6416  & 0.6096  & 0.4838  & 0.5934  & 0.5812  \\
            \hline
        \multirow{5}*{CCC} 
        & 3D-ResNet          & 0.0004  & 0.0002  & -0.0001 & -0.0014 & -0.0003 & 0.0002  \\
        & Slow-fast          & 0.0451  & 0.0371  & 0.0250  & 0.0195  & 0.0313  & 0.0316  \\
        & TPN                & 0.1532  & 0.1608  & 0.1350  & 0.0863  & 0.1328  & 0.1336  \\
        & VAT                & 0.5559  & 0.6243  & 0.5881  & 0.4532  & 0.5667  & 0.5577  \\
			\bottomrule
		\end{tabular}
        }
	\end{center}
	\caption{Models trained on all video frames from ChaLearn2016 impressison dataset}  
\label{tb:BRC}
\end{table}
\setlength{\tabcolsep}{1.4pt}
% +++++++++++++++++++++++++++++++++++++++++++++++++++++++++++++++++++++++++++++++++++++++

%% face images
% +++++++++++++++++++++++++++++++++++++++++++++++++++++++++++++++++++++++++++++++++++
\setlength{\tabcolsep}{2pt}
\begin{table}[H]
	\begin{center}
\resizebox{1\linewidth}{!} {
		\begin{tabular}{|l| c| c c c c c l|}
			\toprule
            & Traits         & Open    & Consc  & Extrav   & Agree   & Neuro   & Avg.   \\
            \hline \hline

        \multirow{5}*{ACC} 
        & 3D-ResNet          & 0.8973 & 0.8926 & 0.8927 & 0.9010 & 0.8902 & 0.8948  \\
        & Slow-fast          & 0.8775 & 0.8608 & 0.8419 & 0.8807 & 0.8600 & 0.8642  \\    
        & TPN                & 0.8867 & 0.8785 & 0.8804 & 0.8933 & 0.8797 & 0.8837 \\
        & VAT                & 0.9035 & 0.9104 & 0.9079 & 0.9061 & 0.9029 & 0.9062  \\
            \hline
        \multirow{5}*{PCC} 
        & 3D-ResNet          & 0.4497 & 0.4728 & 0.4747 & 0.3532 & 0.4561 & 0.4413  \\
        & Slow-fast          & 0.1884 & 0.2500 & 0.1933 & 0.1431 & 0.1808 & 0.1911  \\    
        & TPN                & 0.2231 & 0.2457 & 0.2084 & 0.1144 & 0.1966 & 0.1977 \\
        & VAT                & 0.5952 & 0.6826 & 0.6586 & 0.5152 & 0.6261 & 0.6155 \\
            \hline
        \multirow{5}*{CCC} 
        & 3D-ResNet          & 0.3555 & 0.3811 & 0.3815 & 0.2294 & 0.3514 & 0.3398  \\
        & Slow-fast          & 0.0268 & 0.0390 & 0.0201 & 0.0155 & 0.0232 & 0.0249  \\    
        & TPN                & 0.1227 & 0.1235 & 0.1097 & 0.0379 & 0.0823 & 0.0952  \\
        & VAT                & 0.5818 & 0.6703 & 0.6457 & 0.4956 & 0.6096 & 0.6006  \\
			\bottomrule
		\end{tabular}
        }
	\end{center}
	\caption{Models trained on all face images in frames from ChaLearn2016 impressison dataset}  
\label{tb:BRC}
\end{table}
\setlength{\tabcolsep}{1.4pt}

\subsubsection{Models trained on visual data with video level frames and face images segment in impression dataset}

%% frame images
% +++++++++++++++++++++++++++++++++++++++++++++++++++++++++++++++++++++++++++++++++++
\setlength{\tabcolsep}{2pt}
\begin{table}[H]
	\begin{center}
\resizebox{1\linewidth}{!} {
		\begin{tabular}{|l| c| c c c c c l|}
			\toprule
            & Traits         & Open    & Consc  & Extrav   & Agree   & Neuro   & Avg.   \\
            \hline \hline

        \multirow{5}*{ACC} 
        & 3D-ResNet          & 0.8878 & 0.8806 & 0.8814 & 0.8954 & 0.8809 & 0.8852 \\
        & Slow-fast          & 0.8767 & 0.8602 & 0.8374 & 0.8803 & 0.8595 & 0.8628  \\    
        & TPN                & 0.8842 & 0.8687 & 0.8736 & 0.8910 & 0.8265 & 0.8688 \\
        & VAT                & 0.8937 & 0.8879 & 0.8896 & 0.8989 & 0.8866 & 0.8913 \\
            \hline
        \multirow{5}*{PCC} 
        & 3D-ResNet          & 0.2717 & 0.2797 & 0.2625 & 0.1837 & 0.2474 & 0.2490  \\
        & Slow-fast          & 0.1507 & 0.1488 & 0.1610 & 0.0815 & 0.1126 & 0.1309  \\
        & TPN                & 0.1781 & 0.1380 & 0.0708 & 0.0311 & 0.0380 & 0.0912  \\
        & VAT                & 0.3929 & 0.4223 & 0.4135 & 0.2942 & 0.3861 & 0.3818  \\
            \hline
        \multirow{5}*{CCC} 
        & 3D-ResNet          & 0.1716 & 0.1674 & 0.1593 & 0.0870 & 0.1408 & 0.1452 \\
        & Slow-fast          & 0.0195 & 0.0273 & 0.0120 & 0.0106 & 0.0140 & 0.0167  \\
        & TPN                & 0.1081 & 0.1192 & 0.0440 & 0.0136 & 0.0187 & 0.0607 \\
        & VAT                & 0.2836 & 0.3100 & 0.2994 & 0.1728 & 0.2708 & 0.2673 \\
			\bottomrule
		\end{tabular}
        }
	\end{center}
	\caption{Models trained on video-level frames segment from ChaLearn2016 impressison dataset}  
\label{tb:BRC}
\end{table}
\setlength{\tabcolsep}{1.4pt}
% +++++++++++++++++++++++++++++++++++++++++++++++++++++++++++++++++++++++++++++++++++++++

%% face images
% +++++++++++++++++++++++++++++++++++++++++++++++++++++++++++++++++++++++++++++++++++
\setlength{\tabcolsep}{2pt}
\begin{table}[H]
	\begin{center}
\resizebox{1\linewidth}{!} {
		\begin{tabular}{|l| c| c c c c c l|}
			\toprule
            & Traits         & Open    & Consc  & Extrav   & Agree   & Neuro   & Avg.   \\
            \hline \hline

        \multirow{5}*{ACC} 
        & 3D-ResNet          & 0.8880 & 0.8801 & 0.8830 & 0.8952 & 0.8811 & 0.8855  \\
        & Slow-fast          & 0.8805 & 0.8653 & 0.8432 & 0.8843 & 0.8657 & 0.8678  \\    
        & TPN                & 0.8731 & 0.8114 & 0.8165 & 0.7872 & 0.8632 & 0.8303 \\
        & VAT                & 0.8867 & 0.8809 & 0.8821 & 0.8946 & 0.8807 & 0.8850  \\
            \hline
        \multirow{5}*{PCC} 
        & 3D-ResNet          & 0.2819 & 0.2671 & 0.3004 & 0.1769 & 0.2671 & 0.2587 \\
        & Slow-fast          & 0.2408 & 0.2887 & 0.2780 & 0.1661 & 0.2379 & 0.2423   \\    
        & TPN                & 0.0800 & -0.0238 & 0.0215 & -0.0013 & 0.0775 & 0.0308 \\
        & VAT                & 0.2582 & 0.2923 & 0.2844 & 0.1763 & 0.2661 & 0.2555 \\
            \hline
        \multirow{5}*{CCC} 
        & 3D-ResNet          & 0.1171 & 0.1207 & 0.1199 & 0.0387 & 0.0954 & 0.0983  \\
        & Slow-fast          & 0.0318 & 0.0463 & 0.0255 & 0.0176 & 0.0297 & 0.0302  \\    
        & TPN                & 0.0709 & -0.0188 & 0.0166 & -0.0007 & 0.0654 & 0.0267  \\
        & VAT                & 0.1525 & 0.1794 & 0.1727 & 0.0783 & 0.1457 & 0.1457  \\
			\bottomrule
		\end{tabular}
        }
	\end{center}
	\caption{Models trained on video-level face images segment from ChaLearn2016 impressison dataset}  
\label{tb:BRC}
\end{table}
\setlength{\tabcolsep}{1.4pt}

\subsection{Results of different sessions in True personality}

\subsubsection{talk session with audio data}

%% audio modality
% +++++++++++++++++++++++++++++++++++++++++++++++++++++++++++++++++++++++++++++++++++
\setlength{\tabcolsep}{2pt}
\begin{table}[H]
	\begin{center}
\resizebox{1\linewidth}{!} {
		\begin{tabular}{|l| c| c c c c c l|}
			\toprule
            & Traits         & Open    & Consc  & Extrav   & Agree   & Neuro   & Avg.   \\
            \hline \hline
        \multirow{5}*{MSE} 
        & FFT           & 0.8831 & 0.7001 & 1.4395 & 0.9288 & 1.2765 & 1.0456  \\
        & MFCC/logfbank & 1.0168 & 0.8975 & 1.9919 & 1.2612 & 1.3302 & 1.2995 \\
        & Bi-modal CNN  & 0.9521 & 0.7209 & 1.2583 & 0.8814 & 1.0641 & 0.9754  \\
        & ResNet        & 1.0537 & 0.7124 & 1.2863 & 0.8589 & 1.0631 & 0.9949 \\
        & CR-Netaud     & 0.9735 & 0.8358 & 1.7828 & 1.1014 & 1.3561 & 1.2099   \\
            \hline
        \multirow{5}*{PCC} 
        & FFT           &  0.0595 & -0.3029 & -0.2013 & -0.1759 &  0.4358  & -0.0370  \\
        & MFCC/logfbank & -0.3221 & -0.0992 &  0.2363 &  0.3322 & -0.0129  &  0.0269 \\
        & Bi-modal CNN  & -0.1849 & -0.1889 &  0.2197 & -0.0429 & -0.2558  & -0.0906 \\
        & ResNet        & -0.0759 &  0.1468 &  0.1576 &  0.2281 &  0.3399 &   0.1593 \\
        & CR-Netaud     &  0.0365 &  0.1189 &  0.1967 &  0.1124 & -0.2652 &   0.0399  \\
            \hline
        \multirow{5}*{CCC} 
        & FFT           &  0.0000 &  0.0000 & 0.0000 & 0.0000 &  0.0000  &  0.0000  \\
        & MFCC/logfbank & -0.0009 & -0.0003 & 0.0003 & 0.0013 & -0.0001  &  0.0001  \\
        & Bi-modal CNN  & -0.0002 & -0.0004 & 0.0001 & 0.0000 & -0.0001  & -0.0001  \\
        & ResNet        & -0.0377 &  0.0518 & 0.0321 & 0.0832 &  0.0008  &  0.0260  \\
        & CR-Netaud     &  0.0199 &  0.0696 & 0.0594 & 0.0404 & -0.0875  &  0.0204  \\
			\bottomrule
		\end{tabular}
        }
	\end{center}
	\caption{Models trained on audio data from talk session of true personality dataset}  
\label{tb:BRC}
\end{table}
\setlength{\tabcolsep}{1.4pt}
% +++++++++++++++++++++++++++++++++++++++++++++++++++++++++++++++++++++++++++++++++++++++

\subsubsection{talk session with visual data}

Models are trained with frame data
%% frame images
% +++++++++++++++++++++++++++++++++++++++++++++++++++++++++++++++++++++++++++++++++++
\setlength{\tabcolsep}{2pt}
\begin{table}[H]
	\begin{center}
\resizebox{1\linewidth}{!} {
		\begin{tabular}{|l| c| c c c c c l|}
			\toprule
            & Traits         & Open    & Consc  & Extrav   & Agree   & Neuro   & Avg.   \\
            \hline \hline

        \multirow{7}*{MSE} 
        % & deep-bimodal reg   & 0.5693  & 0.6254  & 0.6070  & 0.4855  & 0.6025  & 0.5779  \\ % only frame no audio 
        & Interpret-img    & 0.9946 & 0.5655 & 1.0808 & 0.9042 & 1.4284 & 0.9947  \\
        & PersEmoN         & 0.8434 & 0.7140 & 1.3382 & 0.9323 & 1.1517 & 0.9959  \\
        & senet            & 1.2619 & 0.6401 & 1.1722 & 0.8547 & 1.4438 & 1.0744 \\
        & hrnet            & 1.1184 & 0.8014 & 1.3216 & 0.7942 & 1.2672 & 1.0606 \\
        & swin-transformer & 1.5054 & 0.6485 & 1.2192 & 0.9950 & 1.3779 & 1.1492 \\
        & 3D-resnet        & 0.8873 & 0.6849 & 1.3178 & 0.9067 & 1.2540 & 1.0101 \\
        & slow-fast        & 0.8949 & 0.8505 & 1.6881 & 1.0978 & 1.3914 & 1.1845 \\
        & tpn              & 0.8802 & 0.6512 & 1.2357 & 0.9094 & 1.1906 & 0.9734 \\
        & vat              & 0.8460 & 0.6723 & 1.2092 & 0.9097 & 1.2590 & 0.9792 \\
            \hline
        \multirow{7}*{PCC} 
        % & deep-bimodal reg   & 0.5693  & 0.6254  & 0.6070  & 0.4855  & 0.6025  & 0.5779  \\ % only frame no audio 
        & Interpret-img    &  0.1861 & 0.5551 &  0.4156 &  0.1984 & -0.0725 &  0.2566  \\
        & PersEmoN         &  0.2684 & 0.1111 & -0.0586 &  0.1143 & -0.0096 &  0.0851 \\
        & senet            &  0.0409 & 0.4306 &  0.1593 &  0.3224 & -0.1490 &  0.1608  \\
        & hrnet            &  0.1838 & 0.2315 & -0.0077 &  0.4094 &  0.1127 &  0.1860  \\
        & swin-transformer & -0.0622 & 0.3491 & -0.0171 &  0.2121 &  0.0725 &  0.1109  \\
        & 3D-resnet        &  0.0368 & 0.0178 &  0.3413 &  0.1808 & -0.2473 &  0.0659  \\
        & slow-fast        & -0.2386 & 0.0246 & -0.2742 & -0.1906 & -0.0454 & -0.1448  \\
        & tpn              &  0.1252 & 0.1614 &  0.3834 &  0.0282 &  0.0906 &  0.1578  \\
        & vat              &  0.2565 & 0.1037 &  0.5175 &  0.1920 & -0.0514 &  0.2037  \\
            \hline
        \multirow{7}*{CCC} 
        % & deep-bimodal reg   & 0.9098 & 0.9106 & 0.9096 & 0.9102 & 0.9061 & 0.9093 \\ 
        & Interpret-img    &  0.1420 & 0.3632 &  0.1736 &  0.1403 & -0.0503 &  0.1538 \\
        & PersEmoN         &  0.0928 & 0.0338 & -0.0004 &  0.0060 &  0.0000 &  0.0264  \\
        & senet            &  0.0235 & 0.3886 &  0.0743 &  0.2551 & -0.0780 &  0.1327  \\
        & hrnet            &  0.1517 & 0.2047 & -0.0037 &  0.3591 &  0.0792 &  0.1582  \\
        & swin-transformer & -0.0494 & 0.3169 & -0.0088 &  0.1815 &  0.0541 &  0.0989  \\
        & 3D-resnet        &  0.0043 & 0.0014 &  0.0028 &  0.0037 & -0.0025 &  0.0019  \\
        & slow-fast        & -0.0080 & 0.0010 & -0.0159 & -0.0047 & -0.0021 & -0.0059 \\
        & tpn              &  0.0427 & 0.0888 &  0.0484 &  0.0071 &  0.0282 &  0.0430   \\
        & vat              &  0.0642 & 0.0165 &  0.0618 &  0.0058 & -0.0029 &  0.0291  \\
			\bottomrule
		\end{tabular}
        }
	\end{center}
	\caption{Models trained on frames from talk session of true personality dataset}  
\label{tb:BRC}
\end{table}
\setlength{\tabcolsep}{1.4pt}
% +++++++++++++++++++++++++++++++++++++++++++++++++++++++++++++++++++++++++++++++++++++++

Models are trained with face data
%% face images
% +++++++++++++++++++++++++++++++++++++++++++++++++++++++++++++++++++++++++++++++++++
\setlength{\tabcolsep}{2pt}
\begin{table}[H]
	\begin{center}
\resizebox{1\linewidth}{!} {
		\begin{tabular}{|l| c| c c c c c l|}
			\toprule
            & Traits         & Open    & Consc  & Extrav   & Agree   & Neuro   & Avg.   \\
            \hline \hline

        \multirow{7}*{MSE} 
        % & deep-bimodal reg   & 0.5693  & 0.6254  & 0.6070  & 0.4855  & 0.6025  & 0.5779  \\ % only frame no audio 
        & Interpret-img    & 1.0126 &   0.5258 &   1.5769 &   0.9724 &   1.1863  &  1.0548 \\
        & PersEmoN         & 0.8924 &   0.6873 &   1.3869 &   0.9388 &   1.1459  &  1.0102  \\
        & senet            & 1.0953 & 0.6715 & 1.4034 & 1.2109 & 1.0157 & 1.0794 \\
        & hrnet            & 1.1065 & 0.6259 & 1.7961 & 0.8482 & 1.2149 & 1.1183 \\
        & swin-transformer & 1.328  & 0.7448 & 1.464 & 1.3229 & 1.2486 & 1.2217  \\
        & 3D-resnet        & 2.2172 & 0.6734 & 1.6037 & 0.9301 & 1.6791 & 1.4206 \\
        & slow-fast        & 1.0743 & 1.0902 & 1.771 & 1.1972 & 1.3242 & 1.2914  \\
        & tpn              & 0.9252 & 0.8072 & 1.1932 & 0.8594 & 1.3149 & 1.02 \\
        & vat              & 0.9076 & 0.6972 & 1.2035 & 0.8925 & 1.2808 & 0.9963  \\
            \hline
        \multirow{7}*{PCC} 
        % & deep-bimodal reg   & 0.5693  & 0.6254  & 0.6070  & 0.4855  & 0.6025  & 0.5779  \\ % only frame no audio
        & Interpret-img    &-0.0072 &   0.5193 &   0.0739 &   0.0673 &   0.4532  &  0.2213 \\
        & PersEmoN         & 0.0497 &   0.1897 &   0.3277 &   0.1575 &   0.2273  &  0.1904 \\
        & senet            & 0.1165 & 0.1975 & 0.2389 & -0.1537 & 0.4485 & 0.1695  \\
        & hrnet            & 0.2237 & 0.4239 & 0.1647 & 0.3573 & 0.2961 & 0.2931  \\
        & swin-transformer & -0.0729 & 0.2214 & 0.2122 & -0.0666 & 0.4054 & 0.1399  \\
        & 3D-resnet        & -0.1164 & 0.2039 & 0.1566 & 0.2199 & -0.2712 & 0.0385  \\
        & slow-fast        & -0.3144 & 0.2742 & 0.2576 & -0.1381 & 0.2115 & 0.0582  \\
        & tpn              & -0.2628 & -0.1833 & -0.0988 & 0.0934 & -0.0347 & -0.0973  \\
        & vat              & -0.2386 & -0.1742 & 0.1149 & -0.0524 & 0.0066 & -0.0687 \\
            \hline
        \multirow{7}*{CCC} 
        % & deep-bimodal reg   & 0.9098 & 0.9106 & 0.9096 & 0.9102 & 0.9061 & 0.9093 \\ 
        & Interpret-img    & -0.0045 &   0.3783 &   0.0281 &   0.0402 &   0.2943  &  0.1473 \\
        & PersEmoN         &  0.006 &   0.0275 &   0.0125 &   0.0083 &   0.0062  &  0.0121  \\
        & senet            & 0.0737 & 0.1403 & 0.13 & -0.0891 & 0.3505 & 0.1211  \\
        & hrnet            & 0.1735 & 0.349 & 0.0785 & 0.2534 & 0.202 & 0.2113  \\
        & swin-transformer & -0.063 & 0.1885 & 0.1271 & -0.0424 & 0.3312 & 0.1083  \\
        & 3D-resnet        & -0.1089 & 0.0658 & 0.0875  & 0.1840 & -0.1289 & 0.0199  \\
        & slow-fast        & -0.0332 & 0.0167 & 0.007 & -0.0084 & 0.0197 & 0.0003 \\
        & tpn              & 0.0427 & 0.0888 & 0.0484 & 0.0071 & 0.0282 & 0.0430   \\
        & vat              & -0.0259 & -0.0137 & 0.0038 & -0.003 & 0.0009 & -0.0076  \\
			\bottomrule
		\end{tabular}
        }
	\end{center}
	\caption{Widely used models trained on face images from talk session of true personality dataset}  
\label{tb:BRC}
\end{table}
\setlength{\tabcolsep}{1.4pt}
% +++++++++++++++++++++++++++++++++++++++++++++++++++++++++++++++++++++++++++++++++++++++

\subsubsection{talk session with video-level visual data}

Models are trained with frame data
%% frame images
% +++++++++++++++++++++++++++++++++++++++++++++++++++++++++++++++++++++++++++++++++++
\setlength{\tabcolsep}{2pt}
\begin{table}[H]
	\begin{center}
\resizebox{1\linewidth}{!} {
		\begin{tabular}{|l| c| c c c c c l|}
			\toprule
            & Traits         & Open    & Consc  & Extrav   & Agree   & Neuro   & Avg.   \\
            \hline \hline

        \multirow{5}*{MSE} 
        & 3D-resnet        & 0.8903 &   0.693 &   1.2789 &   1.1429 &   1.2374  &  1.0485 \\
        & slow-fast        & 0.9951 &   0.7288 &   1.8139 &   1.141 &   1.502  &  1.2361 \\
        & tpn              & 0.8823 &   0.7315 &   1.2743 &   0.8941 &   1.2402  &  1.0045 \\
        & vat              & 0.8814 &   0.6779 &   1.2736 &   0.916 &   1.2497  &  0.9997 \\
            \hline
        \multirow{5}*{PCC} 
        & 3D-resnet        & 0.3686 &   0.2124 &   -0.2108 &   0.1362 &   -0.0035  &  0.1006  \\
        & slow-fast        & -0.2506 &   -0.0039 &   -0.1271 &   -0.0332 &   0.0692  &  -0.0691  \\
        & tpn              & 0.113 &   -0.0148 &   -0.0107 &   0.1335 &   -0.1304  &  0.0181  \\
        & vat              & 0.0945 &   -0.1459 &   0.3484 &   0.0262 &   -0.0861  &  0.0474  \\
            \hline
        \multirow{5}*{CCC} 
        & 3D-resnet        & 0.1476 &   0.0489 &   -0.0134 &   0.0196 &   -0.0016  &  0.0402 \\
        & slow-fast        & -0.014 &   -0.0008 &   -0.0067 &   -0.002 &   0.0058  &  -0.0035 \\
        & tpn              & 0.0316 &   -0.0077 &   -0.0008 &   0.0243 &   -0.0445  &  0.0006   \\
        & vat              & 0.0037 &   -0.004 &   0.0106 &   0.0007 &   -0.0019  &  0.0018  \\
			\bottomrule
		\end{tabular}
        }
	\end{center}
	\caption{Models trained on frames from talk session of true personality dataset}  
\label{tb:BRC}
\end{table}
\setlength{\tabcolsep}{1.4pt}
% +++++++++++++++++++++++++++++++++++++++++++++++++++++++++++++++++++++++++++++++++++++++

Models are trained with face data
%% face images
% +++++++++++++++++++++++++++++++++++++++++++++++++++++++++++++++++++++++++++++++++++
\setlength{\tabcolsep}{2pt}
\begin{table}[H]
	\begin{center}
\resizebox{1\linewidth}{!} {
		\begin{tabular}{|l| c| c c c c c l|}
			\toprule
            & Traits         & Open    & Consc  & Extrav   & Agree   & Neuro   & Avg.   \\
            \hline \hline

        \multirow{5}*{MSE} 
        & 3D-resnet        & 1.129 &   1.038 &   0.7078 &   3.6307 &   1.903  &  1.6817 \\
        & slow-fast        & 0.9681 &   0.8847 &   1.8038 &   1.2567 &   1.2665  &  1.2360 \\
        & tpn              & 1.1651 &   1.3543 &   0.976 &   1.1803 &   1.3715  &  1.2094 \\
        & vat              & 0.8929 &   0.673 &   1.2937 &   0.9147 &   1.2529  &  1.0054 \\
            \hline
        \multirow{5}*{PCC} 
        & 3D-resnet        & 0.0954 &   0.0189 &   0.5887 &   0.1439 &   -0.2265  &  0.1241 \\
        & slow-fast        & -0.1069 &   0.1144 &   0.3656 &   -0.1735 &   -0.064  &  0.0271  \\
        & tpn              & -0.1615 &   -0.1533 &   0.5331 &   -0.2537 &   -0.1629  &  -0.0397  \\
        & vat              & -0.1846 &   0.1697 &   -0.2765 &   0.0015 &   0.0324  &  -0.0515  \\
            \hline
        \multirow{5}*{CCC} 
        & 3D-resnet        & 0.0468 &   0.002 &   0.4706 &   0.089 &   -0.1329  &  0.0951  \\
        & slow-fast        & -0.0077 &   0.015 &   0.0246 &   -0.0043 &   -0.0042  &  0.0047 \\
        & tpn              & -0.1159 &   -0.1285 &   0.1089 &   -0.1593 &   -0.0557  &  -0.0701   \\
        & vat              & -0.0031 &   0.0073 &   -0.0082 &   0.0 &   0.0019  &  -0.0004 \\
			\bottomrule
		\end{tabular}
        }
	\end{center}
	\caption{Widely used models trained on face images from talk session of true personality dataset}  
\label{tb:BRC}
\end{table}
\setlength{\tabcolsep}{1.4pt}
% +++++++++++++++++++++++++++++++++++++++++++++++++++++++++++++++++++++++++++++++++++++++

\subsubsection{talk session with audio-visual data}
Models are trained with frame and audio data
% +++++++++++++++++++++++++++++++++++++++++++++++++++++++++++++++++++++++++++++++++++
\setlength{\tabcolsep}{2pt}
\begin{table}[H]
	\begin{center}
\resizebox{1\linewidth}{!} {
		\begin{tabular}{|l| c| c c c c c l|}
			\toprule
            & Traits         & Open    & Consc  & Extrav   & Agree   & Neuro   & Avg.   \\
            \hline \hline

        \multirow{4}*{MSE} 
                & Deep-bimodal reg    & 0.8471 &   0.5731 &   1.3973 &   1.0411 &   1.4111  &  1.0539  \\
                & Bi-modal lstm       & 0.9178 &   0.7203 &   1.2452 &   0.9072 &   1.3615  &  1.0304   \\
                & Audiovisual-resent  & 1.0075 &   0.5865 &   1.4741 &   0.8976 &   1.1508  &  1.0233   \\
                & CRNet               & 1.2635 &   0.5873 &   1.1236 &   1.1887 &   1.5334  &  1.1393 \\
            \hline
        \multirow{4}*{PCC} 
                & Deep-bimodal reg    & 0.1918 &   0.4475 &   0.4077 &   0.0552 &   -0.25  &  0.1704   \\
                & Bi-modal lstm       & -0.1975 &   -0.4218 &   0.3819 &   0.3025 &   -0.0817  &  -0.0033   \\
                & Audiovisual-resent  & 0.0736 &   0.5089 &   0.1403 &   0.2711 &   0.1133  &  0.2214   \\
                & CRNet               & -0.0168 &   0.4496 &   0.2408 &   -0.0016 &   0.0143  &  0.1373\\
            \hline
        \multirow{4}*{CCC} 
                & Deep-bimodal reg    & 0.0782 &   0.2556 &   0.1187 &   0.0266 &   -0.1124  &  0.0733   \\
                & Bi-modal lstm       & -0.0009 &   -0.0016 &   0.0003 &   0.0002 &   -0.0004  &  -0.0005  \\
                & Audiovisual-resent  & 0.0522 &   0.3184 &   0.0404 &   0.0996 &   0.0001  &  0.1021     \\
                & CRNet               & -0.0126 &   0.3825 &   0.1482 &   -0.0014 &   0.01  &  0.1053 \\
			\bottomrule
		\end{tabular}
        }
	\end{center}
	\caption{Models trained on frames and audio data from talk session of true personality dataset}  
\label{tb:BRC}
\end{table}
\setlength{\tabcolsep}{1.4pt}
% +++++++++++++++++++++++++++++++++++++++++++++++++++++++++++++++++++++++++++++++++++++++

Models are trained with face images and audio data
% +++++++++++++++++++++++++++++++++++++++++++++++++++++++++++++++++++++++++++++++++++
\setlength{\tabcolsep}{2pt}
\begin{table}[H]
	\begin{center}
\resizebox{1\linewidth}{!} {
		\begin{tabular}{|l| c| c c c c c l|}
			\toprule
            & Traits         & Open    & Consc  & Extrav   & Agree   & Neuro   & Avg.   \\
            \hline \hline

        \multirow{4}*{MSE} 
                & Deep-bimodal reg    & 0.9858 &   0.7916 &   2.3313 &   1.0678 &   1.0717  &  1.2496 \\
                & Bi-modal lstm       & 0.9141 &   0.7053 &   1.2407 &   0.913 &   1.3639  &  1.0274  \\
                & Audiovisual-resent  &  1.3495 &   0.9339 &   2.2135 &   1.1284 &   1.1508  &  1.3552   \\
                & CRNet               & 1.2635 &   0.5873 &   1.1236 &   1.1887 &   1.5334  &  1.1393 \\
            \hline
        \multirow{4}*{PCC} 
                & Deep-bimodal reg    &  0.1267 &   0.1168 &   -0.2172 &   -0.2648 &   0.3714  &  0.0266  \\
                & Bi-modal lstm       & -0.1785 &   -0.2531 &   0.4503 &   -0.0926 &   -0.0557  &  -0.0259  \\
                & Audiovisual-resent  &-0.0836 &   0.2517 &   -0.068 &   0.2314 &   0.1059  &  0.0875   \\
                & CRNet               &-0.0168 &   0.4496 &   0.2408 &   -0.0016 &   0.0143  &  0.1373\\
            \hline
        \multirow{4}*{CCC} 
                & Deep-bimodal reg    &  0.0825 &   0.0714 &   -0.0678 &   -0.0828 &   0.1925  &  0.0392 \\
                & Bi-modal lstm       & -0.0007 &   -0.0006 &   0.0005 &   -0.0 &   -0.0002  &  -0.0002 \\
                & Audiovisual-resent  & -0.0407 &   0.1041 &   -0.0197 &   0.0886 &   0.0001  &  0.0265  \\
                & CRNet               & -0.0126 &   0.3825 &   0.1482 &   -0.0014 &   0.01  &  0.1053 \\
			\bottomrule
		\end{tabular}
        }
	\end{center}
	\caption{Models trained on face images and audio data from talk session of true personality dataset}  
\label{tb:BRC}
\end{table}
\setlength{\tabcolsep}{1.4pt}
% +++++++++++++++++++++++++++++++++++++++++++++++++++++++++++++++++++++++++++++++++++++++

% ================================== ANIMAL ================================================
\subsubsection{animal session with audio data}
%% audio modality
% +++++++++++++++++++++++++++++++++++++++++++++++++++++++++++++++++++++++++++++++++++
\setlength{\tabcolsep}{2pt}
\begin{table}[H]
	\begin{center}
\resizebox{1\linewidth}{!} {
		\begin{tabular}{|l| c| c c c c c l|}
			\toprule
            & Traits         & Open    & Consc  & Extrav   & Agree   & Neuro   & Avg.   \\
            \hline \hline

        \multirow{5}*{MSE} 
        & FFT            & 1.2223 & 0.704  & 1.2785 & 0.9243 & 1.2125 & 1.0683  \\
        & MFCC/logfbank  & 0.9256 & 1.0708 & 2.0167 & 1.2297 & 1.3612 & 1.3200 \\
        & Bi-modal CNN   & 0.9203 & 0.8304 & 1.2412 & 0.8406 & 1.2341 & 1.0133 \\
        & Res18-aud      & 0.9012 & 0.6803 & 1.2184 & 0.8450 & 1.0785 & 0.9447 \\
        & crnet-aud      & 0.8811 & 0.6600 & 1.4578 & 0.9603 & 1.1651 & 1.0249   \\
            \hline
        \multirow{5}*{PCC} 
        & FFT            & 0.0321 & 0.3736  & -0.222 & 0.127   & -0.2156 &  0.0190  \\
        & MFCC/logfbank  & 0.1793 & -0.2337 & 0.3521 & -0.3065 & 0.1686  &  0.0320\\
        & Bi-modal CNN   &-0.0303 & -0.2978 & 0.3284 & 0.1797  & -0.2914 &  -0.0223  \\
        & Res18-aud      & 0.0874 & 0.3425  & 0.4417 & 0.2829  & 0.3786  &  0.3066  \\
        & crnet-aud      & 0.0365 & 0.1189  & 0.1967 & 0.1124  & -0.2652 &  0.0399   \\
            \hline
        \multirow{5}*{CCC} 
        & FFT            & 0.0000   & 0.0002  & -0.0000 & 0.0000  & -0.0000 &  0.0000  \\
        & MFCC/logfbank  &  0.0014  & -0.0012 & 0.0007  & -0.0011 & 0.0006  &  0.0001 \\
        & Bi-modal CNN   & -0.0000  & -0.0000 & 0.0000  & 0.0000  & -0.0000 &  0.0000  \\
        & Res18-aud      &  -0.0748 & 0.1794  & -0.3422 & 0.0235  & 0.0026  &  -0.0423  \\
        & crnet-aud      &  -0.0006 & 0.0017  & -0.0018 & 0.0002  & 0.0000  &  -0.0001  \\
			\bottomrule
		\end{tabular}
        }
	\end{center}
	\caption{Models trained on audio data from animal session of true personality dataset}  
\label{tb:BRC}
\end{table}
\setlength{\tabcolsep}{1.4pt}
% +++++++++++++++++++++++++++++++++++++++++++++++++++++++++++++++++++++++++++++++++++++++

\subsubsection{animal session with visual data}

\subsubsection{animal session with video-level visual data}

Models are trained with frame data
%% frame images
% +++++++++++++++++++++++++++++++++++++++++++++++++++++++++++++++++++++++++++++++++++
\setlength{\tabcolsep}{2pt}
\begin{table}[H]
	\begin{center}
\resizebox{1\linewidth}{!} {
		\begin{tabular}{|l| c| c c c c c l|}
			\toprule
            & Traits         & Open    & Consc  & Extrav   & Agree   & Neuro   & Avg.   \\
            \hline \hline

        \multirow{5}*{MSE} 
        & 3D-resnet        & 4.6234 &   5.1403 &   1.4011 &   1.8095 &   3.8527  &  3.3654 \\
        & slow-fast        & 1.1325 &   0.8418 &   2.2351 &   1.148 &   1.4781  &  1.3671 \\
        & tpn              &  1.0882 &   0.6632 &   1.4653 &   0.8994 &   1.2269  &  1.0686 \\
        & vat              & 0.8185 &   0.6544 &   1.1896 &   0.8953 &   1.2285  &  0.9573\\
            \hline
        \multirow{5}*{PCC} 
        & 3D-resnet        & -0.0195 &   0.0527 &   0.1183 &   -0.032 &   0.1541  &  0.0547  \\
        & slow-fast        & -0.1067 &   -0.3351 &   -0.2401 &   -0.1976 &   0.0419  &  -0.1675  \\
        & tpn              & -0.2352 &   0.2179 &   -0.118 &   0.1287 &   -0.1375  &  -0.0288  \\
        & vat              & 0.4658 &   0.1915 &   0.4136 &   0.15 &   0.0294  &  0.2501  \\
            \hline
        \multirow{5}*{CCC} 
        & 3D-resnet        & -0.0143 &   0.0265 &   0.0243 &   -0.0314 &   0.1077  &  0.0226 \\
        & slow-fast        & -0.0081 &   -0.0375 &   -0.0106 &   -0.0079 &   0.0017  &  -0.0125 \\
        & tpn              &  -0.0708 &   0.0789 &   -0.0148 &   0.0565 &   -0.038  &  0.0024   \\
        &5vat              & 0.0795 &   0.0191 &   0.0491 &   0.0117 &   0.0034  &  0.0326  \\
			\bottomrule
		\end{tabular}
        }
	\end{center}
	\caption{Models trained on frames from animal session of true personality dataset}  
\label{tb:BRC}
\end{table}
\setlength{\tabcolsep}{1.4pt}
% +++++++++++++++++++++++++++++++++++++++++++++++++++++++++++++++++++++++++++++++++++++++

Models are trained with face data
%% face images
% +++++++++++++++++++++++++++++++++++++++++++++++++++++++++++++++++++++++++++++++++++
\setlength{\tabcolsep}{2pt}
\begin{table}[H]
	\begin{center}
\resizebox{1\linewidth}{!} {
		\begin{tabular}{|l| c| c c c c c l|}
			\toprule
            & Traits         & Open    & Consc  & Extrav   & Agree   & Neuro   & Avg.   \\
            \hline \hline

        \multirow{5}*{MSE} 
        & 3D-resnet        & 12.1445 &   1.0832 &   3.6685 &   1.4765 &   5.9968  &  4.8739 \\
        & slow-fast        & 1.0911 &   0.7563 &   2.2116 &   1.1413 &   1.4216  &  1.3244 \\
        & tpn              & 1.2322 &   0.8498 &   1.2304 &   0.9942 &   2.206  &  1.3025 \\
        & vat              & 0.9021 &   0.6679 &   1.2988 &   0.9709 &   1.3448  &  1.0369 \\
            \hline
        \multirow{5}*{PCC} 
        & 3D-resnet        & -0.3934 &   0.1301 &   -0.2065 &   -0.114 &   -0.0778  &  -0.1323  \\
        & slow-fast        & -0.103 &   0.195 &   -0.0027 &   -0.0279 &   0.1514  &  0.0426  \\
        & tpn              & -0.117 &   -0.1794 &   0.0888 &   -0.2696 &   -0.2025  &  -0.1359  \\
        & vat              & -0.0432 &   0.1563 &   -0.0911 &   -0.1568 &   0.0116  &  -0.0246 \\
            \hline
        \multirow{5}*{CCC} 
        & 3D-resnet        & -0.2199 &   0.1186 &   -0.1899 &   -0.0999 &   -0.0599  &  -0.0902 \\
        & slow-fast        & -0.0067 &   0.0156 &   -0.0001 &   -0.0021 &   0.0135  &  0.004 \\
        & tpn              & -0.0889 &   -0.1248 &   0.023 &   -0.0795 &   -0.1566  &  -0.0854  \\
        & vat              &-0.0079 &   0.0342 &   -0.0103 &   -0.0398 &   0.0039  &  -0.004\\
			\bottomrule
		\end{tabular}
        }
	\end{center}
	\caption{Widely used models trained on face images from animal session of true personality dataset}  
\label{tb:BRC}
\end{table}
\setlength{\tabcolsep}{1.4pt}
% +++++++++++++++++++++++++++++++++++++++++++++++++++++++++++++++++++++++++++++++++++++++

%% frame images
% +++++++++++++++++++++++++++++++++++++++++++++++++++++++++++++++++++++++++++++++++++
\setlength{\tabcolsep}{2pt}
\begin{table}[H]
	\begin{center}
\resizebox{1\linewidth}{!} {
		\begin{tabular}{|l| c| c c c c c l|}
			\toprule
            & Traits         & Open    & Consc  & Extrav   & Agree   & Neuro   & Avg.   \\
            \hline \hline

        \multirow{7}*{MSE} 
        % & deep-bimodal reg   & 0.5693  & 0.6254  & 0.6070  & 0.4855  & 0.6025  & 0.5779  \\ % only frame no audio 
        & Interpret-img    & 1.0693 & 0.6474 & 1.2433 & 0.8372 & 1.9070 & 1.1408  \\
        & PersEmoN         & 0.8906 & 0.7619 & 1.4443 & 0.9543 & 1.1543 & 1.0410 \\
        & senet            & 1.3736 & 0.5775 & 1.2168 & 0.9303 & 1.5372 & 1.1271 \\
        & hrnet            & 1.2579 & 1.1331 & 1.6776 & 0.9822 & 1.6957 & 1.3493  \\
        & swin-transformer & 1.9168 & 0.6760 & 1.2712 & 0.8628 & 1.7292 & 1.2912  \\
        & 3D-resnet        & 0.9105 & 0.8549 & 1.1854 & 0.9375 & 1.5159 & 1.0809   \\
        & slow-fast        & 0.9700 & 0.9045 & 1.8854 & 1.1839 & 1.2989 & 1.2485   \\
        & tpn              & 1.0443 & 0.6347 & 0.9221 & 0.9632 & 1.1037 & 0.9336 \\
        & vat              & 1.2677 & 1.2324 & 1.1153 & 0.9270 & 1.6506 & 1.2386     \\
            \hline
        \multirow{7}*{PCC} 
        % & deep-bimodal reg   & 0.5693  & 0.6254  & 0.6070  & 0.4855  & 0.6025  & 0.5779  \\ % only frame no audio 
        & Interpret-img    &  0.1762 &  0.4076 &  0.2322 &  0.2734 & -0.4163 &  0.1346  \\
        & PersEmoN         &  0.0809 & -0.0992 & -0.2109 &  0.0328 &  0.0242 & -0.0345  \\
        & senet            & -0.1709 &  0.5200 &  0.1916 &  0.1962 & -0.2652 &  0.0943  \\
        & hrnet            &  0.1672 &  0.1602 & -0.0155 &  0.2072 & -0.0457 &  0.0947   \\
        & swin-transformer & -0.1514 &  0.3808 &  0.1354 &  0.3576 & -0.3163 &  0.0812   \\
        & 3D-resnet        & -0.0189 &  0.1567 &  0.1834 & -0.3627 & -0.2431 & -0.0569  \\
        & slow-fast        & -0.0216 &  0.0947 &  0.3926 &  0.4638 & -0.4422 &  0.0975   \\
        & tpn              & -0.0481 &  0.2326 &  0.2752 &  0.2575 &  0.1441 &  0.1722   \\
        & vat              &  0.0269 & -0.0362 &  0.4567 &  0.3412 & -0.1749 &  0.1227   \\
            \hline
        \multirow{7}*{CCC} 
        % & deep-bimodal reg   & 0.9098 & 0.9106 & 0.9096 & 0.9102 & 0.9061 & 0.9093 \\
        & Interpret-img    &  0.1142 &  0.2942 &  0.0820 &  0.1738 & -0.2997 &  0.0729  \\
        & PersEmoN         &  0.0056 & -0.0117 & -0.0091 &  0.0020 &  0.0009 & -0.0024 \\
        & senet            & -0.1086 &  0.5163 &  0.0846 &  0.1393 & -0.1339 &  0.0995  \\
        & hrnet            &  0.1480 &  0.1505 & -0.0099 &  0.1654 & -0.0398 &  0.0828  \\
        & swin-transformer & -0.1006 &  0.3722 &  0.0891 &  0.2606 & -0.2075 &  0.0827   \\
        & 3D-resnet        & -0.0040 &  0.1496 &  0.0842 & -0.0216 & -0.1184 &  0.0180   \\
        & slow-fast        & -0.0037 &  0.0087 &  0.0179 &  0.0228 & -0.0321 &  0.0027 \\
        & tpn              & -0.0319 &  0.1442 &  0.0944 &  0.0718 &  0.0339 &  0.0625   \\
        & vat              &  0.0245 & -0.0356 &  0.2886 &  0.3077 & -0.1314 &  0.0908   \\
			\bottomrule
		\end{tabular}
        }
	\end{center}
	\caption{Models trained on frame images from animal session of true personality dataset}  
\label{tb:BRC}
\end{table}
\setlength{\tabcolsep}{1.4pt}
% +++++++++++++++++++++++++++++++++++++++++++++++++++++++++++++++++++++++++++++++++++++++

%% face images
% +++++++++++++++++++++++++++++++++++++++++++++++++++++++++++++++++++++++++++++++++++
\setlength{\tabcolsep}{2pt}
\begin{table}[H]
	\begin{center}
\resizebox{1\linewidth}{!} {
		\begin{tabular}{|l| c| c c c c c l|}
			\toprule
            & Traits         & Open    & Consc  & Extrav   & Agree   & Neuro   & Avg.   \\
            \hline \hline

        \multirow{7}*{MSE} 
        % & deep-bimodal reg   & 0.5693  & 0.6254  & 0.6070  & 0.4855  & 0.6025  & 0.5779  \\ % only frame no audio 
        & Interpret-img    & 1.2059 & 0.4850 & 1.4672 & 0.8943 & 1.0455  & 1.0196  \\
        & PersEmoN         & 0.9055 & 0.7012 & 1.3966 & 0.9271 & 1.1476  & 1.0156  \\
        & senet            & 1.4740 & 0.5516 & 1.6512 & 1.1565 & 1.3277  & 1.2322  \\
        & hrnet            & 1.3607 & 0.9452 & 2.0074 & 1.0630 & 1.3894  & 1.3531  \\
        & swin-transformer & 0.8935 & 0.6787 & 1.2692 & 0.9137 & 1.3492  & 1.0209  \\
        & 3D-resnet        & 0.8258 & 0.7275 & 0.9864 & 0.8450 & 1.3303  & 0.9430  \\
        & slow-fast        & 1.0164 & 0.9797 & 2.0597 & 1.1819 & 1.2718  & 1.3019  \\
        & tpn              & 0.9872 & 0.6798 & 1.1668 & 0.9234 & 1.4819  & 1.0478  \\
        & vat              & 0.8701 & 0.6950 & 0.9966 & 0.8233 & 1.2653  & 0.9300  \\

            \hline
        \multirow{7}*{PCC} 
        % & deep-bimodal reg   & 0.5693  & 0.6254  & 0.6070  & 0.4855  & 0.6025  & 0.5779  \\ % only frame no audio 
        & Interpret-img    & -0.1785 &  0.5500 &  0.0458 &  0.2164 &  0.4864 &  0.2240  \\
        & PersEmoN         & -0.1039 &  0.0249 &  0.1973 &  0.3418 &  0.1847 &  0.1290  \\
        & senet            & -0.1868 &  0.5388 & -0.0108 &  0.0653 &  0.4152 &  0.1644  \\
        & hrnet            &  0.0457 &  0.2581 & -0.1192 &  0.1638 &  0.1326 &  0.0962  \\
        & swin-transformer &  0.0110 &  0.0618 &  0.2795 & -0.0194 & -0.0303 &  0.0605  \\
        & 3D-resnet        &  0.3140 & -0.2809 & -0.0471 & -0.1068 & -0.146  & -0.0534  \\
        & slow-fast        & -0.0416 &  0.1630 &  0.0583 &  0.1418 &  0.2216 &  0.1086  \\
        & tpn              & -0.0229 &  0.0732 &  0.3153 & -0.0353 & -0.2657 &  0.0129  \\
        & vat              & -0.2700 &  0.0476 &  0.0946 & -0.3850 & -0.1372 & -0.1300  \\
            \hline
        \multirow{7}*{CCC} 
        % & deep-bimodal reg   & 0.9098 & 0.9106 & 0.9096 & 0.9102 & 0.9061 & 0.9093 \\ 
        & Interpret-img    & -0.1338 &  0.4230 &  0.0180 &  0.1493 &  0.3017 &  0.1516  \\
        & PersEmoN         & -0.0126 &  0.0033 &  0.0069 &  0.0184 &  0.0046 &  0.0041  \\
        & senet            &  0.1667 &  0.4888 & -0.0053 &  0.0457 &  0.3104 &  0.1346  \\
        & hrnet            &  0.0443 &  0.2446 & -0.0621 &  0.1355 &  0.0964 &  0.0918  \\
        & swin-transformer &  0.0015 &  0.0127 &  0.0248 & -0.0023 & -0.0071 &  0.0059  \\
        & 3D-resnet        &  0.0513 & -0.0232 & -0.0013 & -0.0216 & -0.0098 & -0.0009  \\
        & slow-fast        & -0.0015 &  0.0035 &  0.0007 &  0.0035 &  0.0115 &  0.0035  \\
        & tpn              & -0.0117 &  0.0241 &  0.0403 & -0.0095 & -0.1066 & -0.0127  \\
        & vat              & -0.0111 &  0.0009 &  0.0017 & -0.0075 & -0.0093 & -0.0051  \\
			\bottomrule
		\end{tabular}
        }
	\end{center}
	\caption{Models trained on face images from animal session of true personality dataset}  
\label{tb:BRC}
\end{table}
\setlength{\tabcolsep}{1.4pt}
% +++++++++++++++++++++++++++++++++++++++++++++++++++++++++++++++++++++++++++++++++++++++

\subsubsection{animal session with audio-visual data}

Models are trained with frame and audio data
% +++++++++++++++++++++++++++++++++++++++++++++++++++++++++++++++++++++++++++++++++++
\setlength{\tabcolsep}{2pt}
\begin{table}[H]
	\begin{center}
\resizebox{1\linewidth}{!} {
		\begin{tabular}{|l| c| c c c c c l|}
			\toprule
            & Traits         & Open    & Consc  & Extrav   & Agree   & Neuro   & Avg.   \\
            \hline \hline

        \multirow{4}*{MSE} 
                & Deep-bimodal reg    & 0.9239 &   0.6524 &   1.6089 &   0.9596 &   1.0833  &  1.0456  \\
                & Bi-modal lstm       & 0.9194 &   0.7211 &   1.2722 &   0.9110 &   1.3370  &  1.0322   \\
                & Audiovisual-resent  & 0.9817 &   0.6550 &   1.3746 &   0.9179 &   1.1513  &  1.0161   \\
                & CRNet               & 1.3008 &   0.9348 &   1.0743 &   0.7303 &   1.7133  &  1.1507 \\
            \hline
        \multirow{4}*{PCC} 
                & Deep-bimodal reg    &  0.0317   & 0.2293 &   0.0287 &   0.1960 &    0.3063  &  0.1584   \\
                & Bi-modal lstm       & -0.1341   & 0.1164 &   0.2847 &   0.2105 &   -0.3279  &  0.0299   \\
                & Audiovisual-resent  &  0.0798   & 0.4682 &   0.1774 &   0.1555 &   -0.1116  &  0.1539   \\
                & CRNet               & -0.0692   & 0.1520 &   0.2774 &   0.4893 &   -0.0434  &  0.1612\\
            \hline
        \multirow{4}*{CCC} 
                & Deep-bimodal reg    &  0.0154  & 0.1394 & 0.0077 & 0.0964 &  0.1206 &  0.0759  \\
                & Bi-modal lstm       & -0.0006  & 0.0004 & 0.0002 & 0.0002 & -0.0015 & -0.0003  \\
                & Audiovisual-resent  &  0.0547  & 0.3155 & 0.0432 & 0.0490 & -0.0001 &  0.0925    \\
                & CRNet               & -0.0517  & 0.1452 & 0.2303 & 0.4313 & -0.0341 &  0.1442 \\
			\bottomrule
		\end{tabular}
        }
	\end{center}
	\caption{Models trained on frames and audio data from animal session of true personality dataset}  
\label{tb:BRC}
\end{table}
\setlength{\tabcolsep}{1.4pt}
% +++++++++++++++++++++++++++++++++++++++++++++++++++++++++++++++++++++++++++++++++++++++

Models are trained with face images and audio data
% +++++++++++++++++++++++++++++++++++++++++++++++++++++++++++++++++++++++++++++++++++
\setlength{\tabcolsep}{2pt}
\begin{table}[H]
	\begin{center}
\resizebox{1\linewidth}{!} {
		\begin{tabular}{|l| c| c c c c c l|}
			\toprule
            & Traits         & Open    & Consc  & Extrav   & Agree   & Neuro   & Avg.   \\
            \hline \hline

        \multirow{4}*{MSE} 
                & Deep-bimodal reg    & 1.3960 &   0.7406 &   2.2250 &   1.2325 &   1.1166  &  1.3421 \\
                & Bi-modal lstm       & 0.9013 &   0.7473 &   1.1401 &   0.8850 &   1.2606  &  0.9869  \\
                & Audiovisual-resent  & 1.2119 &   0.8100 &   1.5209 &   1.1530 &   1.1513  &  1.1694  \\
                & CRNet               & 1.3008 &   0.9348 &   1.0743 &   0.7303 &   1.7133  &  1.1507 \\
            \hline
        \multirow{4}*{PCC} 
                & Deep-bimodal reg    & -0.2414 &  0.2866 &  -0.0134 &  -0.0615 &    0.2089  &  0.0358  \\
                & Bi-modal lstm       &  0.0967 &  0.2785 &   0.1234 &  -0.1681 &    0.0612  &  0.0783 \\
                & Audiovisual-resent  & -0.2612 &  0.4312 &   0.2331 &   0.0430 &   -0.1289  &  0.0634  \\
                & CRNet               & -0.0692 &  0.1520 &   0.2774 &   0.4893 &   -0.0434  &  0.1612\\
            \hline
        \multirow{4}*{CCC} 
                & Deep-bimodal reg    & -0.1019 &  0.1774 & -0.0040 &  -0.0296 &   0.0436  &  0.0171 \\
                & Bi-modal lstm       &  0.0002 &  0.0000 &  0.0000 &  -0.0004 &   0.0001  & -0.0000 \\
                & Audiovisual-resent  & -0.1652 &  0.2567 &  0.0742 &   0.0205 &  -0.0001  &  0.0372  \\
                & CRNet               & -0.0517 &  0.1452 &  0.2303 &   0.4313 &  -0.0341  &  0.1442 \\
			\bottomrule
		\end{tabular}
        }
	\end{center}
	\caption{Models trained on face images and audio data from animal session of true personality dataset}  
\label{tb:BRC}
\end{table}
\setlength{\tabcolsep}{1.4pt}
% +++++++++++++++++++++++++++++++++++++++++++++++++++++++++++++++++++++++++++++++++++++++

% ================================== GHOST ================================================
\subsubsection{ghost session with audio data}

%% audio modality
% +++++++++++++++++++++++++++++++++++++++++++++++++++++++++++++++++++++++++++++++++++
\setlength{\tabcolsep}{2pt}
\begin{table}[H]
	\begin{center}
\resizebox{1\linewidth}{!} {
		\begin{tabular}{|l| c| c c c c c l|}
			\toprule
            & Traits         & Open    & Consc  & Extrav   & Agree   & Neuro   & Avg.   \\
            \hline \hline

        \multirow{3}*{MSE}
        & FFT            & 1.008  & 0.6564 & 1.6954 & 0.9840 & 1.3705 &  1.1429   \\
        & MFCC/logfbank  & 0.9267 & 1.0670 & 2.0137 & 1.2291 & 1.3592 &  1.3191   \\
        & Bi-modal CNN   & 0.9906 & 0.8100 & 1.2440 & 0.8549 & 1.2061 &  1.0211   \\
        & Res18-aud      & 0.9411 & 0.7085 & 1.2677 & 0.8659 & 1.0765 &  0.9720   \\
        & crnet-aud      & 0.9059 & 0.6471 & 1.3528 & 0.8653 & 1.8698 &  1.1282   \\
            \hline
        \multirow{3}*{PCC} 
        & FFT            & -0.4057 & -0.4291 &  0.2764 & -0.2088 & -0.1979  &  -0.193   \\
        & MFCC/logfbank  & -0.0079 &  0.0725 & -0.1554 &  0.1895 &  0.0625  &  0.0322   \\
        & Bi-modal CNN   &  0.0722 & -0.3649 &  0.3818 & -0.0341 &  0.1585  &  0.0427   \\
        & Res18-aud      & -0.0585 &  0.2891 &  0.1896 &  0.1640 &  0.4224  &  0.2013   \\
        & crnet-aud      & -0.0808 &  0.1607 &  0.0868 &  0.3079 &  0.2019  &  0.1353   \\
            \hline
        \multirow{3}*{CCC} 
        & FFT            &  0.0000 & -0.0000 &  0.0000 &  0.0000 & -0.0000  &  0.0000 \\
        & MFCC/logfbank  &  0.0000 &  0.0004 & -0.0002 &  0.0009 &  0.0002  &  0.0003 \\
        & Bi-modal CNN   &  0.0001 & -0.0001 &  0.0001 &  0.0000 &  0.0000  &  0.0000 \\
        & Res18-aud      & -0.0275 &  0.0776 &  0.0153 &  0.0274 &  0.0027  &  0.0191  \\
        & crnet-aud      & -0.0172 &  0.0445 &  0.0233 &  0.0935 &  0.1945  &  0.0677  \\
			\bottomrule
		\end{tabular}
        }
	\end{center}
	\caption{Models trained on audio from animal session of true personality dataset}  
\label{tb:BRC}
\end{table}
\setlength{\tabcolsep}{1.4pt}
% +++++++++++++++++++++++++++++++++++++++++++++++++++++++++++++++++++++++++++++++++++++++

\subsubsection{ghost session with visual data}

Models are trained with frames from ghost session in UDIVA dataset
%% frame images
% +++++++++++++++++++++++++++++++++++++++++++++++++++++++++++++++++++++++++++++++++++
\setlength{\tabcolsep}{2pt}
\begin{table}[H]
	\begin{center}
\resizebox{1\linewidth}{!} {
		\begin{tabular}{|l| c| c c c c c l|}
			\toprule
            & Traits         & Open    & Consc  & Extrav   & Agree   & Neuro   & Avg.   \\
            \hline \hline

        \multirow{7}*{MSE} 
        & Interpret-img    & 1.1810  & 0.6673 & 1.1209 & 0.8531 & 1.8183 &  1.1281  \\
        & PersEmoN         & 0.8640  & 0.8072 & 1.3441 & 0.9230 & 1.1511 &  1.0179  \\
        & senet            & 0.8827  & 0.7999 & 1.6970 & 0.9647 & 1.2520 &  1.1193 \\
        & hrnet            & 1.3134  & 0.9756 & 1.4561 & 1.0747 & 1.7189 &  1.3077  \\
        & swin-transformer & 1.6180  & 0.7822 & 1.4898 & 1.1696 & 1.3811 &  1.2881  \\
        & 3D-resnet        & 2.3804  & 0.6555 & 1.2846 & 1.9083 & 1.6605 &  1.5779  \\
        & slow-fast        & 1.0009  & 0.9362 & 2.0899 & 1.2640 & 1.3649 &  1.3312   \\
        & tpn              & 1.0323  & 0.6926 & 0.9757 & 0.9033 & 1.2274 &  0.9662 \\
        & vat              & 0.7584  & 0.7026 & 1.1180 & 0.9079 & 1.2688 &  0.9512      \\
            \hline
        \multirow{7}*{PCC} 
        & Interpret-img    &  0.0681  &  0.5305 &  0.4080 &  0.2552  & -0.3828  &  0.1758  \\
        & PersEmoN         &  0.2489  & -0.0456 & -0.0497 &  0.0148  &  0.2328  &  0.0802  \\
        & senet            &  0.2824  &  0.2820 &  0.0585 &  0.3240  &  0.4444  &  0.2783  \\
        & hrnet            &  0.2825  &  0.3552 &  0.3106 &  0.1413  & -0.2506  &  0.1678    \\
        & swin-transformer & -0.0922  &  0.1544 &  0.1217 & -0.0550  &  0.0047  &  0.0267    \\
        & 3D-resnet        &  0.0991  &  0.1570 & -0.2710 & -0.0613  & -0.2524  & -0.0657  \\
        & slow-fast        & -0.2071  & -0.0953 &  0.1178 &  0.0399  &  0.2198  &  0.0150    \\
        & tpn              & -0.0107  & -0.0891 &  0.4064 &  0.1055  &  0.2487  &  0.1322    \\
        & vat              &  0.3954  & -0.0664 &  0.6031 &  0.0559  & -0.0235  &  0.1929   \\
            \hline
        \multirow{7}*{CCC} 
        & Interpret-img    &  0.0441 &  0.3442  &  0.1676 &  0.1677 & -0.2422  &  0.0963 \\
        & PersEmoN         &  0.0267 & -0.0081  & -0.0006 &  0.0001 &  0.0005  &  0.0037  \\
        & senet            &  0.2204 &  0.1873  &  0.0267 &  0.1781 &  0.3019  &  0.1829  \\
        & hrnet            &  0.2464 &  0.2084  &  0.1857 &  0.1250 & -0.1967  &  0.1137  \\
        & swin-transformer & -0.0668 &  0.1384  &  0.0647 & -0.0434 &  0.0026  &  0.0191   \\
        & 3D-resnet        &  0.0880 &  0.0122  & -0.0595 & -0.0604 & -0.1818  & -0.0403   \\
        & slow-fast        & -0.0186 & -0.0071  &  0.0038 &  0.0023 &  0.0117  & -0.0016 \\
        & tpn              & -0.0050 & -0.0258  &  0.0775 &  0.0231 &  0.0122  &  0.0164  \\
        & vat              &  0.2017 & -0.0193  &  0.1318 &  0.0080 & -0.0042  &  0.0636   \\
			\bottomrule
		\end{tabular}
        }
	\end{center}
	\caption{Models trained on frame images from ghost session of true personality dataset}  
\label{tb:BRC}
\end{table}
\setlength{\tabcolsep}{1.4pt}
% +++++++++++++++++++++++++++++++++++++++++++++++++++++++++++++++++++++++++++++++++++++++

Models are trained with face images from ghost session in UDIVA dataset
%% face images
% +++++++++++++++++++++++++++++++++++++++++++++++++++++++++++++++++++++++++++++++++++
\setlength{\tabcolsep}{2pt}
\begin{table}[H]
	\begin{center}
\resizebox{1\linewidth}{!} {
		\begin{tabular}{|l| c| c c c c c l|}
			\toprule
            & Traits         & Open    & Consc  & Extrav   & Agree   & Neuro   & Avg.   \\
            \hline \hline

        \multirow{7}*{MSE} 
        & Interpret-img    & 0.9639 & 0.5597 & 1.4867 & 0.6791 & 1.1713 &  0.9721  \\
        & PersEmoN         & 0.8982 & 0.7044 & 1.4254 & 0.9632 & 1.1483 &  1.0279  \\
        & senet            & 0.8827 & 0.7999 & 1.6970 & 0.9647 & 1.2520 &  1.1193  \\
        & hrnet            & 0.9361 & 0.8271 & 1.9573 & 1.0023 & 1.6187 &  1.2683  \\
        & swin-transformer & 0.9059 & 0.6808 & 1.2752 & 0.9103 & 1.2463 &  1.0037  \\
        & 3D-resnet        & 1.1667 & 0.7493 & 1.1321 & 0.8848 & 1.1850 &  1.0235  \\
        & slow-fast        & 1.0626 & 0.9316 & 2.1434 & 1.3386 & 1.3015 &  1.3554  \\
        & tpn              & 0.9270 & 0.7069 & 1.1732 & 0.9507 & 1.2774 &  1.0070  \\
        & vat              & 0.8192 & 0.7595 & 1.1775 & 0.8464 & 1.1671 &  0.9538  \\
            \hline
        \multirow{7}*{PCC} 
        % & deep-bimodal reg   & 0.5693  & 0.6254  & 0.6070  & 0.4855  & 0.6025  & 0.5779  \\ % only frame no audio 
        & Interpret-img    &  0.1678 &  0.4070  &  0.0242 &  0.5779  & 0.4098  &  0.3173  \\
        & PersEmoN         &  0.0181 &  0.0936  &  0.0076 &  0.0218  & 0.1875  &  0.0657  \\
        & senet            &  0.2824 &  0.2820  &  0.0585 &  0.3240  & 0.4444  &  0.2783  \\
        & hrnet            &  0.4022 &  0.3365  &  0.0096 &  0.2371  & 0.3968  &  0.2765  \\
        & swin-transformer & -0.0415 & -0.0110  &  0.0071 & -0.1015  & 0.1480  &  0.0002  \\
        & 3D-resnet        & -0.1634 &  0.0298  &  0.2992 & -0.0089  & 0.2142  &  0.0742  \\
        & slow-fast        &  0.0344 & -0.0178  & -0.0134 & -0.1143  & 0.2145  &  0.0207  \\
        & tpn              & -0.1552 & -0.0320  &  0.2481 &  0.0903  & 0.1960  &  0.0694  \\
        & vat              & -0.1187 &  0.1571  & -0.0249 &  0.1905  & 0.0216  &  0.0451  \\
            \hline
        \multirow{7}*{CCC} 
        % & deep-bimodal reg   & 0.9098 & 0.9106 & 0.9096 & 0.9102 & 0.9061 & 0.9093 \\ 
        & Interpret-img    &  0.1073  &  0.3033  &  0.0097  &  0.3634  & 0.2506  &  0.2069  \\
        & PersEmoN         &  0.0017  &  0.0129  &  0.0002  &  0.0012  & 0.0051  &  0.0042  \\
        & senet            &  0.2204  &  0.1873  &  0.0267  &  0.1781  & 0.3019  &  0.1829    \\
        & hrnet            &  0.3780  &  0.3317  &  0.0064  &  0.2090  & 0.3315  &  0.2513    \\
        & swin-transformer &  0.0000  &  0.0000  &  0.0000  &  0.0000  & 0.0000  &  0.0000   \\
        & 3D-resnet        & -0.1239  &  0.0082  &  0.0660  & -0.0017  & 0.0648  &  0.0027     \\
        & slow-fast        &  0.0043  & -0.0014  & -0.0003  & -0.0044  & 0.0185  &  0.0033   \\
        & tpn              & -0.0339  & -0.0135  &  0.0702  &  0.0235  & 0.0628  &  0.0218   \\
        & vat              & -0.0071  &  0.0042  & -0.0006  &  0.0038  & 0.0006  &  0.0002    \\
			\bottomrule
		\end{tabular}
        }
	\end{center}
	\caption{Models trained on face images from ghost session of true personality dataset}  
\label{tb:BRC}
\end{table}
\setlength{\tabcolsep}{1.4pt}
% +++++++++++++++++++++++++++++++++++++++++++++++++++++++++++++++++++++++++++++++++++++++

\subsubsection{ghost session with video-level visual data}

Models are trained with frame data
%% frame images
% +++++++++++++++++++++++++++++++++++++++++++++++++++++++++++++++++++++++++++++++++++
\setlength{\tabcolsep}{2pt}
\begin{table}[H]
	\begin{center}
\resizebox{1\linewidth}{!} {
		\begin{tabular}{|l| c| c c c c c l|}
			\toprule
            & Traits         & Open    & Consc  & Extrav   & Agree   & Neuro   & Avg.   \\
            \hline \hline

        \multirow{5}*{MSE} 
        & 3D-resnet        & 1.0256 &   0.8057 &   1.2962 &   1.0806 &   1.2396  &  1.08969 \\
        & slow-fast        & 1.0436 &   0.8525 &   2.1877 &   1.1833 &   1.469  &  1.3472 \\
        & tpn              & 0.9442 &   0.6506 &   1.0865 &   0.8509 &   1.2276  &  0.9519\\
        & vat              & 0.8816 &   0.6653 &   1.2592 &   0.9154 &   1.2588  &  0.9961 \\
            \hline
        \multirow{5}*{PCC} 
        & 3D-resnet        & 0.179 &   -0.3145 &   -0.2791 &   -0.0127 &   0.176  &  -0.0503  \\
        & slow-fast        & -0.0865 &   -0.2118 &   -0.5159 &   -0.0297 &   0.1746  &  -0.1339 \\
        & tpn              & 0.0203 &   0.1214 &   0.1535 &   0.2724 &   -0.0747  &  0.0986 \\
        & vat              & 0.1131 &   0.2543 &   0.5208 &   -0.1318 &   -0.2027  &  0.1108 \\
            \hline
        \multirow{5}*{CCC} 
        & 3D-resnet        & 0.1387 &   -0.1481 &   -0.0489 &   -0.0091 &   0.0224  &  -0.009 \\
        & slow-fast        & -0.0019 &   -0.0117 &   -0.0101 &   -0.0017 &   0.0042  &  -0.0042 \\
        & tpn              & 0.0081 &   0.0563 &   0.0582 &   0.0697 &   -0.0212  &  0.0342  \\
        & vat              & 0.0054 &   0.0086 &   0.0101 &   -0.003 &   -0.0046  &  0.0033  \\
			\bottomrule
		\end{tabular}
        }
	\end{center}
	\caption{Models trained on video-level frames segment from ghost session of true personality dataset}  
\label{tb:BRC}
\end{table}
\setlength{\tabcolsep}{1.4pt}
% +++++++++++++++++++++++++++++++++++++++++++++++++++++++++++++++++++++++++++++++++++++++

Models are trained with face data
%% face images
% +++++++++++++++++++++++++++++++++++++++++++++++++++++++++++++++++++++++++++++++++++
\setlength{\tabcolsep}{2pt}
\begin{table}[H]
	\begin{center}
\resizebox{1\linewidth}{!} {
		\begin{tabular}{|l| c| c c c c c l|}
			\toprule
            & Traits         & Open    & Consc  & Extrav   & Agree   & Neuro   & Avg.   \\
            \hline \hline

        \multirow{5}*{MSE} 
        & 3D-resnet        & 0.9107 &   0.9997 &   1.1969 &   1.1857 &   1.4752  &  1.1536 \\
        & slow-fast        & 0.9967 &   0.9926 &   1.8236 &   1.0453 &   1.347  &  1.2410 \\
        & tpn              & 0.9349 &   0.778 &   1.3029 &   0.9731 &   1.3639  &  1.0706 \\
        & vat              & 0.8813 &   0.6575 &   1.2946 &   0.9181 &   1.2947  &  1.0092 \\
            \hline
        \multirow{5}*{PCC} 
        & 3D-resnet        & -0.0867 &   -0.2053 &   0.4422 &   -0.1384 &   -0.0476  &  -0.0071  \\
        & slow-fast        & 0.0113 &   0.2564 &   0.2707 &   0.1145 &   -0.0559  &  0.1194  \\
        & tpn              &-0.0957 &   -0.1041 &   -0.1219 &   -0.2622 &   0.0206  &  -0.1127  \\
        & vat              & 0.1168 &   0.2199 &   -0.2202 &   -0.092 &   -0.0961  &  -0.0143 \\
            \hline
        \multirow{5}*{CCC} 
        & 3D-resnet        & -0.0189 &   -0.1691 &   0.0765 &   -0.0999 &   -0.0291  &  -0.0481 \\
        & slow-fast        & 0.0004 &   0.0186 &   0.0199 &   0.0195 &   -0.0024  &  0.0112 \\
        & tpn              &  -0.0314 &   -0.0347 &   -0.0215 &   -0.0395 &   0.0076  &  -0.0239  \\
        & vat              & 0.0117 &   0.0453 &   -0.0093 &   -0.0031 &   -0.0112  &  0.0067  \\
			\bottomrule
		\end{tabular}
        }
	\end{center}
	\caption{Widely used models trained on video-level face images segment from ghost session of true personality dataset}  
\label{tb:BRC}
\end{table}
\setlength{\tabcolsep}{1.4pt}
% +++++++++++++++++++++++++++++++++++++++++++++++++++++++++++++++++++++++++++++++++++++++

\subsubsection{ghost session with audio-visual data}

Models are trained with frame and audio data
% +++++++++++++++++++++++++++++++++++++++++++++++++++++++++++++++++++++++++++++++++++
\setlength{\tabcolsep}{2pt}
\begin{table}[H]
	\begin{center}
\resizebox{1\linewidth}{!} {
		\begin{tabular}{|l| c| c c c c c l|}
			\toprule
            & Traits         & Open    & Consc  & Extrav   & Agree   & Neuro   & Avg.   \\
            \hline \hline

        \multirow{4}*{MSE} 
                & Deep-bimodal reg    & 0.8964 &   0.7638 &   1.4031 &   0.9917 &   1.1325  &  1.0375   \\
                & Bi-modal lstm       & 0.9379 &   0.7011 &   1.2632 &   0.9207 &   1.3913  &  1.0429   \\
                & Audiovisual-resent  & 0.9028 &   0.8043 &   1.4217 &   0.9960 &   1.1513  &  1.0552   \\
                & CRNet               & 1.6663 &   0.9454 &   1.3654 &   1.4333 &   1.5648  &  1.3950   \\
            \hline
        \multirow{4}*{PCC} 
                & Deep-bimodal reg    &  0.1158 &   0.2490 &   0.3666 &   0.3009 &   0.1638  &  0.2392   \\
                & Bi-modal lstm       & -0.0157 &   0.3287 &   0.4823 &  -0.3829 &   0.5057  &  0.1836   \\
                & Audiovisual-resent  &  0.0448 &   0.4743 &   0.1644 &   0.0739 &  -0.0358  &  0.1443   \\
                & CRNet               &  0.1964 &   0.1316 &   0.1634 &   0.2077 &   0.2101  &  0.1818   \\
            \hline
        \multirow{4}*{CCC} 
                & Deep-bimodal reg    &  0.0629 &   0.1488 &   0.1184 &   0.1499 &   0.0249  &  0.1010  \\
                & Bi-modal lstm       & -0.0001 &   0.0015 &   0.0008 &  -0.0004 &   0.0037  &  0.0011  \\
                & Audiovisual-resent  &  0.0180 &   0.2706 &   0.0407 &   0.0286 &   0.0000  &  0.0716  \\
                & CRNet               &  0.1416 &   0.1284 &   0.1104 &   0.1827 &   0.1386  &  0.1403  \\
			\bottomrule
		\end{tabular}
        }
	\end{center}
	\caption{Models trained on frames and audio data from animal session of true personality dataset}  
\label{tb:BRC}
\end{table}
\setlength{\tabcolsep}{1.4pt}
% +++++++++++++++++++++++++++++++++++++++++++++++++++++++++++++++++++++++++++++++++++++++

Models are trained with face images and audio data
% +++++++++++++++++++++++++++++++++++++++++++++++++++++++++++++++++++++++++++++++++++
\setlength{\tabcolsep}{2pt}
\begin{table}[H]
	\begin{center}
\resizebox{1\linewidth}{!} {
		\begin{tabular}{|l| c| c c c c c l|}
			\toprule
            & Traits         & Open    & Consc  & Extrav   & Agree   & Neuro   & Avg.   \\
            \hline \hline

        \multirow{4}*{MSE} 
                & Deep-bimodal reg    & 1.1136 &   0.6650 &   1.6572 &   0.9816 &   1.0121  &  1.0859  \\
                & Bi-modal lstm       & 0.9143 &   0.6952 &   1.2385 &   0.9118 &   1.3594  &  1.0239  \\
                & Audiovisual-resent  & 1.1185 &   0.7121 &   1.6486 &   1.0807 &   1.1511  &  1.1422  \\
                & CRNet               & 1.6663 &   0.9454 &   1.3654 &   1.4333 &   1.5648  &  1.3950  \\
            \hline
        \multirow{4}*{PCC} 
                & Deep-bimodal reg    &  0.0001 &   0.2613 &   0.2122 &   0.0686 &   0.4301  &  0.1945  \\
                & Bi-modal lstm       &  0.2108 &   0.3391 &  -0.1412 &  -0.0560 &   0.4558  &  0.1617  \\
                & Audiovisual-resent  & -0.0072 &   0.2118 &   0.1070 &   0.0284 &   0.1855  &  0.1051  \\
                & CRNet               &  0.1964 &   0.1316 &   0.1634 &   0.2077 &   0.2101  &  0.1818  \\
            \hline
        \multirow{4}*{CCC} 
                & Deep-bimodal reg    &  0.0001 &   0.1507 &   0.0732 &   0.0248 &   0.1813  &  0.0860 \\
                & Bi-modal lstm       &  0.0009 &   0.0008 &  -0.0002 &  -0.0000 &   0.0022  &  0.0007 \\
                & Audiovisual-resent  & -0.0048 &   0.1316 &   0.0367 &   0.0133 &   0.0001  &  0.0354 \\
                & CRNet               &  0.1416 &   0.1284 &   0.1104 &   0.1827 &   0.1386  &  0.1403 \\
			\bottomrule
		\end{tabular}
        }
	\end{center}
	\caption{Models trained on face images and audio data from animal session of true personality dataset}  
\label{tb:BRC}
\end{table}
\setlength{\tabcolsep}{1.4pt}
% +++++++++++++++++++++++++++++++++++++++++++++++++++++++++++++++++++++++++++++++++++++++

% ========================================== LEGO ==========================================

\subsubsection{lego session with audio data}

%% audio modality
% +++++++++++++++++++++++++++++++++++++++++++++++++++++++++++++++++++++++++++++++++++
\setlength{\tabcolsep}{2pt}
\begin{table}[H]
	\begin{center}
\resizebox{1\linewidth}{!} {
		\begin{tabular}{|l| c| c c c c c l|}
			\toprule
            & Traits         & Open    & Consc  & Extrav   & Agree   & Neuro   & Avg.   \\
            \hline \hline

        \multirow{3}*{MSE} 
        & FFT            &  0.9355 &  0.7851 & 1.6091 &  0.8997 &  1.2809  &  1.1021 \\
        & MFCC/logfbank  &  0.9265 &  1.0671 & 2.0108 &  1.2307 &  1.3600  &  1.3190 \\
        & Bi-modal CNN   &  0.9544 &  0.7255 & 1.3715 &  0.8933 &  1.2817  &  1.0453 \\
        & Res18-aud      &  0.9199 &  0.6928 & 1.1577 &  0.8745 &  1.0793  &  0.9448 \\
        & crnet-aud      &  0.8976 &  0.6484 & 1.3352 &  0.9222 &  1.1521  &  0.9911 \\
            \hline
        \multirow{3}*{PCC} 
        & FFT            &  0.3000 & -0.3062 &  0.3674 &  0.1172 &  0.2893  &  0.1536  \\
        & MFCC/logfbank  &  0.0674 &  0.0219 & -0.1981 & -0.1029 & -0.1900  & -0.0803  \\
        & Bi-modal CNN   &  0.2363 &  0.1772 & -0.5476 & -0.0202 & -0.0482  & -0.0405  \\
        & Res18-aud      & -0.0786 &  0.2487 &  0.4504 &  0.0954 &  0.0731  &  0.1578  \\
        & crnet-aud      & -0.1188 &  0.1426 & -0.2425 &  0.0121 & -0.0582  & -0.0530  \\
            \hline
        \multirow{3}*{CCC} 
        & FFT            &  0.0000 & -0.0000 &  0.0000 &  0.0000 &  0.0000  &  0.0000  \\
        & MFCC/logfbank  &  0.0003 &  0.0001 & -0.0002 & -0.0005 & -0.0002  & -0.0001  \\
        & Bi-modal CNN   &  0.0001 &  0.0000 &  0.0000 &  0.0000 &  0.0000  &  0.0000  \\
        & Res18-aud      & -0.0435 &  0.1093 &  0.1283 &  0.0376  & 0.0000  &  0.0463  \\
        & crnet-aud      & -0.0002 &  0.0004 & -0.0004 &  0.0000 & -0.0002  & -0.0001  \\
			\bottomrule
		\end{tabular}
        }
	\end{center}
	\caption{Models trained on audio from animal session of true personality dataset}  
\label{tb:BRC}
\end{table}
\setlength{\tabcolsep}{1.4pt}
% +++++++++++++++++++++++++++++++++++++++++++++++++++++++++++++++++++++++++++++++++++++++

\subsubsection{lego session with visual data}

Models are trained with frames from lego session in UDIVA dataset
%% frame images
% +++++++++++++++++++++++++++++++++++++++++++++++++++++++++++++++++++++++++++++++++++
\setlength{\tabcolsep}{2pt}
\begin{table}[H]
	\begin{center}
\resizebox{1\linewidth}{!} {
		\begin{tabular}{|l| c| c c c c c l|}
			\toprule
            & Traits         & Open    & Consc  & Extrav   & Agree   & Neuro   & Avg.   \\
            \hline \hline

        \multirow{7}*{MSE} 
        % & deep-bimodal reg   & 0.5693  & 0.6254  & 0.6070  & 0.4855  & 0.6025  & 0.5779  \\ % only frame no audio 
        & Interpret-img    & 1.2310 & 0.7035 & 0.8241 & 0.9213 & 2.0767 &  1.1513  \\
        & PersEmoN         & 0.8978 & 0.7094 & 1.4197 & 0.9479 & 1.1532 &  1.0256  \\
        & senet            & 1.6945 & 0.6956 & 1.1760 & 0.9696 & 1.6169 &  1.2304  \\
        & hrnet            & 1.1884 & 1.4485 & 1.3374 & 1.0223 & 1.8183 &  1.3630  \\
        & swin-transformer & 1.9875 & 0.7044 & 1.3137 & 1.3174 & 1.4387 &  1.3523  \\
        & 3D-resnet        & 1.3462 & 0.6648 & 1.1616 & 0.8355 & 1.1885 &  1.0392  \\
        & slow-fast        & 0.9501 & 0.9238 & 2.0718 & 1.2580 & 1.3781 &  1.3164  \\
        & tpn              & 0.8971 & 0.6501 & 1.1323 & 0.9710 & 1.2257 &  0.9752  \\
        & vat              & 0.8010 & 0.6879 & 1.1261 & 0.8358 & 1.1965 &  0.9294  \\
            \hline
        \multirow{7}*{PCC} 
        % & deep-bimodal reg   & 0.5693  & 0.6254  & 0.6070  & 0.4855  & 0.6025  & 0.5779  \\ % only frame no audio 
        & Interpret-img    &  0.1314 &  0.3531  &  0.5827  &  0.1311  & -0.3176  &  0.1762  \\
        & PersEmoN         & -0.0425 &  0.0039  & -0.1414  &  0.1294  &  0.0225  & -0.0056  \\
        & senet            & -0.1173 &  0.3756  &  0.1839  &  0.1473  & -0.2213  &  0.0737  \\
        & hrnet            &  0.1247 & -0.0520  &  0.1170  &  0.2299  & -0.1361  &  0.0567  \\
        & swin-transformer & -0.1302 &  0.2453  &  0.3210  & -0.0599  &  0.0458  &  0.0844  \\
        & 3D-resnet        & -0.0300 &  0.1883  &  0.0059  &  0.1228  &  0.0131  &  0.0600  \\
        & slow-fast        &  0.1383 & -0.4440  & -0.1044  & -0.0598  & -0.0335  & -0.1007  \\
        & tpn              &  0.1485 &  0.0767  &  0.0879  & -0.0968  &  0.0661  &  0.0565  \\
        & vat              &  0.0912 & -0.2273  & -0.1694  &  0.01620 & -0.1623  & -0.0903  \\
            \hline
        \multirow{7}*{CCC} 
        % & deep-bimodal reg   & 0.9098 & 0.9106 & 0.9096 & 0.9102 & 0.9061 & 0.9093 \\ 
        & Interpret-img    &  0.0831 &  0.2209  &  0.3305  &  0.0796  & -0.211   &  0.1006  \\
        & PersEmoN         & -0.0050 &  0.0005  & -0.0055  &  0.0068  &  0.0006  & -0.0005  \\
        & senet            & -0.0799 &  0.3223  &  0.0984  &  0.1041  & -0.1377  &  0.0614  \\
        & hrnet            &  0.1122 & -0.0371  &  0.0763  &  0.2093  & -0.1197  &  0.0482  \\
        & swin-transformer & -0.0934 &  0.2111  &  0.1909  & -0.0537  &  0.0298  &  0.0569  \\
        & 3D-resnet        & -0.0209 &  0.1022  &  0.0013  &  0.0532  &  0.0009  &  0.0273  \\
        & slow-fast        &  0.0218 & -0.0485  & -0.0063  & -0.0042  & -0.002   & -0.0078  \\
        & tpn              &  0.0631 &  0.0163  &  0.0144  & -0.0339  &  0.0347  &  0.0189  \\
        & vat              &  0.0126 & -0.0389  & -0.0101  &  0.0008  & -0.0064  & -0.0084  \\
			\bottomrule
		\end{tabular}
        }
	\end{center}
	\caption{Widely used models trained on frame images from lego session of true personality dataset}  
\label{tb:BRC}
\end{table}
\setlength{\tabcolsep}{1.4pt}
% +++++++++++++++++++++++++++++++++++++++++++++++++++++++++++++++++++++++++++++++++++++++

Models are trained with face images from lego session in UDIVA dataset
%% face images
% +++++++++++++++++++++++++++++++++++++++++++++++++++++++++++++++++++++++++++++++++++
\setlength{\tabcolsep}{2pt}
\begin{table}[H]
	\begin{center}
\resizebox{1\linewidth}{!} {
		\begin{tabular}{|l| c| c c c c c l|}
			\toprule
            & Traits         & Open    & Consc  & Extrav   & Agree   & Neuro   & Avg.   \\
            \hline \hline

        \multirow{7}*{MSE} 
        & Interpret-img    & 0.9067 & 0.6180 & 1.3894 & 0.8325 & 1.1943 &  0.9882 \\
        & PersEmoN         & 0.9177 & 0.7010 & 1.4301 & 0.9493 & 1.1458 &  1.0287 \\
        & senet            & 1.1072 & 0.6053 & 1.4589 & 1.1930 & 1.4978 &  1.1725 \\
        & hrnet            & 1.0863 & 0.8670 & 1.7965 & 1.2311 & 1.9556 &  1.3873 \\
        & swin-transformer & 1.0320 & 0.7432 & 1.3494 & 0.9277 & 1.2628 &  1.0630 \\
        & 3D-resnet        & 0.9967 & 0.6704 & 1.2219 & 0.8366 & 1.3475 &  1.0146 \\
        & slow-fast        & 1.1074 & 0.9433 & 2.1837 & 1.3347 & 1.2851 &  1.3708 \\
        & tpn              & 0.8642 & 0.8446 & 1.4764 & 1.0498 & 1.5847 &  1.1639 \\
        & vat              & 0.8245 & 0.6787 & 1.1815 & 0.8457 & 1.2012 &  0.9463 \\

            \hline
        \multirow{7}*{PCC} 
        & Interpret-img    &  0.2175 &  0.3111  &  0.1230  &  0.3595 &  0.4314  &  0.2885 \\
        & PersEmoN         & -0.3087 &  0.0796  & -0.0461  &  0.1041 &  0.2123  &  0.0082 \\
        & senet            &  0.3046 &  0.3615  &  0.1277  & -0.0208 &  0.4371  &  0.2420 \\
        & hrnet            &  0.3024 &  0.2977  & -0.0568  &  0.0937 &  0.1859  &  0.1646 \\
        & swin-transformer & -0.0774 & -0.0407  & -0.0521  & -0.0418 &  0.0569  & -0.0310 \\
        & 3D-resnet        & -0.0151 & -0.0533  &  0.2669  &  0.3111 & -0.0891  &  0.0841 \\
        & slow-fast        & -0.0761 &  0.1643  & -0.1691  & -0.2679 &  0.1309  & -0.0436 \\
        & tpn              &  0.2545 &  0.1495  & -0.3532  & -0.2092 & -0.2409  & -0.0799 \\
        & vat              & -0.1600 &  0.1084  &  0.1139  &  0.0591 &  0.1627  &  0.0568 \\
            \hline
        \multirow{7}*{CCC} 
        & Interpret-img    &  0.1652  &  0.2389  &  0.0522 &  0.2529 &  0.2371  &  0.1893 \\
        & PersEmoN         & -0.033   &  0.0095  & -0.0017 &  0.0061 &  0.0072  & -0.0024 \\
        & senet            &  0.2106  &  0.2939  &  0.0637 & -0.0151 &  0.2744  &  0.1655 \\
        & hrnet            &  0.2741  &  0.2739  & -0.0384 &  0.0743 &  0.1481  &  0.1464 \\
        & swin-transformer & -0.0476  & -0.013   & -0.0073 & -0.0119 &  0.0197  & -0.0120 \\
        & 3D-resnet        & -0.0099  & -0.0099  &  0.0391 &  0.0391 & -0.0221  &  0.0072 \\
        & slow-fast        & -0.0103  &  0.0116  & -0.0035 & -0.0159 &  0.0147  & -0.0007 \\
        & tpn              &  0.1822  &  0.0400  & -0.044  & -0.0918 & -0.0969  & -0.0021 \\
        & vat              & -0.0114  &  0.0022  &  0.0017 &  0.0016 &  0.0053  & -0.0001 \\
			\bottomrule
		\end{tabular}
        }
	\end{center}
	\caption{Widely used models trained on face images from lego session of true personality dataset}  
\label{tb:BRC}
\end{table}
\setlength{\tabcolsep}{1.4pt}
% +++++++++++++++++++++++++++++++++++++++++++++++++++++++++++++++++++++++++++++++++++++++

\subsubsection{lego session with video-level visual data}

Models are trained with frame data
%% frame images
% +++++++++++++++++++++++++++++++++++++++++++++++++++++++++++++++++++++++++++++++++++
\setlength{\tabcolsep}{2pt}
\begin{table}[H]
	\begin{center}
\resizebox{1\linewidth}{!} {
		\begin{tabular}{|l| c| c c c c c l|}
			\toprule
            & Traits         & Open    & Consc  & Extrav   & Agree   & Neuro   & Avg.   \\
            \hline \hline

        \multirow{5}*{MSE} 
        & 3D-resnet        & 1.3034 &   0.6785 &   1.6971 &   1.0813 &   1.9622  &  1.3445\\
        & slow-fast        & 1.0398 &   0.836 &   1.9512 &   1.2053 &   1.5072  &  1.3080 \\
        & tpn              & 0.9442 &   0.6506 &   1.0865 &   0.8509 &   1.2276  &  0.9519 \\
        & vat              & 0.8815 &   0.6925 &   1.2734 &   0.9083 &   1.2595  &  1.0030 \\
            \hline
        \multirow{5}*{PCC} 
        & 3D-resnet        & 0.0639 &   0.1928 &   0.2124 &   -0.0003 &   -0.0615  &  0.0815  \\
        & slow-fast        & -0.1597 &   0.0913 &   -0.5591 &   -0.0377 &   -0.0738  &  -0.1478  \\
        & tpn              & 0.0203 &   0.1214 &   0.1535 &   0.2724 &   -0.0747  &  0.0986 \\
        & vat              & 0.1241 &   -0.2047 &   0.239 &   0.1339 &   -0.0431  &  0.0498 \\
            \hline
        \multirow{5}*{CCC} 
        & 3D-resnet        & 0.0295 &   0.0313 &   0.0382 &   -0.0001 &   -0.0146  &  0.0169 \\
        & slow-fast        & -0.0235 &   0.0101 &   -0.047 &   -0.003 &   -0.0019  &  -0.0131\\
        & tpn              &  0.0081 &   0.0563 &   0.0582 &   0.0697 &   -0.0212  &  0.0342  \\
        & vat              & 0.0065 &   -0.0174 &   0.0093 &   0.0061 &   -0.0018  &  0.0006 \\
			\bottomrule
		\end{tabular}
        }
	\end{center}
	\caption{Models trained on video-level frames segment from lego session of true personality dataset}  
\label{tb:BRC}
\end{table}
\setlength{\tabcolsep}{1.4pt}
% +++++++++++++++++++++++++++++++++++++++++++++++++++++++++++++++++++++++++++++++++++++++

Models are trained with face data
%% face images
% +++++++++++++++++++++++++++++++++++++++++++++++++++++++++++++++++++++++++++++++++++
\setlength{\tabcolsep}{2pt}
\begin{table}[H]
	\begin{center}
\resizebox{1\linewidth}{!} {
		\begin{tabular}{|l| c| c c c c c l|}
			\toprule
            & Traits         & Open    & Consc  & Extrav   & Agree   & Neuro   & Avg.   \\
            \hline \hline

        \multirow{5}*{MSE} 
        & 3D-resnet        & 0.8107 &   0.6244 &   1.3232 &   0.9364 &   1.252  &  0.9894 \\
        & slow-fast        & 1.0042 &   0.9923 &   1.9656 &   1.2506 &   1.2382  &  1.2902 \\
        & tpn              & 1.2592 &   0.8766 &   1.5572 &   1.033 &   1.3165  &  1.2085\\
        & vat              & 0.9003 &   0.6758 &   1.2838 &   0.9127 &   1.2517  &  1.0049\\
            \hline
        \multirow{5}*{PCC} 
        & 3D-resnet        & 0.3053 &   0.2696 &   0.149 &   -0.0935 &   0.2545  &  0.177  \\
        & slow-fast        & -0.431 &   0.1096 &   -0.2098 &   0.1436 &   -0.1099  &  -0.0995 \\
        & tpn              & -0.1007 &   -0.2531 &   0.233 &   -0.178 &   -0.1501  &  -0.0898  \\
        & vat              & -0.0581 &   0.1641 &   -0.1375 &   0.1197 &   0.2241  &  0.0625  \\
            \hline
        \multirow{5}*{CCC} 
        & 3D-resnet        & 0.124 &   0.079 &   0.0083 &   -0.0106 &   0.0341  &  0.047 \\
        & slow-fast        & -0.0172 &   0.004 &   -0.0044 &   0.0059 &   -0.0076  &  -0.0039\\
        & tpn              &-0.0746 &   -0.1475 &   0.1193 &   -0.0701 &   -0.0407  &  -0.0427   \\
        & vat              & -0.008 &   0.0095 &   -0.01 &   0.0058 &   0.0136  &  0.0022  \\
			\bottomrule
		\end{tabular}
        }
	\end{center}
	\caption{Widely used models trained on video-level face images segment from lego session of true personality dataset}  
\label{tb:BRC}
\end{table}
\setlength{\tabcolsep}{1.4pt}
% +++++++++++++++++++++++++++++++++++++++++++++++++++++++++++++++++++++++++++++++++++++++

\subsubsection{lego session with audio-visual data}

Models are trained with frame and audio data
% +++++++++++++++++++++++++++++++++++++++++++++++++++++++++++++++++++++++++++++++++++
\setlength{\tabcolsep}{2pt}
\begin{table}[H]
	\begin{center}
\resizebox{1\linewidth}{!} {
		\begin{tabular}{|l| c| c c c c c l|}
			\toprule
            & Traits         & Open    & Consc  & Extrav   & Agree   & Neuro   & Avg.   \\
            \hline \hline

        \multirow{4}*{MSE} 
                & Deep-bimodal reg    & 0.9427 &   0.8394 &   1.5333 &   1.1341 &   1.1690  &  1.1237  \\
                & Bi-modal lstm       & 1.0299 &   0.6821 &   1.5326 &   0.7560 &   1.2819  &  1.0565  \\
                & Audiovisual-resent  & 0.8191 &   0.7605 &   1.5265 &   1.1074 &   1.1516  &  1.0730  \\
                & CRNet               & 0.9029 &   0.9902 &   1.5870 &   0.8247 &   1.9834  &  1.2577  \\
            \hline
        \multirow{4}*{PCC} 
                & Deep-bimodal reg    & -0.0221 &  0.0501 &   0.2460 &    0.2176 &   0.0064  &  0.0996  \\
                & Bi-modal lstm       & -0.3049 &  0.1226 &   0.2066 &    0.0669 &  -0.0441  &  0.0094  \\
                & Audiovisual-resent  &  0.2880 &  0.2773 &   0.4055 &   -0.0505 &   0.0441  &  0.1929  \\
                & CRNet               &  0.3475 &  0.0672 &  -0.0554 &    0.3337 &   0.0932  &  0.1572  \\
            \hline
        \multirow{4}*{CCC} 
                & Deep-bimodal reg    & -0.0104 &   0.0279 &   0.0753 &   0.1271 &   0.0014  &  0.0442  \\
                & Bi-modal lstm       & -0.0004 &   0.0007 &   0.0003 &   0.0001 &  -0.0001  &  0.0001  \\
                & Audiovisual-resent  &  0.1795 &   0.1702 &   0.1378 &  -0.0223 &   0.0001  &  0.0930  \\
                & CRNet               &  0.3220 &   0.0560 &  -0.0255 &   0.2547 &   0.0650  &  0.1345  \\
			\bottomrule
		\end{tabular}
        }
	\end{center}
	\caption{Models trained on frames and audio data from animal session of true personality dataset}  
\label{tb:BRC}
\end{table}
\setlength{\tabcolsep}{1.4pt}
% +++++++++++++++++++++++++++++++++++++++++++++++++++++++++++++++++++++++++++++++++++++++

Models are trained with face images and audio data
% +++++++++++++++++++++++++++++++++++++++++++++++++++++++++++++++++++++++++++++++++++
\setlength{\tabcolsep}{2pt}
\begin{table}[H]
	\begin{center}
\resizebox{1\linewidth}{!} {
		\begin{tabular}{|l| c| c c c c c l|}
			\toprule
            & Traits         & Open    & Consc  & Extrav   & Agree   & Neuro   & Avg.   \\
            \hline \hline

        \multirow{4}*{MSE} 
                & Deep-bimodal reg    & 1.1673 &   0.7461 &   1.9958 &   1.1203 &   1.0975  &  1.2254  \\
                & Bi-modal lstm       & 0.9826 &   0.7781 &   1.5581 &   0.7402 &   1.2767  &  1.0671  \\
                & Audiovisual-resent  & 0.8869 &   0.6588 &   1.3523 &   0.9308 &   1.1514  &  0.9960  \\
                & CRNet               & 0.9029 &   0.9902 &   1.5870 &   0.8247 &   1.9834  &  1.2577  \\
            \hline
        \multirow{4}*{PCC} 
                & Deep-bimodal reg    &  0.0429 &   0.1059 &   -0.1703 &   -0.0024 &   0.2746  &  0.0502  \\
                & Bi-modal lstm       &  0.1334 &  -0.1660 &    0.0126 &    0.3451 &  -0.0438  &  0.0563  \\
                & Audiovisual-resent  & -0.0149 &   0.2306 &    0.0670 &    0.0483 &   0.0771  &  0.0816  \\
                & CRNet               &  0.3475 &   0.0672 &   -0.0554 &    0.3337 &   0.0932  &  0.1572  \\
            \hline
        \multirow{4}*{CCC} 
                & Deep-bimodal reg    &  0.0226 &   0.0621 &  -0.0543 &  -0.0011 &   0.0701  &  0.0199  \\
                & Bi-modal lstm       &  0.0001 &  -0.0005 &   0.0000 &   0.0001 &  -0.0001  & -0.0001  \\
                & Audiovisual-resent  & -0.0011 &   0.0234 &   0.0009 &   0.0014 &   0.0005  &  0.0050  \\
                & CRNet               &  0.3220 &   0.0560 &  -0.0255 &   0.2547 &   0.0650  &  0.1345  \\
			\bottomrule
		\end{tabular}
        }
	\end{center}
	\caption{Models trained on face images and audio data from animal session of true personality dataset}  
\label{tb:BRC}
\end{table}
\setlength{\tabcolsep}{1.4pt}
% +++++++++++++++++++++++++++++++++++++++++++++++++++++++++++++++++++++++++++++++++++++++

\subsubsection{Results summery of visual data in frame from four session}

\setlength{\tabcolsep}{2pt}
\begin{table}[H]
	\begin{center}
        \resizebox{1\linewidth}{!} {
		    \begin{tabular}{|l| c| c c c c c l|}
			\toprule
                & Traits         & Open    & Consc   & Extrav  & Agree   & Neuro   & Avg.      \\
                \hline \hline

            \multirow{7}*{Video} 
                & Interpret-img    & 1.1190  & 0.6459  & 1.0673  & 0.8790  & 1.8076  & 1.1037  \\
                & PersEmoN         & 0.8740  & 0.7481  & 1.3866  & 0.9394  & 1.1526  & 1.0201  \\
                & SENet            & 1.3032  & 0.6783  & 1.3155  & 0.9298  & 1.4625  & 1.1378  \\
                & HRNet            & 1.2195  & 1.0896  & 1.4482  & 0.9684  & 1.6250  & 1.2702  \\
                & Swin-transformer & 1.7569  & 0.7028  & 1.3235  & 1.0862  & 1.4817  & 1.2702  \\
                & 3D-Resnet        & 1.3811  & 0.7150  & 1.2374  & 1.1470  & 1.4047  & 1.1770  \\
                & Slow-fast        & 0.9540  & 0.9038  & 1.9338  & 1.2009  & 1.3583  & 1.2701  \\
                & TPN              & 0.9259  & 0.7596  & 1.2524  & 0.9458  & 1.4147  & 1.0597  \\
                & VAT              & 0.9183  & 0.8238  & 1.1422  & 0.8951  & 1.3437  & 1.0246  \\
                \hline
                
            \multirow{3}*{Aud-vis}
                & Deep-bimodal reg    & 0.9025 &  0.7072 &  1.4857 &  1.0316 &  1.1990 &  1.0652  \\
                & Bi-modal lstm       & 0.9512 &  0.7062 &  1.3283 &  0.8737 &  1.3429 &  1.0405  \\
                & Audiovisual-resnet  & 0.9278 &  0.7016 &  1.4492 &  0.9797 &  1.1513 &  1.0419  \\
                & CRNet               & 1.2834 &  0.8644 &  1.2876 &  1.0442 &  1.6987 &  1.2357  \\
			\bottomrule
		\end{tabular}
        }
	\end{center}
	\caption{The MSE results achieved for the true personality recognition on the UDIVA dataset. The reported results are obtained by averaging the results from frame data for four sessions.}  
\label{tb:MSE-UDIVA}
\end{table}
\setlength{\tabcolsep}{1.4pt}

\setlength{\tabcolsep}{2pt}
\begin{table}[H]
	\begin{center}
        \resizebox{1\linewidth}{!} {
		    \begin{tabular}{|l| c| c c c c c l|}
			\toprule
                & Traits         & Open    & Consc   & Extrav  & Agree   & Neuro   & Avg.            \\
                \hline \hline

            \multirow{7}*{Video} 
                & Interpret-img    &  0.0958  &  0.3056  &  0.1884  &  0.1404  & -0.2008  &  0.1059  \\
                & PersEmoN         &  0.0300  &  0.0036  & -0.0039  &  0.0037  &  0.0005  &  0.0068  \\
                & SENet            &  0.0138  &  0.3536  &  0.0710  &  0.1692  & -0.0119  &  0.1191  \\
                & HRNet            &  0.1646  &  0.1316  &  0.0621  &  0.2147  & -0.0693  &  0.1007  \\
                & Swin-transformer & -0.0655  &  0.2233  &  0.0687  &  0.0977  & -0.0280  &  0.0592  \\
                & 3D-Resnet        &  0.0168  &  0.0664  &  0.0072  & -0.0063  & -0.0754  &  0.0017  \\
                & Slow-fast        & -0.0021  & -0.0115  & -0.0001  &  0.0041  & -0.0061  & -0.0032  \\
                & TPN              &  0.0172  &  0.0559  &  0.0587  &  0.0170  &  0.0273  &  0.0352  \\
                & VAT              &  0.0758  & -0.0193  &  0.1180  &  0.0806  & -0.0362  &  0.0438  \\
                \hline
                
            \multirow{3}*{Aud-vis}
                & Deep-bimodal reg    &  0.0365  &  0.1429 &  0.0800 &  0.1000 &  0.0086 &  0.0736  \\
                & Bi-modal lstm       & -0.0005  &  0.0002 &  0.0004 &  0.0000 &  0.0004 &  0.0001  \\
                & Audiovisual-resnet  &  0.0761  &  0.2687 &  0.0655 &  0.0387 &  0.0000 &  0.0898  \\
                & CRNet               &  1.2834  &  0.8644 &  1.2876 &  1.0442 &  1.6987 &  1.2357  \\
			\bottomrule
		\end{tabular}
        }
	\end{center}
	\caption{The CCC results achieved for the true personality recognition on the UDIVA dataset. The reported results are obtained by averaging the results from frame data for four sessions.}  
\label{tb:MSE-UDIVA}
\end{table}
\setlength{\tabcolsep}{1.4pt}
